\documentclass{article}
\usepackage{caption}
\usepackage{graphicx}
\usepackage{amsmath}
\usepackage{amssymb}
\usepackage{booktabs}
\usepackage{multirow}
\usepackage{makecell}
\usepackage{pdfpages}
\usepackage{times}
\usepackage{microtype}
\usepackage{epsfig}
\usepackage{float}
\usepackage{placeins}
\usepackage{color, colortbl}
\usepackage{stfloats}
\usepackage{enumitem}
\usepackage{pifont}
\usepackage{tabularx}
\usepackage{xstring}
\usepackage{multirow}
\usepackage{xspace}

\usepackage{siunitx}

\usepackage[accepted, nohyperref]{icml2025}
\usepackage[british, american]{babel}
\usepackage{graphicx, amsmath, amssymb, subcaption, multirow, overpic, textpos}
\usepackage{array}

\definecolor{mydarkblue}{rgb}{0,0.08,0.45}
\definecolor{citecolor}{HTML}{0071BC}
\definecolor{linkcolor}{HTML}{ED1C24}
\definecolor{baselinecolor}{gray}{.9}
\definecolor{red}{HTML}{FF8988}
\definecolor{yellow}{HTML}{FECC81}
\newcommand{\baseline}[1]{\cellcolor{baselinecolor}{#1}}
\usepackage{tikz}

\usepackage{url}

\usepackage[pagebackref=false, breaklinks=true, colorlinks, citecolor=citecolor, linkcolor=linkcolor, urlcolor=mydarkblue, bookmarks=false, pdftitle={Idiosyncrasies in Large Language Models}]{hyperref}

\PassOptionsToPackage{hyphens}{url}
\newcolumntype{x}[1]{>{\centering\arraybackslash}p{#1pt}}
\newcolumntype{y}[1]{>{\raggedright\arraybackslash}p{#1pt}}
\newcolumntype{z}[1]{>{\raggedleft\arraybackslash}p{#1pt}}

\newlength\savewidth\newcommand\shline{\noalign{\global\savewidth\arrayrulewidth
  \global\arrayrulewidth 1pt}\hline\noalign{\global\arrayrulewidth\savewidth}}

\newcommand{\tablestyle}[2]{\setlength{\tabcolsep}{#1}\renewcommand{\arraystretch}{#2}\centering\footnotesize}
\renewcommand{\paragraph}[1]{\vspace{1.25mm}\noindent\textbf{#1}}

\newcommand{\cmark}{\ding{51}} 

\newcommand{\eg}{\emph{e.g}.}
\newcommand{\ie}{\emph{i.e}.}

\newcommand{\vs}{\emph{vs}.\ }

\newcommand{\gpt}{ChatGPT}
\newcommand{\claude}{Claude}
\newcommand{\grok}{Grok}
\newcommand{\gemini}{Gemini}
\newcommand{\deepseek}{DeepSeek}

\icmltitlerunning{Idiosyncrasies in Large Language Models}

\begin{document}

\twocolumn[
    \icmltitle{Idiosyncrasies in Large Language Models}
    \icmlsetsymbol{equal}{*}
    
    \begin{icmlauthorlist}
    \icmlauthor{Mingjie Sun}{equal,cmu}\;\; 
    \icmlauthor{Yida Yin}{equal,cal}\;\;
    \icmlauthor{Zhiqiu Xu}{penn}\;\;
    \icmlauthor{J. Zico Kolter}{cmu}\;\;
    \icmlauthor{Zhuang Liu}{princeton}
    \end{icmlauthorlist}

    \icmlaffiliation{cmu}{Carnegie Mellon University}
    \icmlaffiliation{princeton}{Princeton University}
    \icmlaffiliation{cal}{UC Berkeley}
    \icmlaffiliation{penn}{University of Pennsylvania}

    \icmlcorrespondingauthor{Mingjie Sun}{mingjies@andrew.cmu.edu}
    \icmlcorrespondingauthor{Yida Yin}{davidyinyida0609@berkeley.edu}

\vskip 0.3in
]

\printAffiliationsAndNotice{\icmlEqualContribution}

\begin{figure*}
  \centering
  \includegraphics[width=\linewidth]{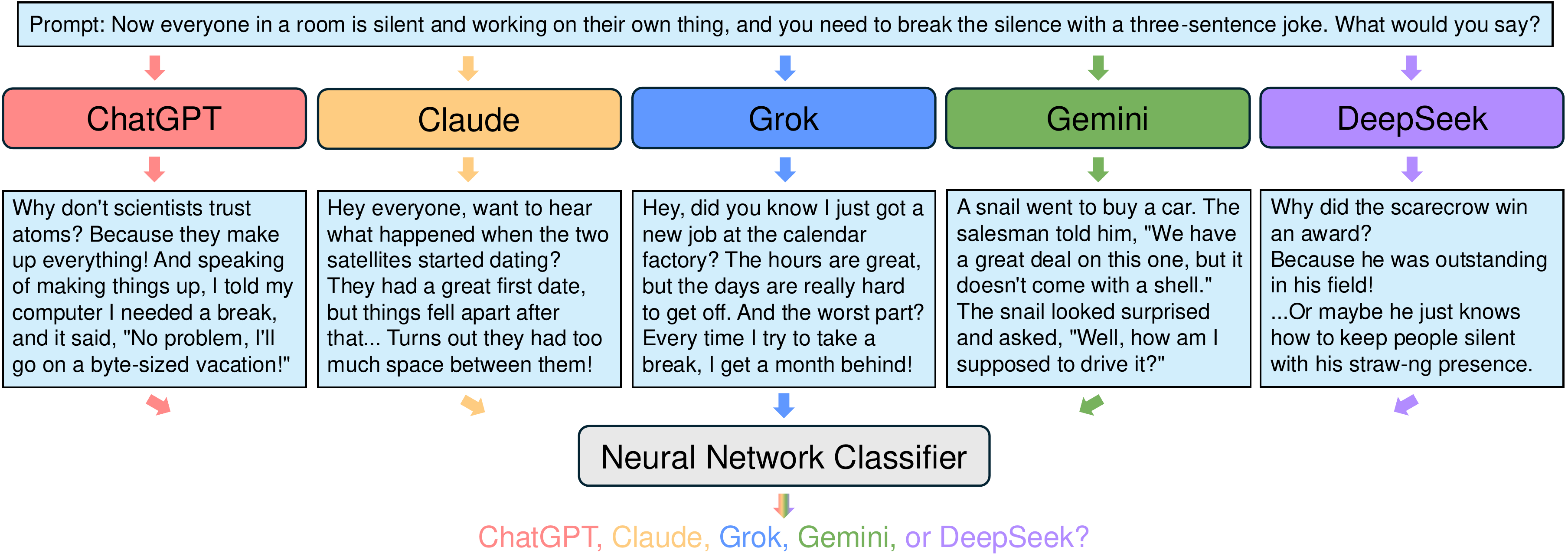}
  \vspace{-2.em}
  \caption{\textbf{Our framework for studying idiosyncrasies in Large Language Models (LLMs).} We show that  each LLM is unique in its expression. In the example shown here on \gpt, \claude, \grok, \gemini, and \deepseek, a neural network classifier is able to distinguish them with a near-perfect 97.1\% accuracy.}
  \label{fig:teaser} 
  \vspace{-.5em}
\end{figure*}

\begin{abstract}
In this work, we unveil and study idiosyncrasies in Large Language Models (LLMs) -- unique patterns in their outputs that can be used to distinguish the models. To do so, we consider a simple classification task: given a particular text output, the objective is to predict the source LLM that generates the text. We evaluate this synthetic task across various groups of LLMs and find that simply fine-tuning text embedding models on LLM-generated texts yields excellent classification accuracy. Notably, we achieve 97.1\% accuracy on held-out validation data in the five-way classification problem involving \gpt, \claude, \grok, \gemini, and \deepseek. Our further investigation reveals that these idiosyncrasies are rooted in word-level distributions. These patterns persist even when the texts are rewritten, translated, or summarized by an external LLM, suggesting that they are also encoded in the semantic content. Additionally, we leverage LLM as judges to generate detailed, open-ended descriptions of each model's idiosyncrasies. Finally, we discuss the broader implications of our findings, including training on synthetic data, inferring model similarity, and robust evaluation of LLMs. Code is available at \href{https://github.com/locuslab/llm-idiosyncrasies}{github.com/locuslab/llm-idiosyncrasies}. 
\end{abstract}

\section{Introduction}
As the adoption of generative models such as LLMs accelerates, it becomes increasingly important to understand the origin and provenance of such generated content.  While a great deal of past work has focused on the classification of human-written and AI-written content~\citep{krishna2023paraphrasing,mitchell2023detectgpt,sadasivan2023can}, there has been little work on classifying \emph{between} content generated by different LLMs, either between the outputs of entirely different models or between those of different variants of the same model family. If possible, the ability to distinguish between source models in this manner would be valuable for a number of applications: it could shed light on the relative uptake of different LLMs, beyond what is reported by individual companies, and on the nature of data used to build different models. Additionally, it could offer insights into what features of generated text are most ``unique'' to each LLM.

In this paper, we investigate whether LLMs exhibit idiosyncrasies that enable their outputs to be reliably differentiated. Inspired by recent studies on dataset bias in computer vision ~\citep{liu2024datasetbias,zeng2024bias}, which showed that images from different large-scale datasets can be accurately distinguished by standard neural networks, we consider a similar synthetic classification task to assess the separability of responses generated between different LLMs. Specifically, we sample a large number of text outputs from each LLM using the same set of prompts and then train a classifier to recognize which model generates a specific text. Figure~\ref{fig:teaser} provides an overview of our framework. The illustrated example on \gpt, \claude, \grok, \gemini, and {\deepseek} presents a five-way classification problem. 

We find that a classifier based upon simple fine-tuning text embedding models on LLM outputs is able to achieve remarkably high accuracy on this task. This indicates the clear presence of idiosyncrasies in LLMs. The observation is highly robust over a large variety of LLM combinations. For instance, trained on the combined set of texts from \gpt, \claude, \grok, \gemini, and \deepseek, a model can achieve 97.1\% classification accuracy on the \textit{held-out} validation data, compared to a 20.0\% chance-level guess. Within the same model family, we obtain a non-trivial 59.8\% accuracy across 4 model sizes in Qwen-2.5 series~\cite{qwen2025qwen25technicalreport}. Further, we observe strong out-of-distribution generalization of these classifiers when tested on responses from prompts outside the training distribution.

We observe several interesting properties of this task. When controlling the length and format of outputs through prompt instructions, we still obtain high classification accuracy. Furthermore, for post-trained LLMs, the classifier demonstrates non-trivial accuracy even with only the first few tokens of the generated text. However, when classifying generations from the \emph{same} LLM but using different sampling strategies, we achieve accuracy only slightly above the chance level. In addition, we observe certain behaviors of this task that resemble those of standard text classification, where improvements in text embeddings and availability of larger training datasets lead to better classification performance.

We analyze the sources of these idiosyncrasies by applying text transformations that isolate different levels of information. We find that randomly shuffling words in LLM-generated responses leads to only a slight decrease in classification accuracy. This suggests that a substantial portion of distinctive features is encoded in the word-level distribution. We then highlight distinct sets of characteristic phrases that are consistently associated with each LLM. We also observe that markdown formatting contributes to a moderate degree of idiosyncrasies in the LLMs following post-training.

At the same time, we obtain over 90\% accuracy when the word distribution is disrupted through transformations that preserve semantics, such as rephrasing or translating. Even with the most aggressive transformation -- summarizing, classification accuracy remains well above chance-level guess. This finding implies that semantic information also shapes the idiosyncrasies in LLMs. Through open-ended language analysis, we provide further insights into these characteristics. For instance, {\gpt} has a preference for detailed, in-depth explanations, whereas {\claude} produces more concise and direct responses, prioritizing clarity.

Last, we discuss the broader implications of our findings. One should be cautious when using synthetic data to train LLMs, as we show that many of these idiosyncrasies can be inherited in such a process. Our framework also serves as a tool for assessing model similarities among frontier models, either open-source or proprietary. In addition, we discuss how the idiosyncrasies in LLMs can be used maliciously to manipulate voting-based leaderboards, therefore highlighting the need for more robust evaluation methodologies.

\begin{table*}[th]
\centering
\normalsize 
\begin{subtable}[t]{\linewidth}
\centering
\tablestyle{4pt}{1.15}
\begin{tabular}{x{50}x{50}x{50}x{50}x{50}x{55}}
   \gpt   &  \claude  &  \grok   &   \gemini  & \deepseek &  acc. (chat) \\ 
   \shline 
\cmark & \cmark &  &  & & 99.3 \\
\cmark &  &   \cmark   &  &  & 97.7 \\
\cmark &  &        &     \cmark   & & 98.7 \\
\cmark &  &  &     & \cmark &  97.2 \\
 &   \cmark     & \cmark &  & & 99.7 \\
 & \cmark   &  &  \cmark     & & 99.6 \\
   &   \cmark     &  &    & \cmark  & 99.6 \\
 &        & \cmark &   \cmark   & &  99.4\\
  &        & \cmark &     & \cmark &  98.7 \\
     &        &  & \cmark  & \cmark &  99.9 \\
\midrule

\cmark & \cmark & \cmark & \cmark & \cmark & 97.1 \\
\\
\\
Llama   &  Gemma    &  Qwen   &   Mistral    & acc. (instruct) & acc. (base) \\ 
\shline 
\cmark & \cmark &  &   & 99.9    & 98.3    \\
\cmark &  &   \cmark     &   & 97.8 & 81.7 \\
\cmark &  &        &     \cmark  & 97.0 & 96.3 \\
 &   \cmark     & \cmark &  & 99.9 & 98.3\\
 & \cmark   &  &  \cmark      &  99.9 & 98.4 \\
 &        & \cmark &   \cmark     &  96.1 & 95.7\\
\midrule 
\cmark & \cmark & \cmark & \cmark   &  96.3 & 87.3 \\
\end{tabular}
\vspace{-13.5em}
\caption{chat APIs}
\label{tab:main_api_v2}
\vspace{13em}
\end{subtable}
\begin{subtable}[t]{\linewidth}
\centering 
\tablestyle{4pt}{1.15}
\begin{tabular}{cccccc}
\end{tabular}
\vspace{-1.5em}
\caption{instruct and base LLMs}
\vspace{-1.5em}
\label{tab:main_base_v2}
\end{subtable}
\vspace{-4ex}
\caption{\textbf{Classification accuracies for various LLM combinations.} \emph{Top}: results for chat APIs. \emph{Bottom}: results for instruct and base LLMs. Check marks (\cmark) denote the models included in each combination. We observe high classification accuracies consistently across all model combinations, indicating the presence of distinct idiosyncrasies in LLMs.}
\label{tab:main}
\vspace{-1ex}

\end{table*}

\section{Evaluating Idiosyncrasies in LLMs}\label{sec-evaluate-idiosyncrasies}
Large Language Models (LLMs) share several core characteristics. The majority of them are based on the Transformer architecture~\citep{vaswani2017attention}, which is shared by all models we consider in this paper. Second, they are trained using an auto-regressive objective~\citep{radford2019gpt2}, where they predict the next token in a sequence based on preceding context. Lastly, their training datasets significantly overlap, often incorporating vast and diverse sources such as Common Crawl, Wikipedia, and Stack Overflow. Given these similarities, it is natural to ask: do LLMs speak in the same way? If not, how can we effectively measure the degree of their differences?

To address these questions, we construct a synthetic task focused on classifying outputs from different LLMs. Consider $N$ LLMs, denoted as $f_{1},\ldots,f_{N}$, where each $f_{i}$ takes an input prompt $p$ and outputs a text completion $o$. For a given dataset $\mathcal{D}$ of prompts, the outputs produced by each LLM $f_{i}$ are denoted as  $\mathcal{O}_{i}$. We approach this problem with a straightforward setup. For $N$ output sets $\mathcal{O}_{i}$, we formulate a $N$-way classification task, where the objective is to predict which LLM produced each output. If outputs of different LLMs were drawn from the same distribution, classification accuracy would not be better than random chance. Thus, we use the classification performance of this synthetic task as a measure of idiosyncrasies in LLMs.

Our task is formulated as a sequence classification problem, for which fine-tuning BERT-style models is a common approach~\citep{Sun2019HowToFineTuneBERT}. In this work, we fine-tune a more recent and competitive sequence embedding model based on decoder-only Transformers: LLM2vec~\cite{parishad2024llm2vec}. We attach a $N$-way classification head to the extracted embeddings and use LoRA-based fine-tuning~\cite{hu2022lora} to the model weights. Input sequences are truncated to a maximum length of 512 tokens. We report the classification accuracy on a held-out validation set. Additional training details are provided in Appendix~\ref{appendix:finetune}.

\subsection{Main Observations}
\label{sec:main_results}
We observe surprisingly high accuracies by neural networks to classify LLM outputs. This observation is robust across different settings, \eg, across model families and sizes. 

We describe the LLMs we use to generate the output datasets $\mathcal{O}_{1,\cdots,N}$. For a comprehensive and fair comparison across model families, we categorize three groups of LLMs:
\setlist{nosep}
\begin{enumerate}[topsep=-2.5pt,itemsep=2pt,parsep=0pt,partopsep=0pt,leftmargin=15pt]
    \item Chat APIs (``chat''): This category includes state-of-the-art LLMs that are primarily accessible via APIs. We consider GPT-4o~\citep{openai2024gpt4o}, Claude-3.5-Sonnet~\citep{anthropic2024}, Grok-2~\cite{grok2}, Gemini-1.5-Pro~\cite{team2024gemini}, and DeepSeek-V3~\cite{deepseekai2024deepseekv3technicalreport}. For simplicity, we refer to them as ChatGPT, Claude, Grok, Gemini and DeepSeek. Their architectures and weights remain proprietary and undisclosed, with the exception of DeepSeek. 
    \item Instruct LLMs (``instruct''): These models are trained to generate high-quality responses from human instructions. We consider four LLMs of similar sizes across different families: Llama3.1-8b~\citep{Llama3}, Gemma2-9b~\citep{gemma2}, Qwen2.5-7b~\citep{qwen2025qwen25technicalreport} and Mistral-v3-7b~\citep{jiang2023mistral}. We will refer to them as Llama, Gemma, Qwen and Mistral. 
    \item Base LLMs (``base''): These are base versions of instruct LLMs. They are obtained by pretraining on extensive text corpora without any post-training stage.
\end{enumerate}

Throughout the paper, we refer to these three categories as “chat”, “instruct”, and “base” respectively. For a given prompt dataset, we collect 11K text sequences, splitting them into 10K and 1K as training and validation sets, respectively. The same split is used across all LLMs. For chat APIs and instruct LLMs, we generate outputs from UltraChat~\citep{ding2023ultrachat}, a diverse dialogue and instruction dataset. For base LLMs, we synthesize new texts using prompts from FineWeb~\citep{penedo2024fineweb}, a high-quality LLM pretraining dataset. More details on response generation are in Appendix~\ref{appendix:response_gen}.

\textbf{Across model families.} In Table~\ref{tab:main}, we report the results for classifying outputs from various combinations of chat APIs (Table~\ref{tab:main_api_v2}) and instruct / base LLMs (Table~\ref{tab:main_base_v2}). In each of the three LLM groups, we enumerate all ($C_N^2$) possible pairwise combinations when choosing 2 out of $N$ models in the top panel of each table, as well as the case including $N$ models in the bottom row. For the binary classification task, the neural network consistently achieves over 90\% accuracy, with only one exception. Notably, for chat APIs and instruct LLMs, many combinations reach as high as 99\% accuracy. In the more challenging $N$-way classification tasks, our classifiers maintain strong performance, achieving at least 87.3\% accuracy across three groups. These results highlight the idiosyncrasies across different LLMs. We refer readers to Appendix~\ref{appendix:section_confusion_matrix} for the confusion matrices of our classifiers.

\textbf{Within the same model family.} We evaluate sequence classification performance when distinguishing responses from LLMs within the same model family. Note that models from the same family typically share common training procedures, \eg, pretraining datasets and optimization schedule. First, we analyze the impact of model size by considering four Qwen2.5 instruct LLMs with 7B, 14B, 32B, and 72B parameters. As shown in Table~\ref{tab:main_model_size}, the classification task is more difficult here, but our classifiers remain well above chance accuracy when distinguishing LLMs within the same family. In the binary classification setup, the highest accuracy reaches 85.5\%, whereas in the full combination setup, the accuracy becomes 59.8\%. In addition, we observe high accuracies when classifying responses from base and instruct versions of the same model. For example, our classifiers achieve 96.8\% accuracy when distinguishing outputs from Qwen2.5-7b base and instruct models.

\begin{table}[h]
\centering
\vspace{-1ex}
\tablestyle{6pt}{1.15}
\begin{tabular}{ccccc}
   7b   &  14b    &  32b   &   72b    & instruct \\ 
   \shline
\cmark & \cmark &  &      &    77.0 \\
\cmark &  &   \cmark     &   &  81.2  \\
\cmark &  &        &     \cmark &   83.4 \\
 &   \cmark     & \cmark &  &  63.1     \\
 & \cmark   &  &  \cmark       & 85.5      \\
 &        & \cmark &   \cmark    &  84.8 \\
 \hline
\cmark & \cmark & \cmark & \cmark  &  59.8 \\
\end{tabular}
\vspace{-.5ex}
\caption{\textbf{Classification within Qwen2.5 model family.} The classifier can differentiate responses between LLMs within the same model family with reasonably well accuracies.
}
\label{tab:main_model_size}
\end{table}

\textbf{Generalization to out-of-distribution responses.} We find that our classifiers generalize robustly to responses beyond their training distribution. To evaluate this, we collect responses from instruct LLMs across four diverse datasets: \ie,  UltraChat, Cosmopedia~\cite{benallal2024cosmopedia}, LmsysChat~\cite{zheng2024lmsyschat1mlargescalerealworldllm}, and WildChat~\cite{zhao2024wildchat1mchatgptinteraction}. These datasets originate from different sources and are designed for various purposes -- Cosmopedia is intended for synthetic data generation, LmsysChat and WildChat capture real-world user interactions, while UltraChat consists primarily of synthetic responses. For each dataset, we train a classifier on a group of model responses and evaluate the classifier on the remaining three datasets. Table~\ref{tab:main_it_ood} shows the results on instruct LLMs. Our classifiers generalize well across different datasets, indicating that they learn very robust and transferable patterns.

\begin{table}[ht]
\centering
\small 
\tablestyle{2pt}{1.15}
\begin{tabular}{lcccc}
 train / test  & UltraChat & Cosmopedia & LmsysChat & WildChat \\
    \shline
UltraChat  & 96.3 & 98.9 & 89.9 & 92.4\\
Cosmopedia & 95.7 & 99.8 & 88.3 & 94.9 \\
LmsysChat  & 94.7 & 97.2 & 91.8 & 92.0 \\
WildChat   & 95.1 & 99.1 & 90.2 & 95.7 \\
\end{tabular}

\vspace{-.5ex}
\caption{\textbf{Robust generalization to out-of-distribution responses.} We train classifiers on LLM outputs from one prompt dataset and tested on those from another.
}
\label{tab:main_it_ood}
\end{table}

\subsection{Controlled Experiments}\label{sec:main_control}
We analyze the behaviors of the synthetic classification task in several controlled settings. \textit{From now on, we only report accuracies of the $N$-way classification task in each group.}

\textbf{Prompt-level interventions.} We assess the degree of idiosyncrasies in LLM outputs with explicit prompt-level interventions. Specifically, we modify the original prompt by incorporating additional instructions to constrain response length and format. We then perform sequence classification on the resulting outputs. Our interventions are: 
\begin{itemize}[topsep=-1pt, ,itemsep=0pt,parsep=0pt,partopsep=-0.5pt, leftmargin=12pt]
    \item Length control: \textit{Please provide a concise response in a single paragraph, limited to a maximum of 100 words.}
    \item Format control: \textit{Please provide your response in plain text only, avoiding the use of italicized or bold text, lists, markdown, or HTML formatting.}
\end{itemize}
LLM outputs after these interventions are presented in Appendix~\ref{appendix:response_demonstration}. We find that LLMs can follow the additional instructions in generating responses.

\begin{table}[ht]
\centering
\small 

\tablestyle{3.pt}{1.15}
\begin{tabular}{lccc}
   & original & length control & format control  \\
  \shline
instruct LLMs   & 96.3 & 93.0 & 91.4 \\

\end{tabular}
\caption{\textbf{Controlling LLM outputs with prompts.} An instruction is added to the original prompt to specify the output length and format. \textit{Length control} limits responses to one paragraph. \textit{Format control} ensures that responses are in plain text without any format.
}
\label{tab:main_prompt}
\end{table}

The results are shown in Table~\ref{tab:main_prompt}, where ``original'' means the classification accuracy without interventions. We can see that neural networks still perform excellently for classifying LLM outputs applied with length and format control prompts. These findings suggest that LLM characteristics are deeply embedded in the generated text, persisting despite surface-level constraints on length and formatting.

\begin{figure}[t]
\centering
    \includegraphics[width=.98\linewidth]{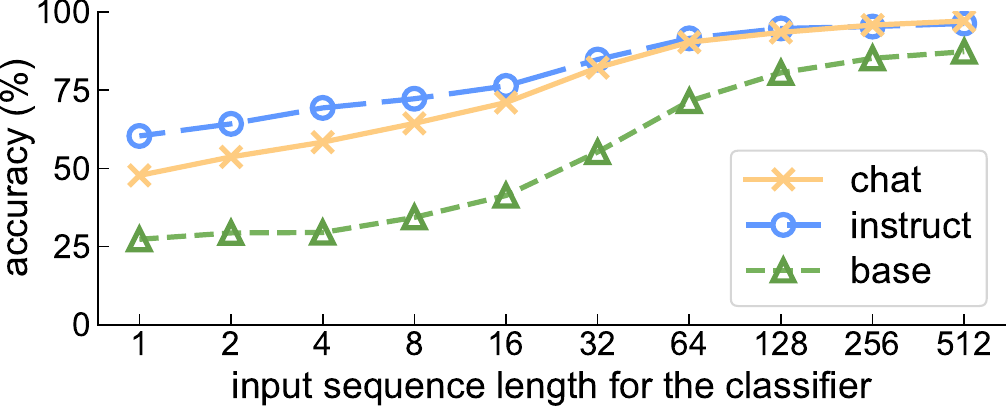}
    \vspace{-.5em}
    \caption{\textbf{Ablations on input length of text embedding models.} Classification accuracies improve as the text embedding models capture more context. Performance begins to saturate beyond an input sequence length of 256. Note that the three lines represent different groups of LLMs and are not directly comparable.}
    ~\label{fig:seq_len}
    \vspace{-1em}
\end{figure}

\textbf{Input length of text embedding models.} We control the number of input tokens to the text embedding models. Specifically, we truncate each response to a fixed number of tokens in a left-to-right fashion. Figure~\ref{fig:seq_len} presents the results. Across three groups of LLMs, the classification task benefits from seeing an increased number of tokens. Intriguingly, for chat APIs and instruct LLMs, we observe around 50\% accuracy using only a single text token. This suggests that the initial token in a response contains certain distinctive signals for the classification problem. In Section~\ref{sec:words_letters}, we provide further evidence supporting this observation.

\textbf{Sampling methods.} We consider outputs when sampled using different decoding strategies. Specifically, we use four widely used sampling methods: greedy decoding, temperature softmax, top-k, and top-p sampling. For each method, we generate a set of responses from the same LLM. We then fine-tune the LLM2vec embedding model to predict the sampling method responsible for each response. 

\begin{table}[h]
\vspace{-1em}
\centering
\begin{tabular}{lcccc}
 & greedy & softmax & top-k & top-p \\ \shline 
 greedy & - & - & - & - \\
 softmax & 59.6 & - & - & -\\
 top-k   & 58.2 & 50.0 & - & -\\
 top-p & 52.9 & 51.0 & 52.1 & -\\
\end{tabular}
\vspace{-0.3em}
\caption{\textbf{Classifications with different sampling methods.} Distinguishing responses generated by the same model using different sampling strategies is only marginally better than chance accuracy. The results are on Llama3.1-8b instruct model's responses.}
\label{tab:sampling}
\end{table}

Table~\ref{tab:sampling} presents the results for all pairs of sampling methods. Notably, the accuracy of distinguishing between responses generated by the same LLM remains relatively low, with the highest accuracy across all configurations being 59\%. Furthermore, in a more fine-grained 5-way classification task distinguishing softmax sampling at five different temperatures ($T=$ 0, 0.25, 0.5, 0.75, 1), we obtain an accuracy of 37.9\%, only marginally better than the random chance level of 20\%. These results suggest that outputs from the same LLM are not easily separable based on decoding strategies.

\textbf{Text embedding models.} 
We vary the underlying pretrained embedding models for sequence classification. The default setting we used in previous parts is fine-tuning the LLM2vec embedding models. We consider various generations of embeddings models spanning across architectures and training methods: ELMo~\cite{peters2018deepcontextualizedwordrepresentations},  BERT~\cite{devlin2018bert}, T5~\citep{raffel2023T5}, GPT-2~\citep{radford2019gpt2}, and LLM2vec~\citep{parishad2024llm2vec}. 
Details on the fine-tuning setting can be found in Appendix~\ref{appendix:finetune}. 

Table~\ref{tab:main_embedding} shows the results. All sequence embedding models can achieve very high accuracies. The classification performance improves with more advanced sequence embedding models. Among all methods, LLM2vec demonstrates the best performance, achieving 97.1\% on chat APIs, 96.3\% on instruct LLMs, and 87.3\% on base LLMs. 

\begin{table}[h]
\centering
\small 
\tablestyle{6pt}{1.15}
\begin{tabular}{lccc}
    method & chat & instruct & base \\
    \shline 
    ELMo & 90.8 & 91.0 & 69.8 \\
    BERT   & 91.1 & 91.5 & 66.0 \\
    T5 & 90.5 & 89.8 & 67.9 \\
    GPT-2 & 92.1 & 92.3 & 80.2 \\
    LLM2vec & \textbf{97.1} & \textbf{96.3} & \textbf{87.3}\\
\end{tabular}
\vspace{-.25em}
\caption{\textbf{Different sequence embedding models.} LLM2vec achieves the best performance in classifying outputs from various LLMs among the five embedding models we study.}
\label{tab:main_embedding}
\end{table}

\begin{figure}[h]
    \vspace{-.5em}
\centering 
    \includegraphics[width=\linewidth]{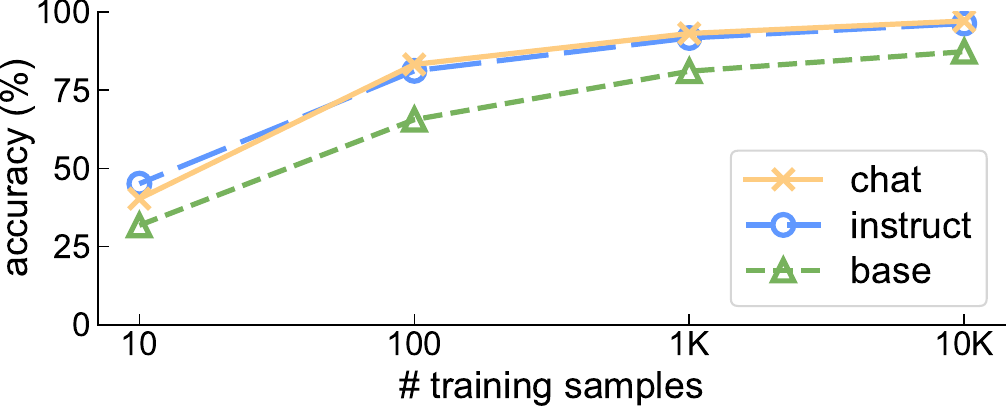}
    \vspace{-2em}
    \caption{\textbf{Different numbers of training samples.} Our sequence classifiers benefit from more training samples. The classification performance converges when using about 10K training samples.}~\label{fig:train_samples}
    \vspace{-1em}
\end{figure}

\textbf{Training data size.} We vary the number of training samples generated by LLMs and train the classifier with the same total number of iterations. We present the results in Figure~\ref{fig:train_samples}. The classification performance increases with more training samples. This trend holds consistently across chat APIs, instruct LLMs, and base LLMs. Furthermore, as few as 10 samples, the classifier achieves non-trivial accuracy (\eg, 40.3\% on chat APIs), surpassing 20\% chance-level guess.

\section{Concrete Idiosyncrasies in LLMs}

We have shown that modern neural networks can achieve excellent accuracies in classifying which LLM generates a given response. 
Here we use text similarity metrics to quantify differences between LLM outputs. We consider three standard metrics -- ROUGE-1~\cite{lin-2004-rouge}, ROUGE-L~\cite{lin-2004-rouge}, and BERTScore~\cite{devlin2018bert} -- to measure lexical and semantic similarity. We compute the mean F1-score for each metric across all response pairs generated by any two different chat API models given the same prompt. For comparison, we also measure the similarity between responses sampled within the same model. As shown in Table~\ref{tab:text_sim}, responses from different LLMs exhibit lower text similarities than those from the same model. 

\begin{table}[t]
    \vspace{-2ex}
    \tablestyle{6pt}{1.15}
    \begin{tabular}{lccc}
    & across LLMs & within an LLM\\
    \shline
    ROUGE-1   & 0.499 & 0.660 \\
    ROUGE-L   & 0.256 & 0.414 \\
    BERTScore$^*$ & 0.220 & 0.482
    \end{tabular}
    \vspace{-1ex}
    \caption{\textbf{Text similarity scores.}  We evaluate the text similarity of LLM outputs using ROUGE-1, ROUGE-L, and BERTScore. Responses from different LLMs exhibit low lexical similarity.}
    \label{tab:text_sim}
    \vspace{-1ex}
\end{table}

In the following, we identify concrete idiosyncrasies in LLMs across three dimensions: words and letters, markdown formatting elements, and semantic meaning. For each dimension, we apply text transformations to isolate potential idiosyncrasies and assess their impacts on classification performance. We then highlight specific patterns within each dimension that distinguish LLMs.

\begin{figure*}[th]
    \centering
    \begin{subfigure}[h]{0.49\textwidth}
    \centering
        \includegraphics[width=\linewidth]{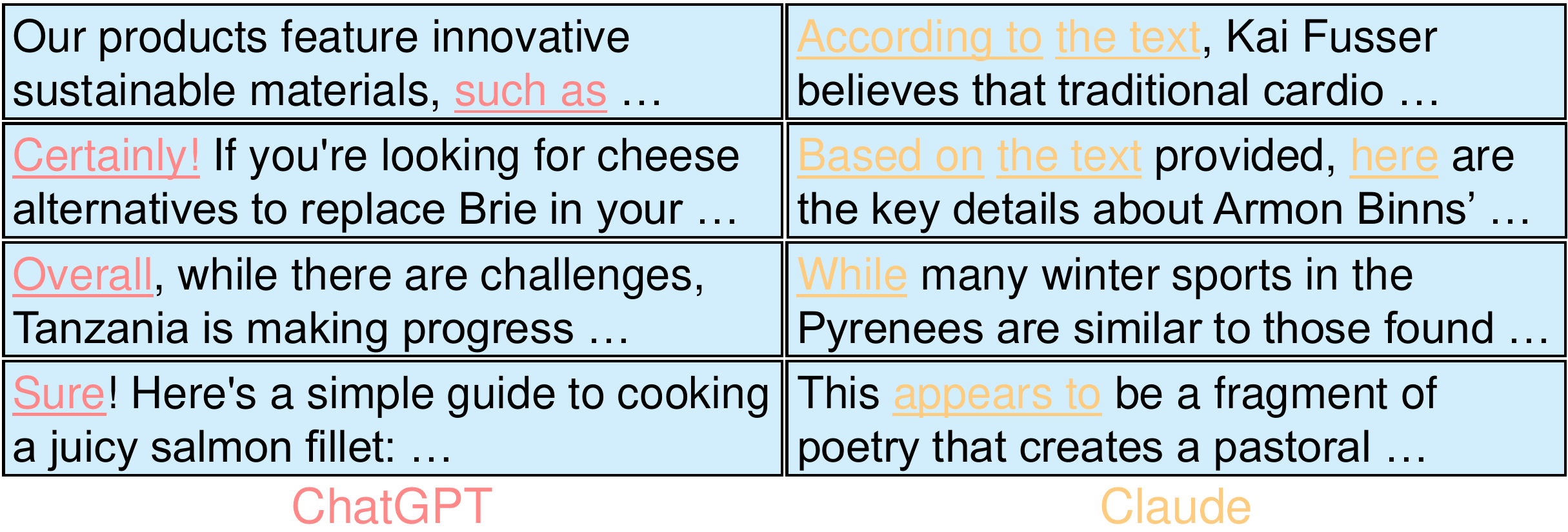}
    \vspace{-1.5em}
    \subcaption{characteristic phrases}
    \label{fig:example_phrases}
    \end{subfigure}~
    \begin{subfigure}[h]{0.49\textwidth}
    \centering
        \includegraphics[width=\linewidth]{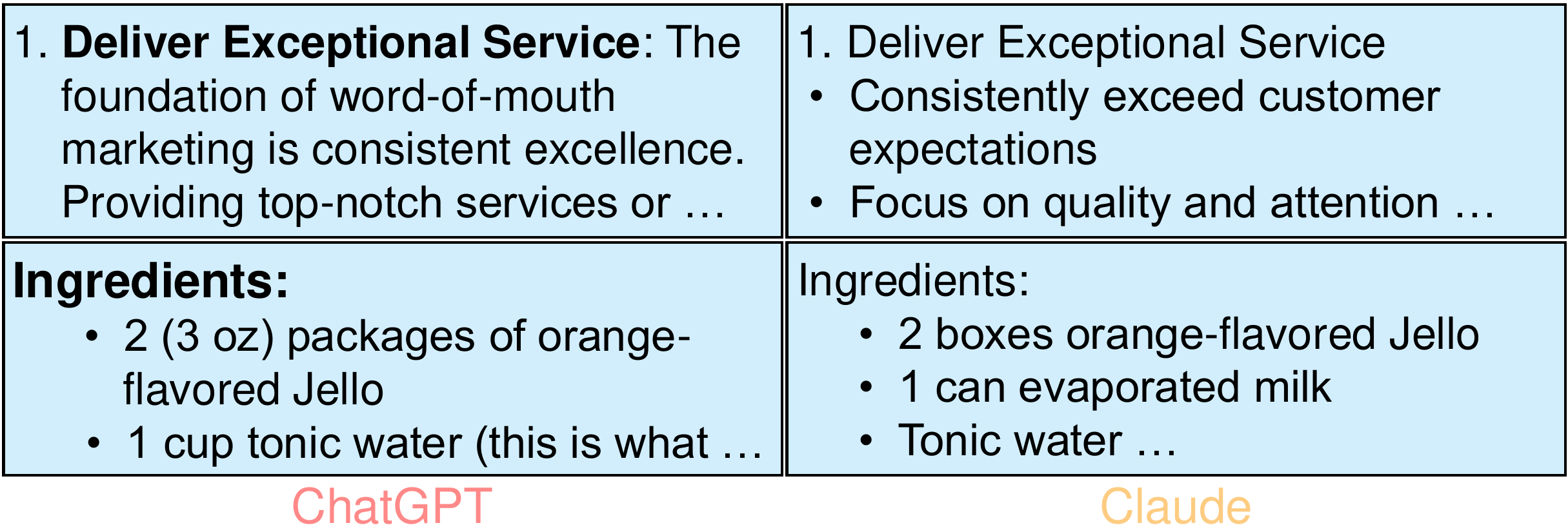}
    \vspace{-1.5em}
    \subcaption{unique markdown formatting}
    \label{fig:example_special_characters}
    \end{subfigure}
    \vspace{-.5em}
    \caption{\textbf{Example responses from \textcolor{red}{\gpt} and \textcolor{yellow}{\claude}, showcasing their idiosyncrasies}: characteristic phrases (\textit{left}) and unique markdown formatting (\textit{right}). For clarity, we highlight each characteristic phrase with \underline{underline} and model-specific color.
    }
    \label{fig:example_responses}
\end{figure*}

\subsection{Words and Letters}
\label{sec:words_letters}

\textbf{Text shuffling.} To decouple the effects of words and letters from other factors, we remove special characters in LLM-generated responses, such as punctuations, markdown elements, and excessive white spaces. This ensures that each response consists solely of words separated by a white space. Additionally, we apply two shuffling strategies to the preprocessed text: word-level and letter-level shuffling. These transformations disrupt the natural order and force the classifier to learn patterns from raw text statistics. Table~\ref{tab:shuffle} presents the classification results. 

\begin{table}[bp]
    \vspace{-2ex}
    \centering
    \tablestyle{6pt}{1.15}
    \begin{tabular}{lccc}
    & chat & instruct & base \\
    \shline
    original  & \baseline{97.1} & \baseline{96.3} & \baseline{87.3} \\
    removing special characters & 95.1 & 93.8 & 75.4 \\
    shuffling words             & 88.9 & 88.9 & 68.3 \\
    shuffling letters        & 39.1 & 38.6 & 38.9
    \end{tabular}
    \caption{\textbf{Classifications with only words and letters.} While removing special characters and shuffling words have little impact on accuracies, shuffling letters greatly reduces the performance.}
    \label{tab:shuffle}
\end{table}

\begin{figure}[h]
    \vspace{-2ex}
    \centering
    \includegraphics[width=.495\linewidth]{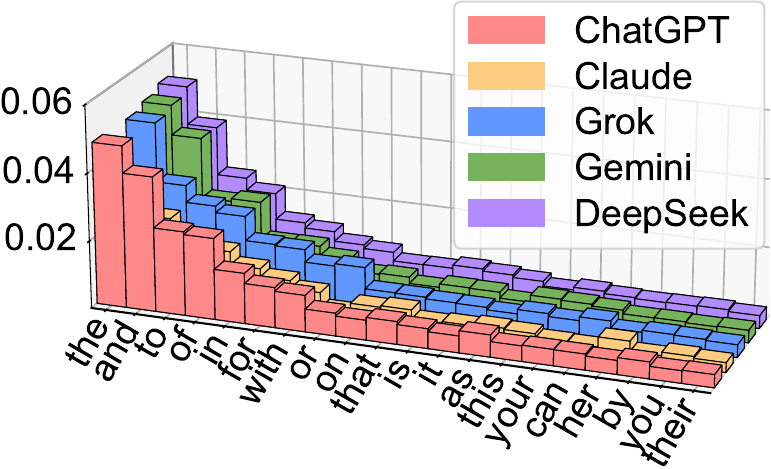}
    \includegraphics[width=.495\linewidth]{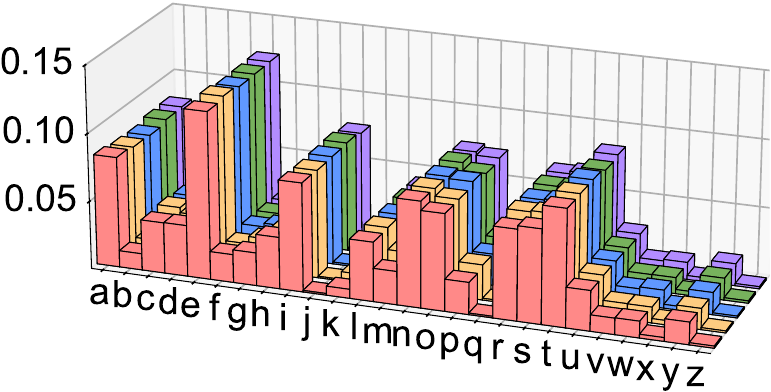}
    \caption{\textbf{Frequencies of words and letters.} The top 20 most frequently used words of LLMs (left) exhibit distinct patterns for each model, but their letter frequencies (right) are very similar. Results are on the chat API models.}
    \label{fig:word_letter_dist}
    \vspace{-2ex}
\end{figure}

Classifiers trained on responses without special characters achieve accuracies close to those using the original responses, \ie, 95.1\% for chat APIs, 93.8\% for instruct LLMs, and 75.4\% for base LLMs. Likewise, using word-shuffled responses yields high accuracies comparable to the original ones. Further, we plot the frequencies of several commonly used words from five chat APIs in Figure~\ref{fig:word_letter_dist} (\textit{left}). We observe distinct patterns among models, even for frequent English words: Claude has much lower frequencies for words like ``the'', ``and'', ``to'', and ``of'' than other chat APIs. These results suggest that \emph{special characters and word order are not essential for distinguishing LLMs}; \emph{word choices reflect substantial idiosyncrasies across models.}

In contrast, shuffling at the letter level results in a substantial drop in accuracy (49\%-56\%), approaching chance-level performance. This indicates that letter-level statistics alone are not sufficient for predicting LLM identities. To qualitatively visualize distinctions in letter distributions across models, Figure~\ref{fig:word_letter_dist} (\textit{right}) shows the frequency distribution of letters in responses generated by chat APIs. Different LLMs share almost identical letter distributions, indicating that \emph{letters contribute minimally to idiosyncrasies in LLMs.}

\begin{figure}[t]
    \centering
    \includegraphics[width=.98\linewidth]{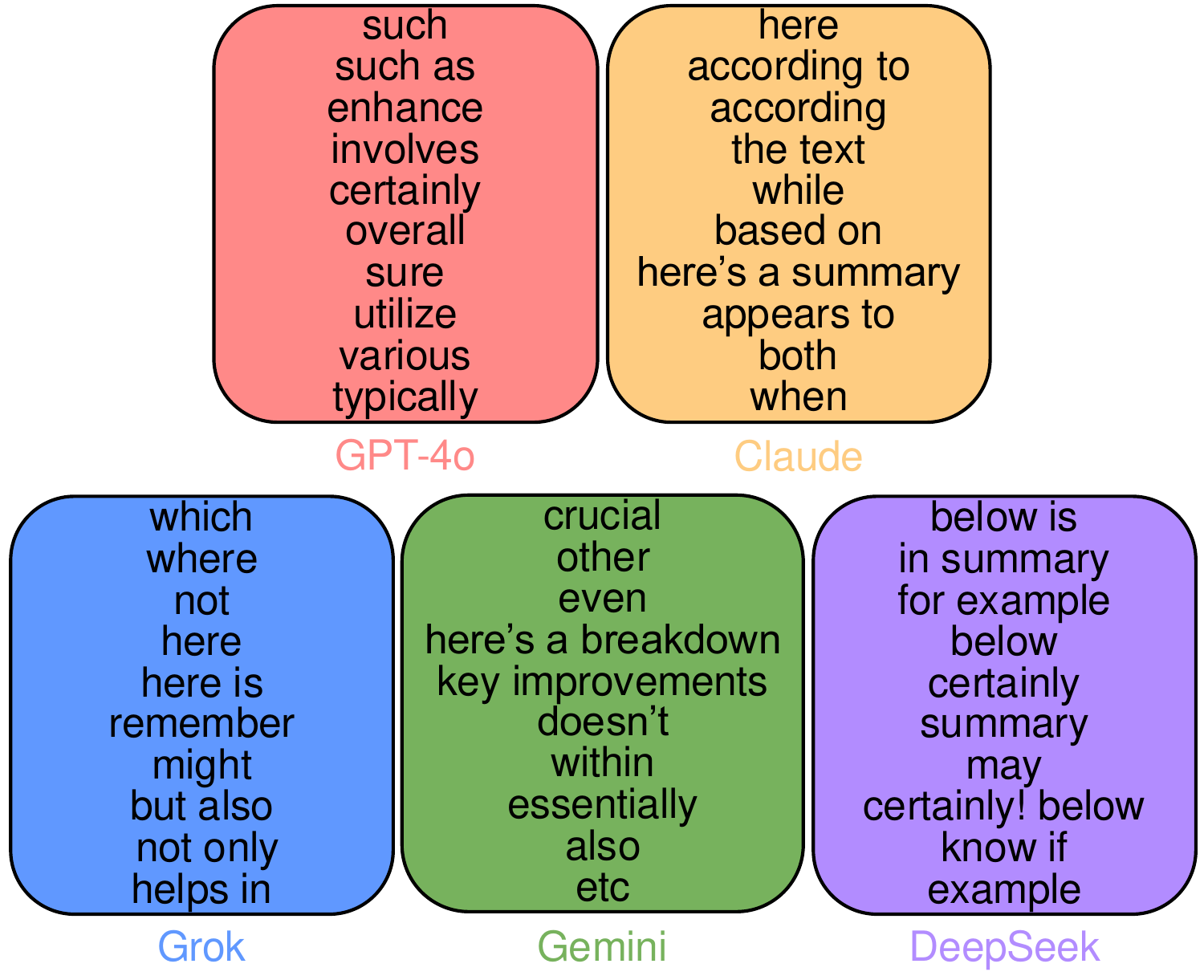}
    \vspace{-.5em}
\caption{\textbf{Characteristic phrases.} We train a logistic regression model on TF-IDF features of chat APIs' outputs and extract the top 10 phrases for each LLM based on the coefficients of these features. We remove common words shared across these LLMs.}

\vspace{.2em}
\label{fig:top_words}
\end{figure}

\begin{figure}[t]
    \centering
    \includegraphics[width=\linewidth]{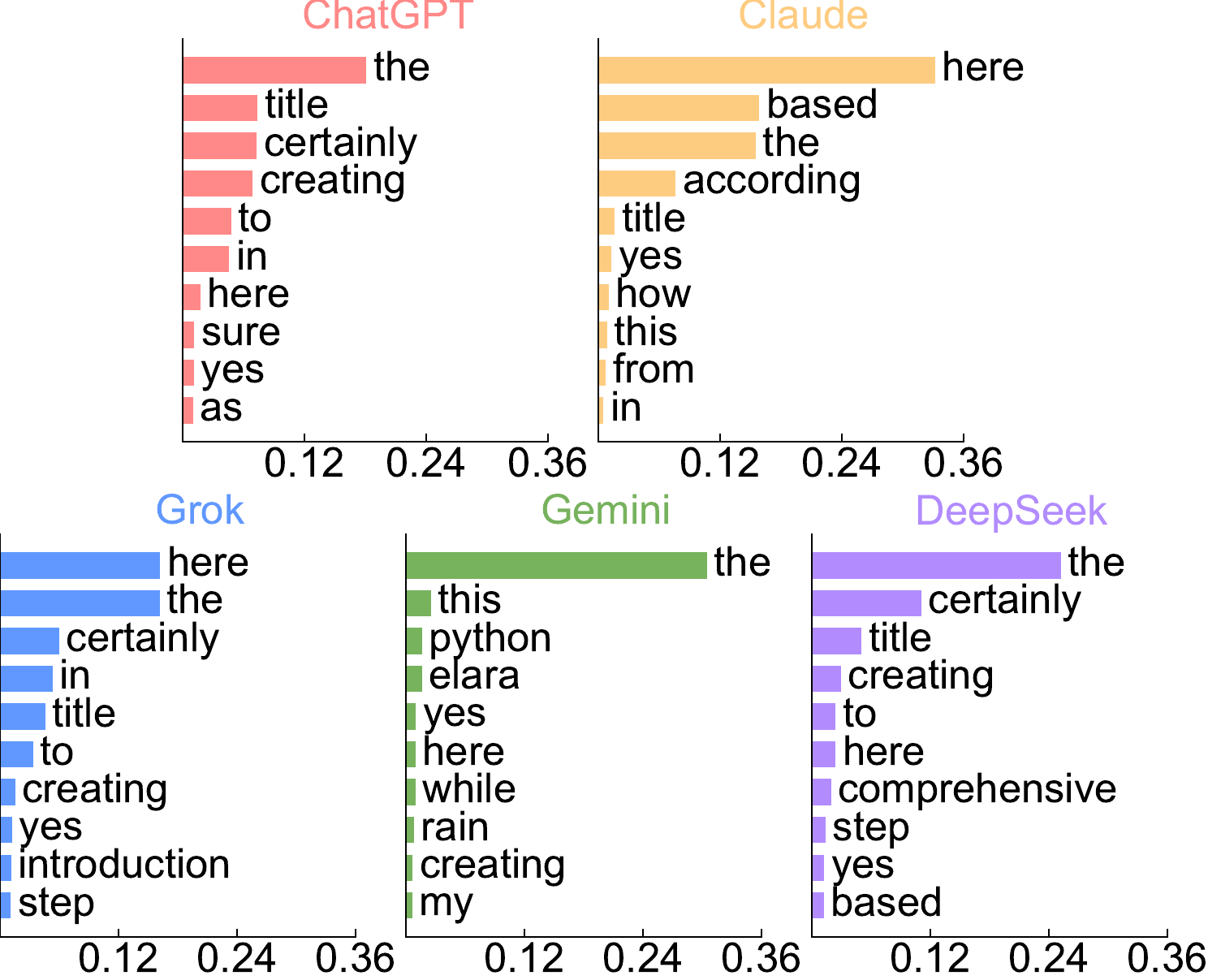}
    \vspace{-2em}
    \caption{\textbf{First word.} We analyze the distribution of the first word in chat APIs' responses, with the top 10 most frequent words for each model. These differences in the first-word usage explain the non-trivial accuracy with only the first word in Figure~\ref{fig:seq_len}.}
    \label{fig:first_words_chat}
\end{figure}

\textbf{Characteristic phrases.} We use Term Frequency-Inverse Document Frequency (TF-IDF) to highlight characteristic phrases inside LLM-generated responses that reflect each model's word choices. Formally, we treat each LLM response as a document and then extract TF-IDF features on all uni-gram and bi-gram words. We then train a $N$-way logistic regression model to predict the origin of responses on the extracted features. This simple linear classifier achieves 85.5\% / 83.7\% accuracy on chat APIs / instruct LLMs, close to 95.1\% / 93.8\% achieved with fine-tuning embedding models on responses without special characters (Table~\ref{tab:shuffle}).

Since the coefficients of a logistic regression model provide a natural ranking for its features, we leverage these coefficients to highlight important phrases in the classification task. Figure~\ref{fig:top_words} presents the top 10 phrases with the largest logistic regression coefficients for each of the five chat API models (excluding the 20 most frequently used words in Figure~\ref{fig:top_words}). Notably, these phrases often serve as transitions or emphasis in sentences. For example, {\gpt} likes to generate ``such as", ``certainly'', and ``overall'', whereas {\claude} prefers ``here'', ``according to'', and ``based on''.

Figure~\ref{fig:example_phrases} illustrates these characteristic phrases with example responses from {\gpt} and {\claude}. While {\gpt} begins responses with ``certainly'' and ``below is'', Claude usually references the original prompt using the phrases like ``according to the text'' and ``based on the text''. Moreover, Figure~\ref{fig:first_words_chat} reveals noticeable differences in the distribution of first word choices among chat APIs. Appendix~\ref{appendix:top_phrases} provides characteristic phrases for other LLMs.

\subsection{Markdown Formatting}
\label{sec:markdown}

\begin{table}[bp]
    \vspace{-1ex}
    \centering
    \tablestyle{6pt}{1.15}
    \begin{tabular}{lccc}
    & chat & instruct & base \\
    \shline
    original                & \baseline{97.1} & \baseline{96.3} & \baseline{87.3} \\
    markdown elements only & 73.1            & 77.7            &38.5 \\
    \end{tabular}
    \vspace{-.5ex}
    \caption{\textbf{Classifications with only markdown elements.} Using markdown elements can achieve high accuracies for chat APIs and instruct LLMs, but marginally better results for base LLMs.}
    \label{tab:remove_text}
    \vspace{-1ex}
\end{table}

We seek to understand how each LLM formats their responses, particularly in markdown. To this end, we focus on common markdown elements used by LLMs: (1) bold text, (2) italic text, (3) header, (4) enumeration, (5) bullet point, (6) code block. We transform the LLM outputs by retaining only these formatting components while replacing other text with the marker ``xxx''. Appendix~\ref{appendix:response_demonstration} provides examples of the transformed outputs. Table~\ref{tab:remove_text} shows the classification results after this transformation.

Surprisingly, we observe our classifiers achieve high accuracies of 73.1\% for chat APIs and 77.7\% for instruct LLMs. However, the classification accuracies with base LLMs' responses are near chance-level guess (25\%). This is likely because base LLMs tend to generate responses in plain text.

\begin{figure}[h]
    \centering
    \includegraphics[width=0.49\linewidth]{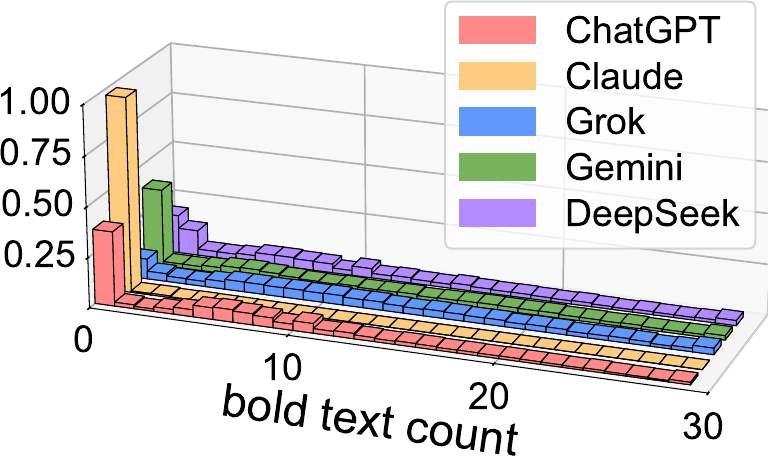}
    \includegraphics[width=0.49\linewidth]{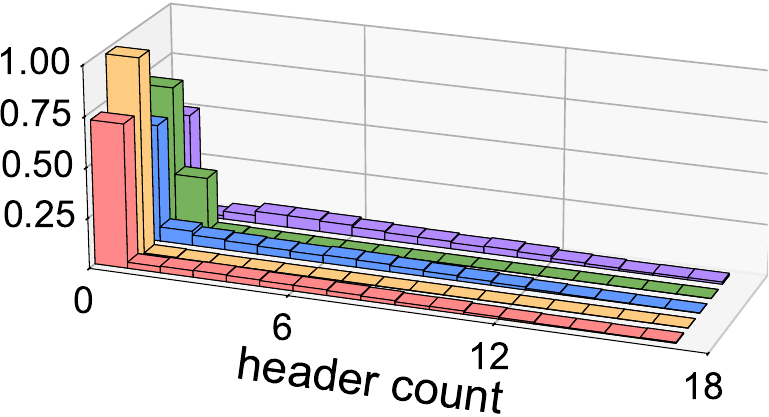}
    \includegraphics[width=0.49\linewidth]{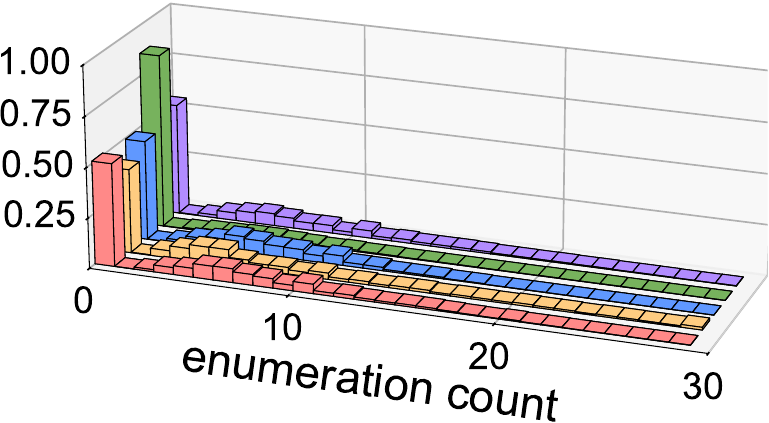}
    \includegraphics[width=0.49\linewidth]{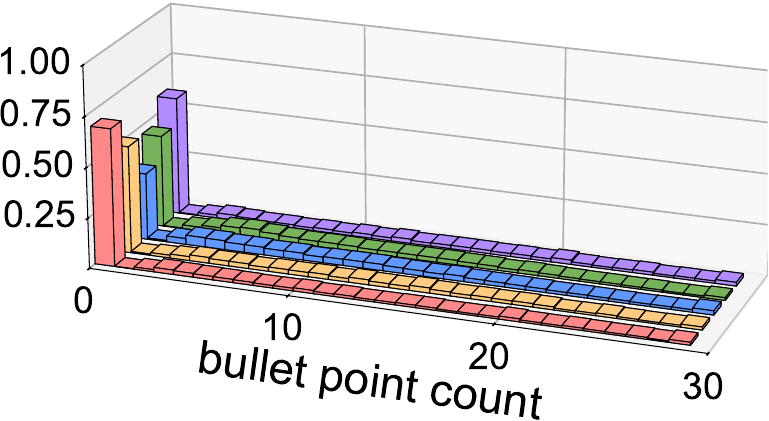}
    \caption{\textbf{Markdown formatting elements.} Each LLM has a distinctive distribution of markdown formatting elements.}
    \label{fig:features_dist}
\end{figure}

We count the occurrence of a markdown formatting element in each response and then plot the distribution of these counts over all responses in Figure~\ref{fig:features_dist}. Each model exhibits a unique way to format its responses. For instance, Claude (\textcolor{yellow}{\textbf{yellow}}) has a high density at zero in the bold text and header count distributions, indicating that it generates many responses without bold texts or headers. On the contrary, other LLMs exhibit lower values at zero and thus decorate text with these formatting elements more often.

Figure~\ref{fig:example_special_characters} visualizes how {\gpt} and {\claude} structure their responses in markdown. Interestingly, {\gpt} tends to emphasize each key point within enumerations in bold and highlight a title with markdown headers, but {\claude} formats text with simple enumeration and bullet points. More analysis for other models can be found in Appendix~\ref{appendix:markdown}.

\begin{figure*}[t]
    \centering
    \includegraphics[width=.49\linewidth]{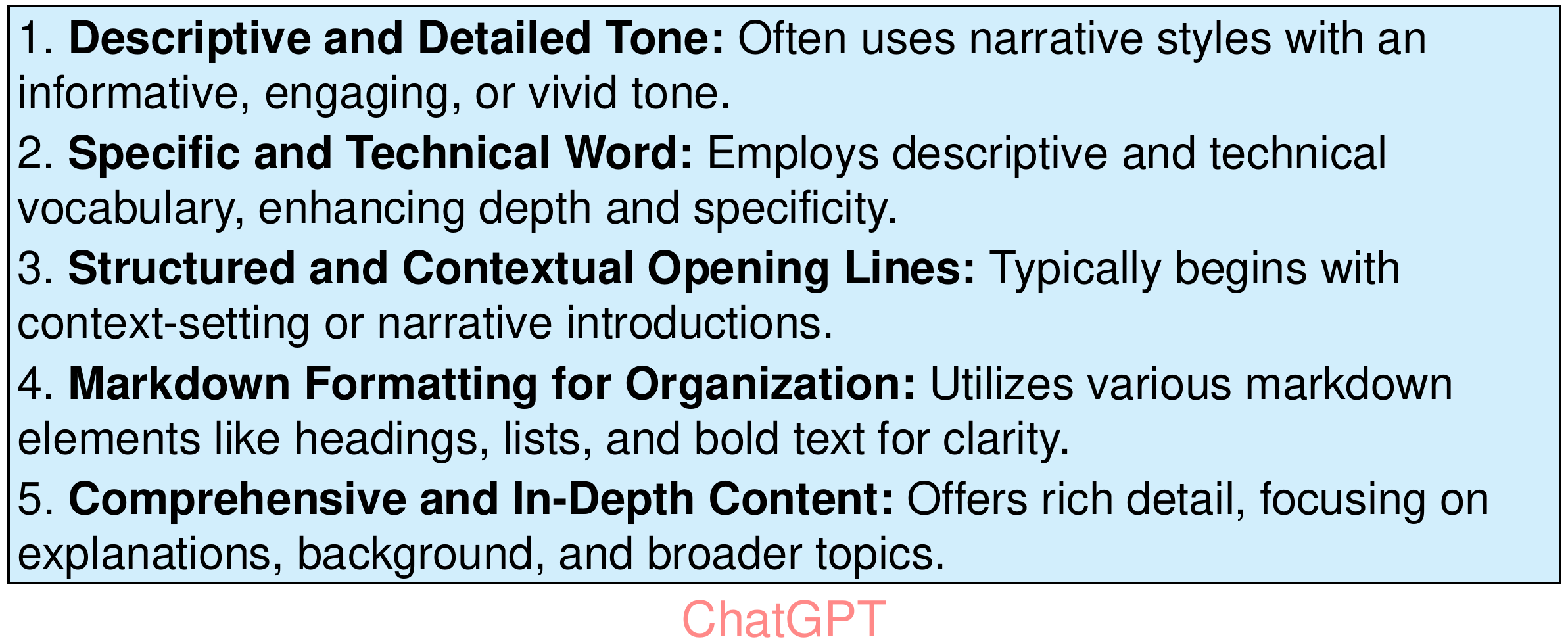}~
    \includegraphics[width=.49\linewidth]{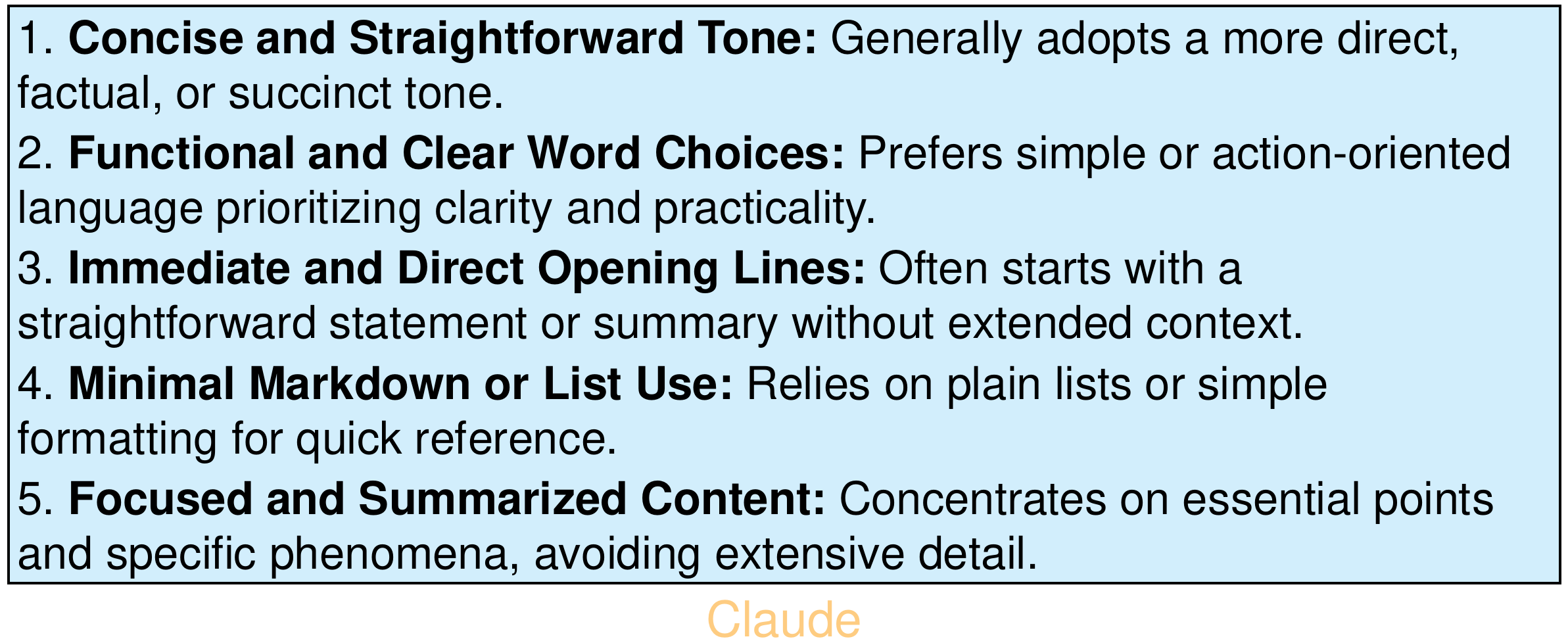}
    \vspace{-.5em}
    \caption{\textbf{Results of our open-ended language analysis on \textcolor{red}{\gpt} and \textcolor{yellow}{\textbf{\claude}}.} \textcolor{red}{\textbf{\gpt}} features descriptive language, sophisticated markdown formatting, and in-depth details, while \textcolor{yellow}{\textbf{\claude}} highlights straightforward tone, minimal structure, and summarized content.}
    \label{fig:open-ended}
\end{figure*}

\subsection{Semantics}
\label{sec:semantics}
\textbf{Rewriting.} One potential reason for the high classification accuracy is the unique writing style (\eg, word choice, sentence structure) of each LLM. To isolate this factor, we leverage another LLM (\eg, GPT-4o mini) to rewrite LLM responses. Our rewriting approaches include (see Appendix~\ref{appendix:response_demonstration} for example  responses after rewriting):

\begin{itemize}[topsep=0.pt, itemsep=0.7pt,parsep=0pt,partopsep=-.5pt,leftmargin=12pt]
    \item Paraphrasing: \emph{Paraphrase the above text while maintaining the semantic meaning of the original text.}
    \item Translating: \emph{Translate the above text into Chinese.}
    \item Summarizing: \emph{Summarize the above text in one paragraph.}
\end{itemize}

\begin{table}[h]
    \vspace{-1ex}
    \centering
    \tablestyle{6pt}{1.15}
    \begin{tabular}{lccc}
    & chat & instruct & base \\
    \shline
    original  &  \baseline{97.8} & \baseline{96.3} & \baseline{87.3} \\
    paraphrasing    & 91.4 & 92.2 & 71.7 \\
    translating & 91.8 & 92.7 & 74.0 \\ 
    summarizing   & 58.1 & 57.5 & 44.7 \\
    \end{tabular}
    \vspace{-1ex}
    \caption{\textbf{Classifications on rewritten responses.} Paraphrasing or translating LLM outputs achieves an accuracy comparable to that using original counterparts. However, summarizing these texts makes the model less capable of predicting LLM identities.}
    \label{tab:rewriting}
\end{table}

We show the results in Table~\ref{tab:rewriting} (see Appendix~\ref{appendix:semantics} for results using alternative LLM for rewriting). The classifiers trained on paraphrased LLM responses maintain similar accuracy levels to those using original responses. Likewise, when using translated text, the classifiers are also able to differentiate between LLMs. These findings suggest that \emph{the semantic meanings of words play a more significant role in predicting LLM origins than the exact word choice.}

Moreover, despite a noticeable accuracy drop (\ie, $>$38\%) with the summarized text, the resulting performance remains well above chance-level guess. This remarkable ability to classify the summarized texts shows the \emph{high-level semantic difference in LLM-generated responses.}

\textbf{Open-ended language analysis.} In this part, we focus on studying the semantic difference in responses generated by LLMs. We employ another LLM (\eg, {\gpt}) as a judge to provide open-ended, descriptive characterizations for each LLM's outputs. The results with other LLM judges for our language analysis are available in Appendix~\ref{appendix:language}. 

Specifically, we present an LLM judge with two responses -- generated by different models based on the same prompt -- and ask it to analyze these responses from different angles (\eg, tone and content). This process is repeated multiple times to gather a comprehensive collection of analyses. Finally, we query the LLM judge to summarize these analyses into bullet points that capture the characteristics of each model. The prompts are detailed in Appendix~\ref{appendix:language_prompt}.

The results of open-ended language analysis on {\gpt} \vs {\claude} are shown in Figure~\ref{fig:open-ended}. For a detailed pairwise comparison of the responses, see Figure~\ref{appendix:fig:open-ended-language-analysis} in Appendix~\ref{appendix:response_demonstration}. {\gpt} is characterized by descriptive and detailed responses in an engaging tone. In contrast, {\claude} prioritizes simplicity with only key points and straightforward language. Additional results on chat API models and instruct LLMs are provided in Appendix~\ref{appendix:language}.

\section{Implications}
In this section, we explore the broader implications of our framework, regarding synthetic data and model similarity.

\textbf{Idiosyncrasies via synthetic data.} 
Using synthetic data has become a common practice when training frontier LLMs~\cite{abdin2024phi3technicalreporthighly,abdin2024phi4,liu2024bestpractices}. We conduct supervised fine-tuning (SFT) on two base LLMs (Llama3.1-8b and Gemma2-9b) using Ultrachat, \ie, dialogues generated by {\gpt}. After the SFT stage, we train a classifier to distinguish between responses from two fine-tuned models. We find that SFT on the same synthetic dataset significantly reduces the classification accuracy from 96.5\% to 59.8\%, narrowing down the differences between these two models. 

In addition, we generate responses from Llama3.1-8B and Gemma2-9B in instruct LLMs using UltraChat prompts. Then we fine-tune Qwen2.5-7B base LLM on each set of responses respectively. Interestingly, responses from the two resulting fine-tuned models can be classified with 98.9\% accuracy, suggesting that each fine-tuned model retains the unique characteristics in its SFT data. These findings suggest that training with synthetic data can propagate the idiosyncrasies in the source model. 

We note that this behavior is not limited to synthetic datasets; in fact, training on different datasets often leads to idiosyncratic patterns in model outputs~\citep{mansour2024bias}.

\textbf{Inferring model similarity.} Our framework offers a quantitative approach for assessing similarities between proprietary and open-weight LLMs. Given a set of $N$ LLMs, we omit one model and train a classifier on responses from the remaining $N-1$ models. We then evaluate which LLM the classifier associates the responses of the excluded model with. The model that is most frequently predicted as the
\begin{figure}[h]
    \centering
    \includegraphics[width=\linewidth]{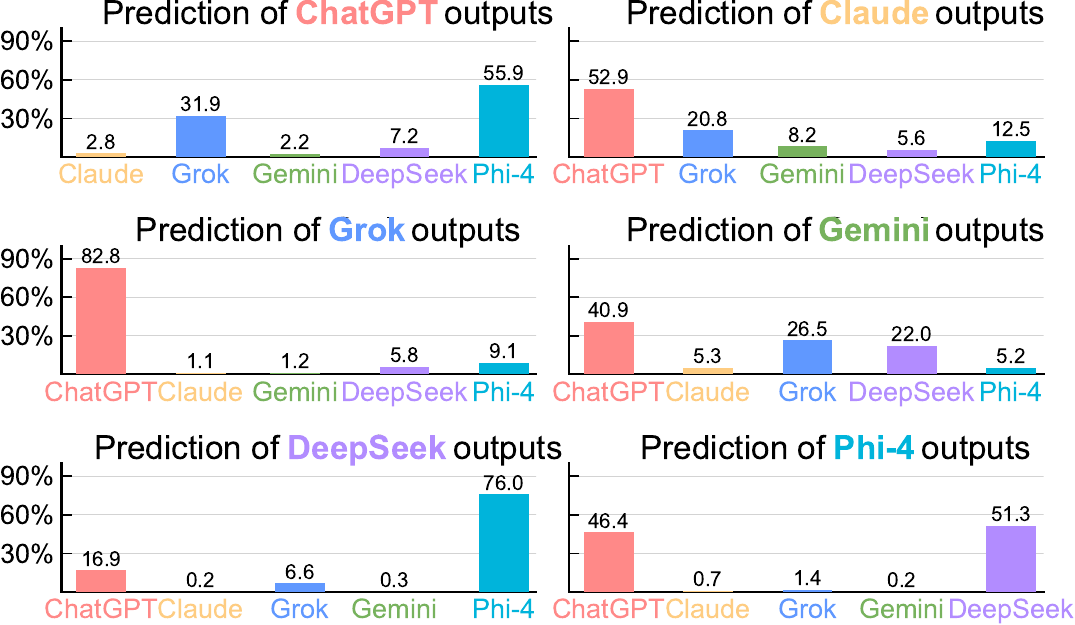}
    \vspace{-2.ex}
    \caption{\textbf{Inferring model similarity.} We consider 6 LLMs, including 5 chat API models and Phi-4. In each subfigure, we evaluate a five-way classifier on outputs from the excluded LLM and present the distribution of predicted model origins. There is a strong tendency for LLM outputs to be predicted as ChatGPT.}
    \label{fig:leave_one_out}
\end{figure}
source is considered the closest match to the excluded LLM. This process is repeated for each of the $N$ models. For this 
analysis, we include the open-weight Phi-4~\citep{abdin2024phi4} alongside 5 chat API models. Note that Phi-4 uses a substantial amount of synthetic data in its training.

Figure~\ref{fig:leave_one_out} shows the results. Intriguingly, for {\claude}, {\grok}, and {\gemini}, we observe a strong tendency for their outputs to be classified as {\gpt}. For instance, when {\grok} is the excluded model, 82.8\% of its responses are classified as {\gpt}. In addition, responses from {\gpt} and {\deepseek} are frequently identified as coming from Phi-4, with 55.9\% and 76.0\% of their responses, respectively. In turn, most of Phi-4's outputs are classified as originating from {\gpt} or {\deepseek}.

\textbf{Robust evaluation of LLMs.} Our findings reveal a potential vulnerability in widely used LLM evaluation methodologies. It has become a common strategy to incorporate human judgement in evaluating  LLMs, for instance, Chatbot Arena~\citep{Chiang2024ChatbotArena}. It is a voting-based leaderboard where users submit preferences of the responses from two randomly chosen models. This benchmark has gained significant traction and is now a key reference point for frontier model development. However, exploiting the idiosyncratic property of LLM outputs, a malicious attacker can identify the model behind the candidate responses and consistently vote for the target model, thereby manipulating the leaderboard rankings. Concurrent work by \citet{Huang2025Exploring} has demonstrated the feasibility of this attack in simulation. We hope our work brings attention to potential weaknesses in current evaluation pipelines, as they can misguide model development and optimization efforts~\citep{Singh2025Leaderboard}.

\section{Related Work}

\textbf{Dataset classification.} \citet{torralba2011unbiased} introduced the ``Name That Dataset” experiment a decade ago to highlight the bias present in visual datasets at that time. Recently, \citet{liu2024datasetbias} revisited this problem (termed dataset classification) and found that current large-scale, supposedly more diverse visual datasets are still very biased. \citet{zeng2024bias} further identified structural and semantic components in images as key contributors to these biases. \citet{you2024images} and \citet{mansour2024bias} applied the dataset classification framework to study bias in synthetic images and LLM pretraining datasets, respectively. While the synthetic task shown in  Figure~\ref{fig:teaser} is conceptually similar to dataset classification, we focus not on training datasets but on the distinctive characteristics inherent to LLMs. Shortly after our initial arXiv preprint, \citet{suzuki2025natural} demonstrated that subtle variations in the training process can result in distinguishable outputs across LLMs.

\textbf{Human \vs machine-generated texts.} Many prior works have studied the problem of determining whether a text is authored by a human or an AI system~\citep{mitchell2023detectgpt,Wu2023Survey,cai2025areyougetting}. Model-free approaches typically use linguistic properties such as n-gram frequencies~\citep{badaskar-etal-2008-identifying,openai_gpt2_dataset}, entropy~\citep{thomas2008entropy,gehrmann2019gltr} or negative probability curvature~\citep{mitchell2023detectgpt,bao2023fastdetectgpt}. Other works leverage neural network features to perform this task, such as fine-tuning BERT~\citep{uchendu2021turingbench,ippolito2019automatic}. \citet{Russell2025People} found that experienced chatbot users are good at distinguishing between AI- and human-written articles. Neural authorship attribution~\citep{uchendu-etal-2020-authorship,Antoun2023FromTextToSource,huang2024authorship} seeks not only to identify machine-generated text but also to attribute it to specific text generators.  In this work, we focus on the distinguishability between LLMs rather than between AI \vs human. 

\textbf{Characteristics of machine-generated texts.} Beyond  detection, it is also important to understand the distinctive properties of machine-generated texts. N-gram frequencies have long served as a basic signal for such purpose. \citet{mcgovern2024fingerprints} showed that LLM outputs contain unique lexical and syntactic fingerprints that distinguish them from human writing. Other works have examined stylometric features in AI-generated content~\citep{zaitsu2023distinguishing}. Most recently, \citet{chakrabarty2024can} studied the human-AI alignment in the writing process, finding that professional writers can effectively identify and edit undesirable idiosyncrasies common in LLM-generated text. Our work seeks to provide a deeper and principled understanding behind the observed idiosyncrasies from our framework in Figure~\ref{fig:teaser}.

\textbf{Understanding differences between distributions.} A line of research~\citep{dunlap2023describing, zhong2024explaining} has used foundation models to describe qualitative differences between pairs of data distributions (\eg, image datasets). \citet{gao2024model} and \citet{cai2025areyougetting} conducted auditing studies on LLM APIs to detect instances of model substitution, such as watermarking and quantization.
The most relevant work to us is \citet{dunlap2024vibecheck}, which proposed VibeCheck to understand user-aligned traits in LLM outputs. They found that LLMs often vary in styles, such as being more formal or friendly. In contrast, our work aims to identify broader generalizable patterns to interpret the high classification performance.

\section{Conclusion}
We demonstrate the presence of idiosyncrasies in Large Language Models (LLMs) and investigate a synthetic task designed to quantify their extent. We find that simply fine-tuning pretrained text embedding models on LLM outputs leads to exceedingly high accuracy in predicting the origins of the text. This phenomenon persists across diverse prompt datasets, LLM combinations, and many other settings. We also pinpoint concrete forms of these idiosyncrasies within LLMs. We hope our work encourages further research into understanding idiosyncrasies in LLMs.

We conclude by outlining several directions for future work:
\begin{itemize}[leftmargin=3.0ex, topsep=0ex, itemsep=1.0ex]
    \item It is worth investigating whether our observations generalize to LLMs beyond the Transformer architecture, such as state space models~\citep{gu2023mamba} and diffusion-based language models~\citep{nie2025largelanguagediffusionmodels}.
    \item Understanding how the training process results in these idiosyncrasies remains an important open question.
    \item Third, our setup does not consider scenarios where the list of source LLMs could be large and even unknown beforehand, which require further investigation. Our initial results suggest it is promising: a 10-way classification problems involving 5 chat APIs and 5 instruct LLMs including Phi-4 achieves 92.2\% accuracy.
    \item It is interesting to examine how these idiosyncrasies relate to model distillation~\citep{hinton2015distilling}, a technique that has become increasingly prevalent in practice.
\end{itemize}

\section*{Acknowledgments} 
We thank Zekai Wang for valuable discussions. Mingjie Sun was supported by funding from the Bosch Center for Artificial Intelligence. 

\section*{Impact Statement}
Our study investigates the distinguishability of LLMs. On the positive side, our results offer insights into LLM behaviors that enhance transparency, accountability and provenance in tracking AI generated content. This has valuable applications such as combating misinformation, and ensuring responsible use of LLMs. However, our results highlight challenges in model training. Reliance on synthetic data or model distillation risks propagating biases and potentially infringing on the intellectual property of proprietary models.

\bibliography{refs}
\bibliographystyle{icml2025}

\newpage
\clearpage

\appendix
\onecolumn

\section*{Appendix}
\section{Implementation Details}\label{appendix:implementation}
\subsection{Response Generation}\label{appendix:response_gen}
We report our procedure for generating responses from chat APIs, instruct LLMs, and base LLMs. For chat APIs, we access a stable version of each model, including GPT-4o-2024-08-06, Claude-3.5-Sonnet-20241022, Grok-Beta, Gemini-1.5-Pro-002, and DeepSeek-Chat, through its official API between November 28, 2024, and February 6, 2025, generating responses with their default sampling setting. For instruct LLMs, we use greedy decoding to sample outputs. For base LLMs, we set the temperature to $T=$ 0.6 and apply a repetition penalty of 1.1 to avoid repetitive completions.

\subsection{Training Setup}
\label{appendix:finetune}
In this part, we describe our fine-tuning process using the text embedding models on LLM responses. We use the first 512 tokens of each generated response for training and evaluation. To perform sequence classification, we add a linear layer as the classification head on top of each text embedding model. For ELMo, BERT, LLM2vec, this layer is applied to the average embeddings over all tokens in a sequence. For T5 and GPT-2, we follow the original setups~\cite{radford2019gpt2, raffel2023T5} and apply the head on the output of the last token. 

For smaller text embedding models, such as ELMo, BERT, T5, and GPT-2, we fine-tune the entire model along with the classification head, searching over base learning rates $\{$3e-3, 1e-3, 3e-4, 1e-4, 3e-5, 1e-5, 3e-6, 1e-6$\}$. For the largest LLM2vec model, we employ the parameter-efficient LoRA~\cite{hu2022lora} fine-tuning method with a rank of 16, LoRA $\alpha$ of 32, a dropout rate of 0.05, and a base learning rate of 5e-5. Table~\ref{tab:recipe} details our basic training recipe.

\begin{table}[ht]
\centering
\small
\addtolength{\tabcolsep}{-.3pt}
\def\arraystretch{1.2}
\begin{tabular}{y{108}|x{96}}
config & value \\
\Xhline{0.7pt}
optimizer & AdamW \\
weight decay & 0.001 \\
optimizer momentum & $\beta_1, \beta_2=0.9, 0.999$ \\
training epochs & 3 \\
batch size & 8 \\
learning rate schedule & cosine decay \\
warmup schedule & linear \\
warmup ratio & 10\% \\
gradient clip & 0.3 \\
\end{tabular}
\caption{Our fine-tuning recipe.}
\label{tab:recipe}
\end{table}

\clearpage 
\newpage 
\subsection{Prompts for Open-ended Language Analysis}
\label{appendix:language_prompt}

\begin{figure}[ht]
    \centering
    \begin{subfigure}[h]{0.45\linewidth}
    \includegraphics[width=\linewidth]{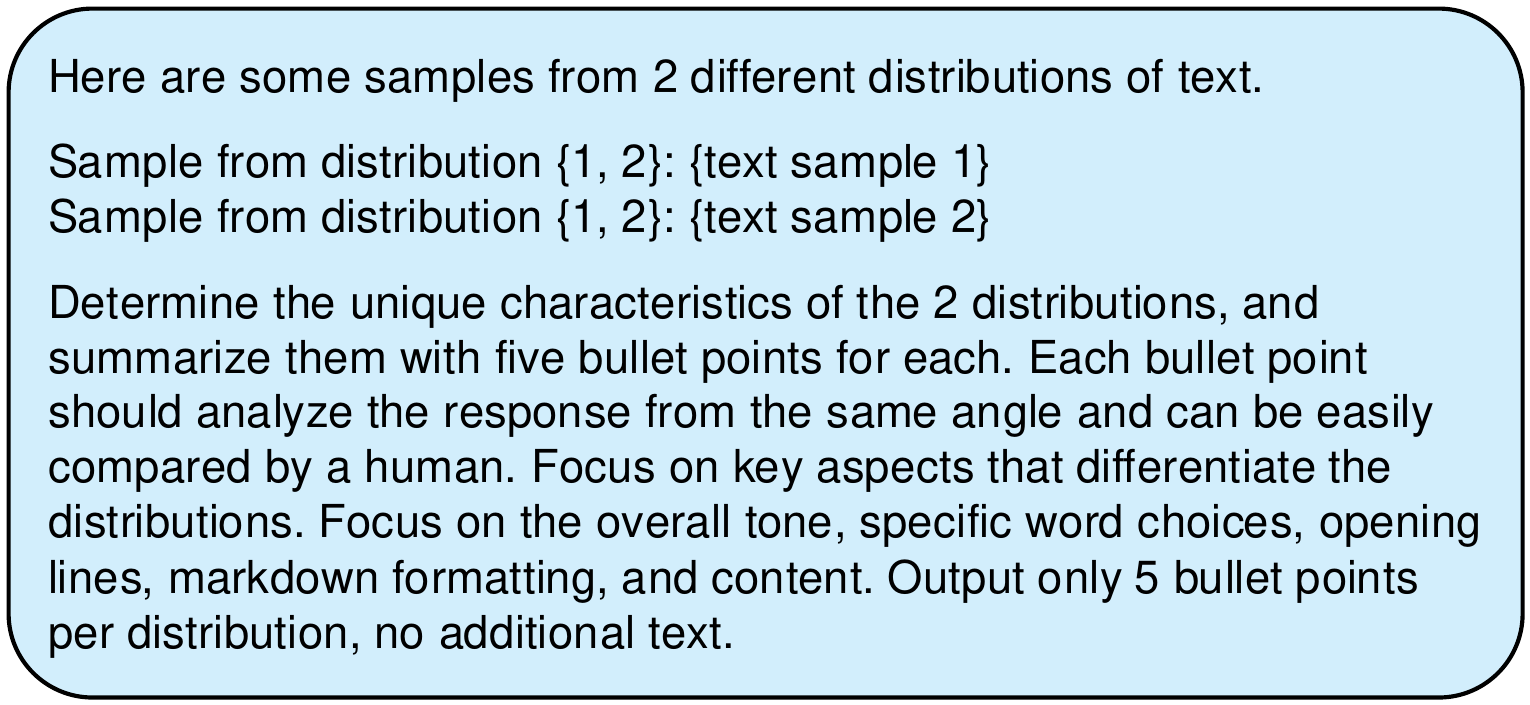}
    \subcaption{analysis prompt}
        \vspace{2ex}
    \label{fig:analysis_prompt}
    \end{subfigure}
    \hspace{2ex}
    \begin{subfigure}[h]{0.45\linewidth}
    \includegraphics[width=\linewidth]{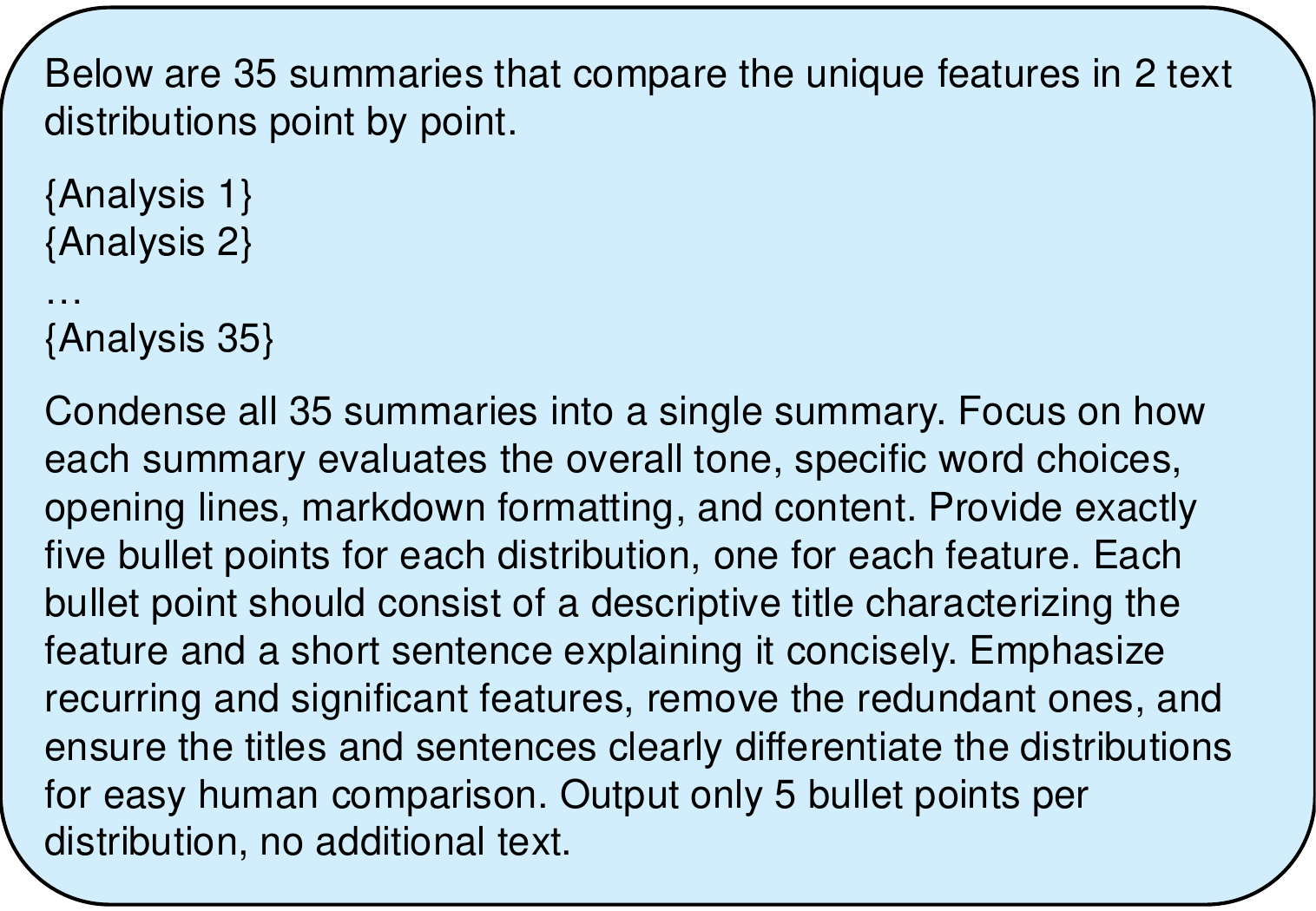}
    \subcaption{summarization prompt}
    \label{fig:summary_prompt}
    \end{subfigure}
    \caption{Prompts in our open-ended language analysis.}
    \label{fig:prompt_open_ended_language_analysis}
\end{figure}

We detail the procedures of our open-ended language analysis in Section~\ref{sec:semantics}. Given the same input, we sample a pair of responses from two LLMs and present them, along with an analysis prompt (see Figure~\ref{fig:analysis_prompt}), to an LLM judge for comparison. To avoid the LLM judge exploiting any prior knowledge of the models, we anonymize model identities using an index distribution. This process is repeated for 35 response pairs, yielding a set of detailed analyses. Finally, we use the summarization prompt (see Figure~\ref{fig:summary_prompt}) to distill these analyses into 5 bullet points that characterize the idiosyncrasies of each model.

\section{Additional Results}
\label{appendix:additional_results}

\subsection{Confusion Matrix}\label{appendix:section_confusion_matrix}
\begin{figure}[H]
    \centering
    \begin{subfigure}[h]{.32\linewidth}
        \centering
        \includegraphics[width=\linewidth]{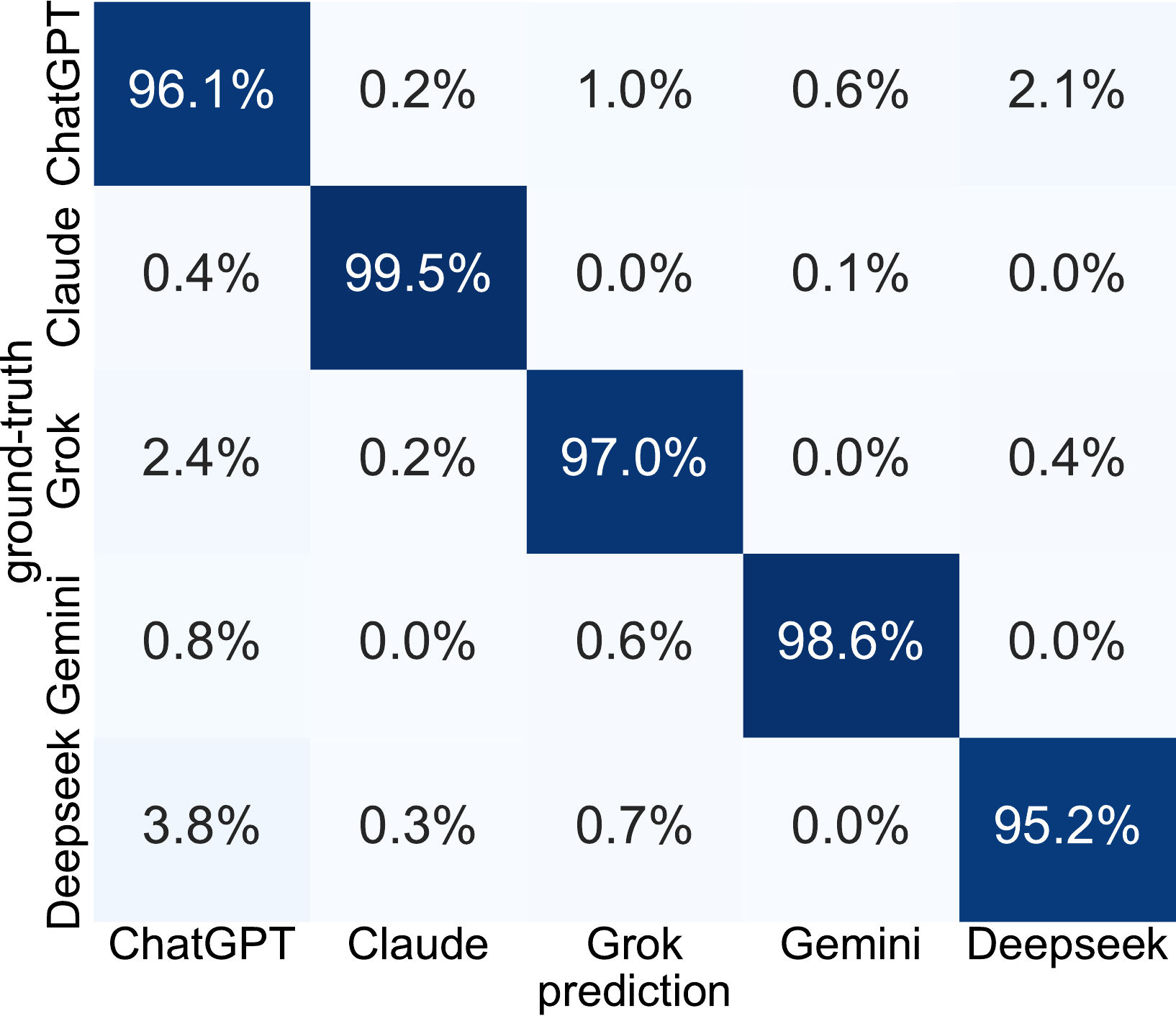}
        \vspace{-3ex}
        \subcaption{chat APIs}
    \end{subfigure}~
    \begin{subfigure}[h]{.32\linewidth}
        \centering
        \includegraphics[width=\linewidth]{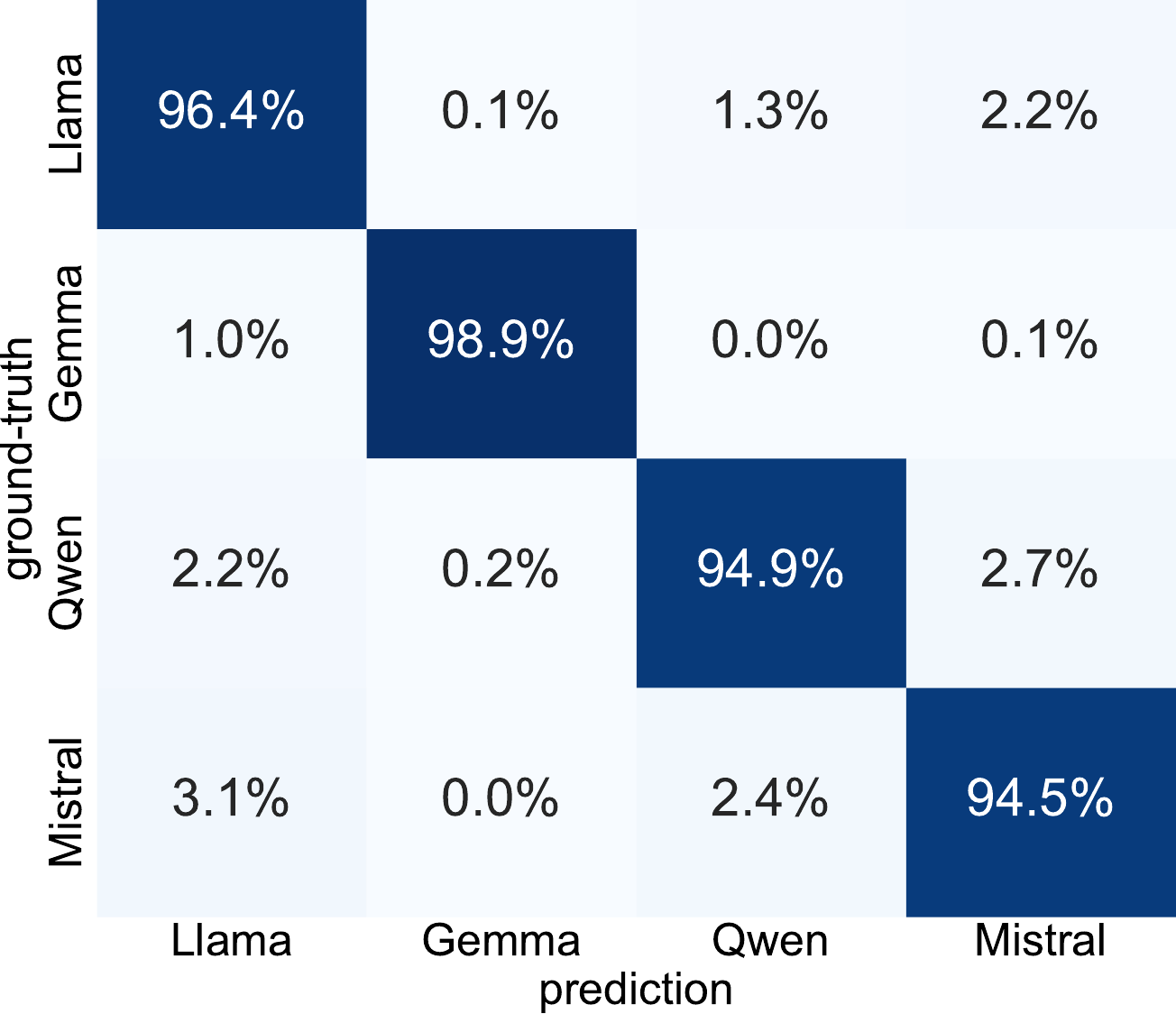}
        \vspace{-3ex}
        \subcaption{instruct LLMs}
    \end{subfigure}~
    \begin{subfigure}[h]{.32\linewidth}
        \centering
        \includegraphics[width=\linewidth]{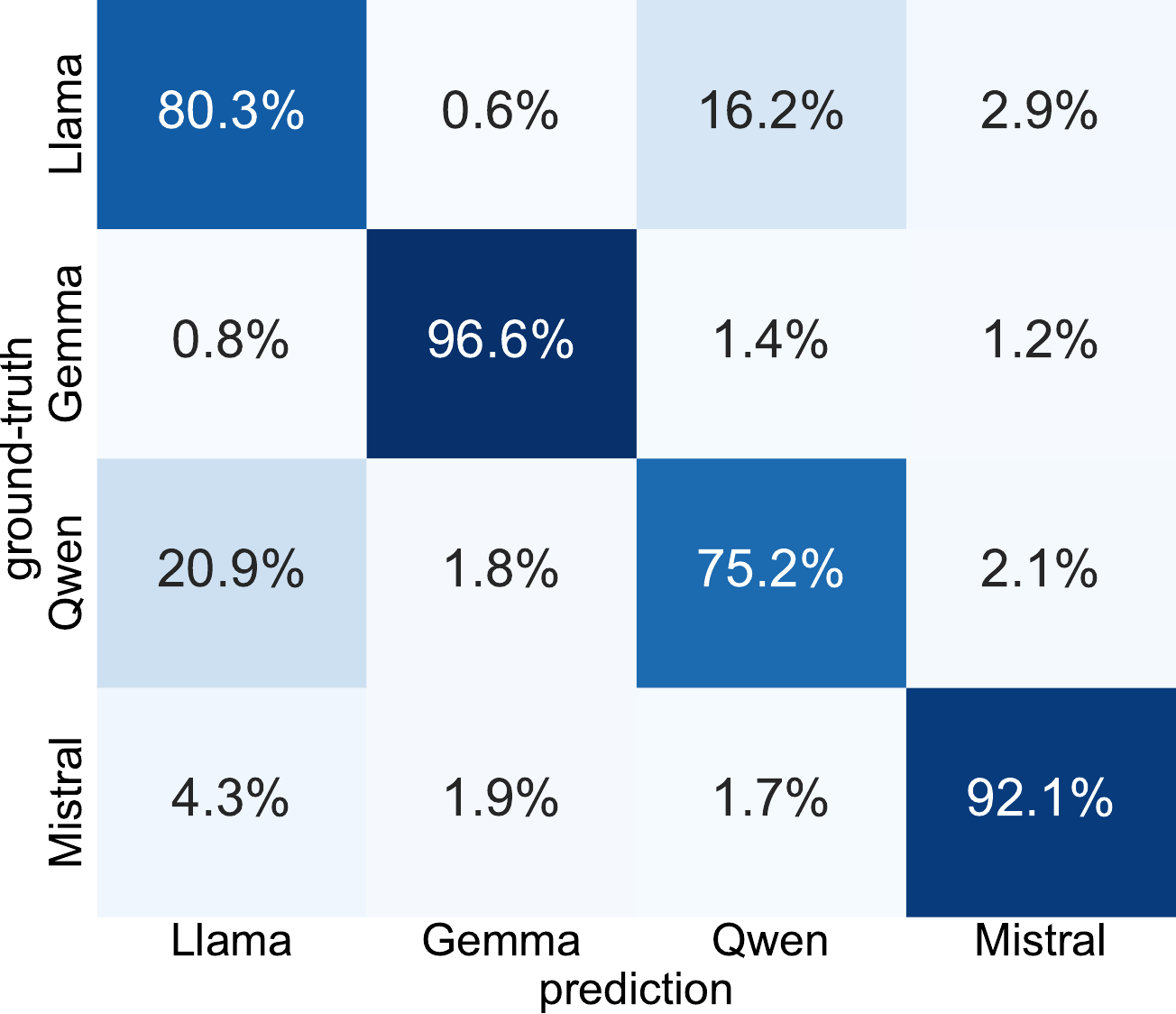}
        \vspace{-3ex}
        \subcaption{base LLMs}
    \end{subfigure}
    \caption{Confusion matrices for $N$-way classifiers on three groups of LLMs: chat APIs, instruct LLMs, and base LLMs.}
    \label{fig:confusion}
    \vspace{-1em}
\end{figure}

In Figure~\ref{fig:confusion}, we present the confusion matrix for the $N$-way classifiers that are trained on responses generated by chat APIs, instruct LLMs, and base LLMs, respectively. The results demonstrate that our classifiers can accurately predict the origin of LLM-generated responses, with minimal confusion between different LLMs.

\label{appendix:confusion_matrix}

\subsection{Words and Letters}
\label{appendix:char_distribution}
\begin{figure}[H]
    \centering
    \begin{subfigure}[h]{0.49\linewidth}
    \centering
    \includegraphics[width=0.49\linewidth]{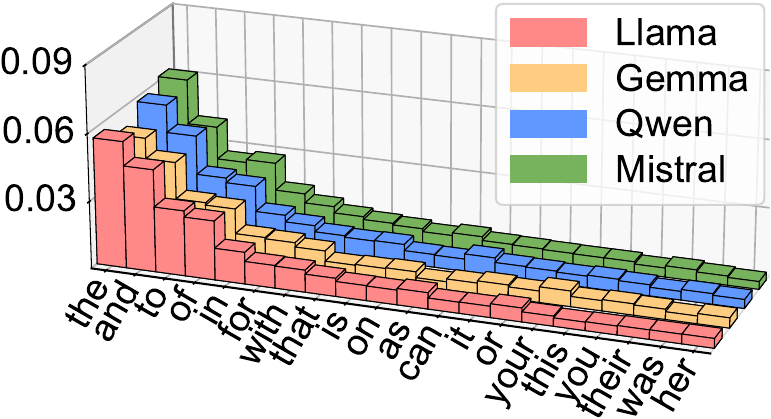}
    \includegraphics[width=0.49\linewidth]{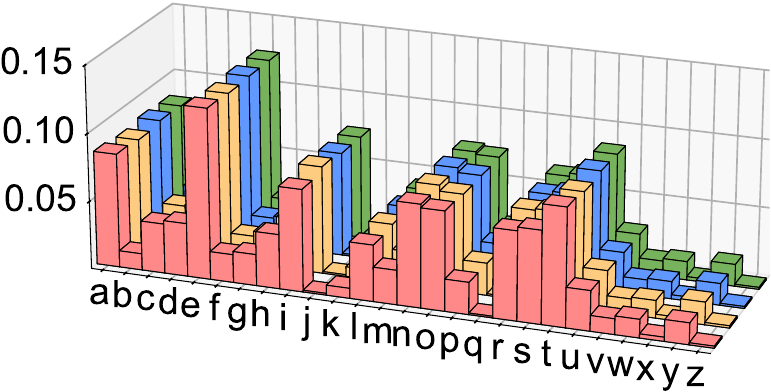}
    \subcaption{instruct LLMs}
    \label{fig:top_word_instruct}
    \end{subfigure}
    \begin{subfigure}[h]{0.49\linewidth}
    \centering
    \includegraphics[width=0.49\linewidth]{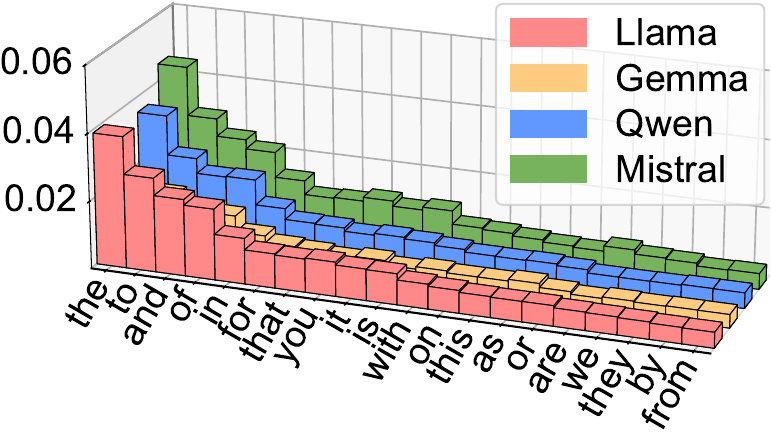}
    \includegraphics[width=0.49\linewidth]{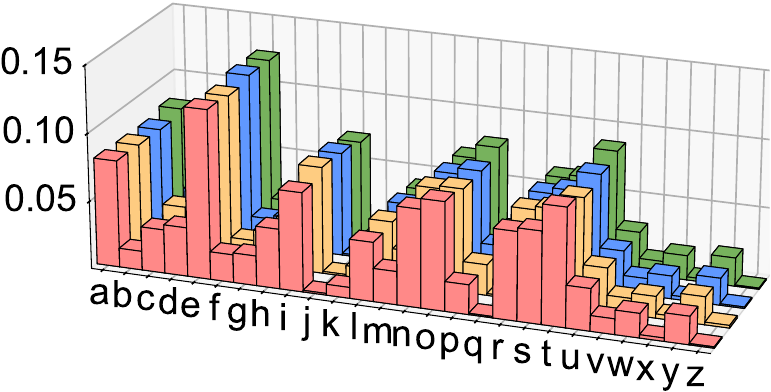}
    \subcaption{base LLMs}
    \label{fig:top_word_base}
    \end{subfigure}
    \caption{Word and letter frequencies in instruct and base LLMs.}
    \label{fig:words_and_letters_instruct_base}
\end{figure}

Figure~\ref{fig:words_and_letters_instruct_base} presents the frequencies of the 20 most commonly used words (\textit{left}) and all English letters (\textit{right}) across instruct and base LLMs. Consistent with our observations in Section~\ref{sec:words_letters}, we find notable differences in the distribution of commonly used words between these models, such as ``the'', ``and'', ``to''. In contrast, the letter distributions are nearly identical.

\subsection{Characteristic Phrases}
\label{appendix:top_phrases}
We provide additional results for characteristic phrases as presented in Section~\ref{sec:words_letters}. We follow the same methodology in Figure~\ref{fig:top_words} to extract characteristic phrases of instruct and base LLMs. Specifically, we train a four-way logistic regression classifier on the TF-IDF features of their responses and use the coefficients to select important phrases of each model. 

\begin{figure}[ht]
    \centering
    \begin{subfigure}[h]{.49\linewidth}
        \centering
        \includegraphics[width=\linewidth]{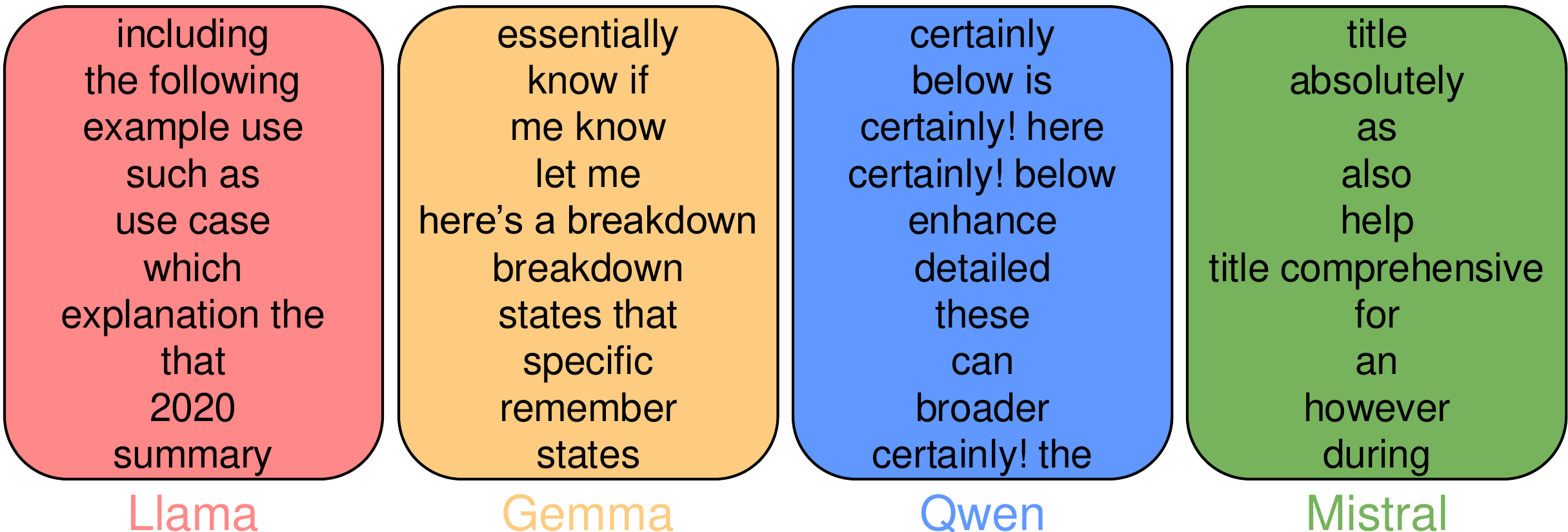}
        \subcaption{instruct LLMs}
        \vspace{-1ex}
    \end{subfigure}
    \begin{subfigure}[h]{.49\linewidth}
        \centering
        \includegraphics[width=\linewidth]{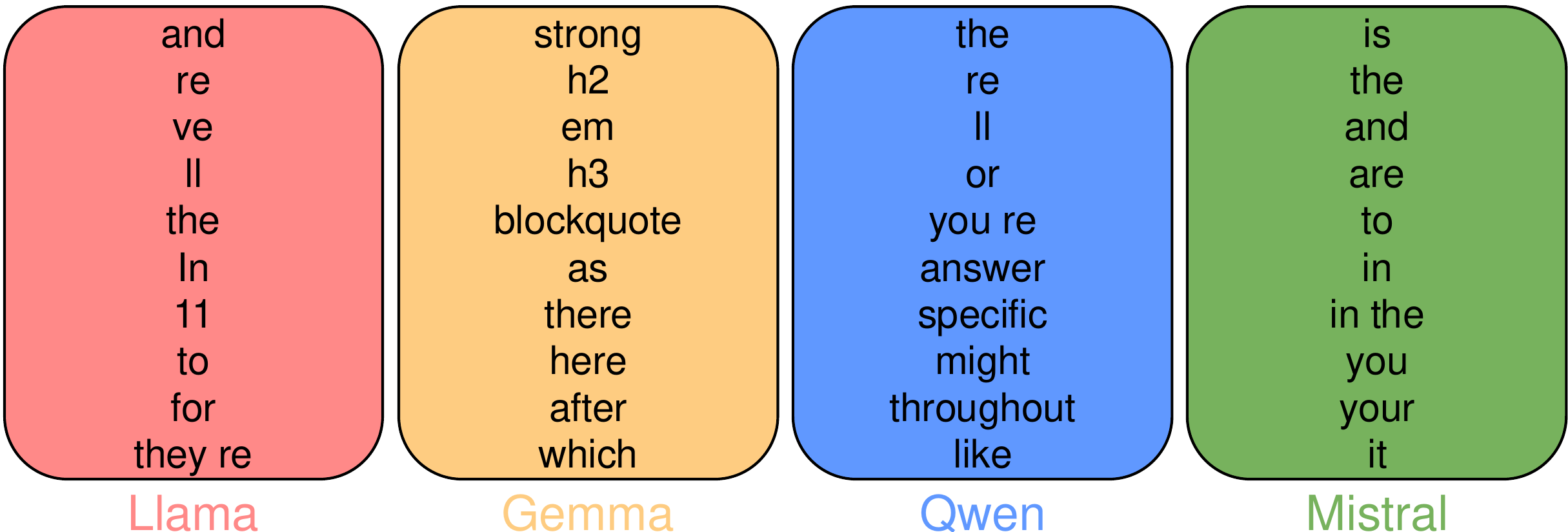}
        \subcaption{base LLMs}
        \vspace{-1ex}
    \end{subfigure}
    \centering
    \caption{Characteristic phrases for instruct\protect\footnotemark and base LLMs.}
    \label{fig:char_phrases_instruct_base}
\end{figure}

As shown in Figure~\ref{fig:char_phrases_instruct_base}, each instruct LLM contains quite distinct characteristic phrases. For example, Llama frequently employs terms ``including'' and ``such as'' to introduce specific examples in the output, whereas Gemma tends to engage with users using phrases ``let me'' and ``know if''. In contrast, the extracted phrases from base LLMs are less distinctive, primarily consisting of common words such as “the”, “to”, and “you”. 

\begin{figure*}[h]
\vspace{-.25em}
    \centering
    \begin{subfigure}[h]{.495\linewidth}
        \centering
        \includegraphics[width=\linewidth]{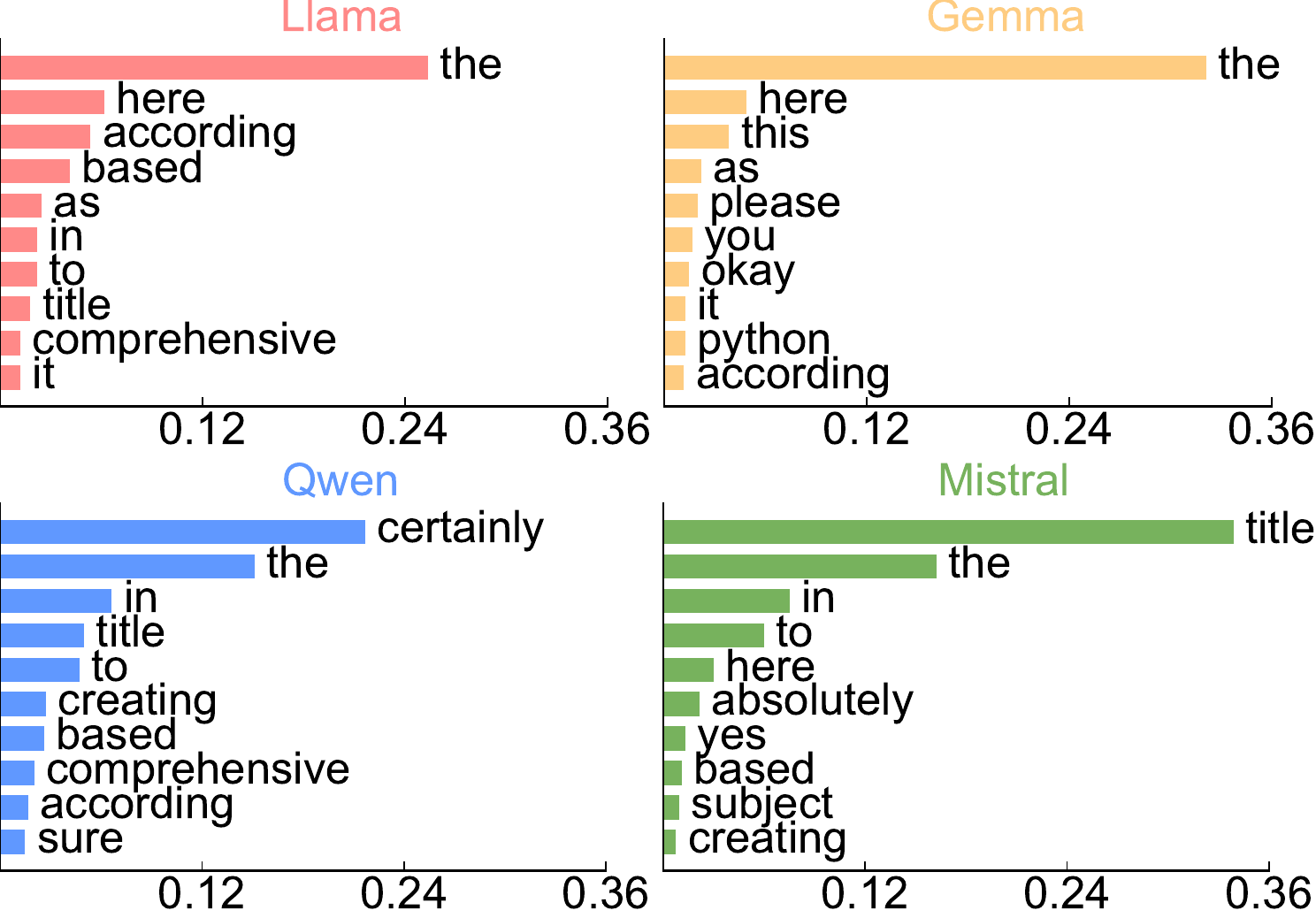}
        \subcaption{instruct LLMs}
    \end{subfigure}
    \begin{subfigure}[h]{.495\linewidth}
        \centering
        \includegraphics[width=\linewidth]{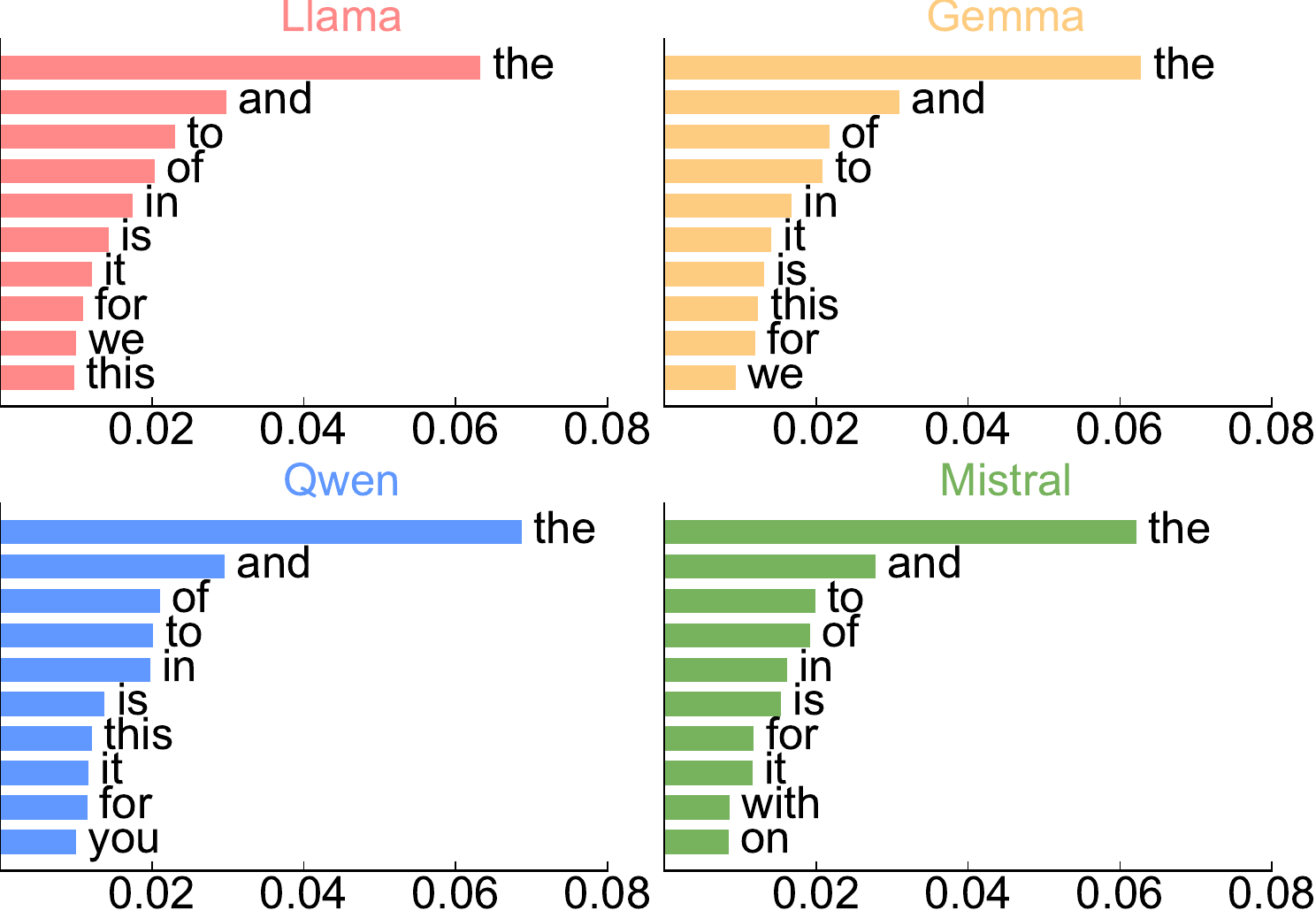}
        \subcaption{base LLMs}
    \end{subfigure}
    \centering
    \vspace{-1ex}
    \caption{Distribution of first word choices in instruct and base LLMs.}
    \label{fig:first_words_instruct_base}
\end{figure*}

Figure~\ref{fig:first_words_instruct_base} illustrates the distribution of first word choices in instruct and base LLMs. Similar to chat APIs (Figure~\ref{fig:first_words_chat}), instruct LLMs display varied distributions. However, base LLMs exhibit substantial overlap in their most frequent first words, \eg, ``the'', ``and'', ``of'', ``to'', and ``in''.

\footnotetext{In Llama of instruct LLMs, the phrase ``explanation the''  corresponds to a markdown header or bold text for ``explanation'' followed by a new sentence starting with ``the''.}

\clearpage
\newpage
\subsection{Unique Markdown Formatting}
\label{appendix:markdown}

\begin{figure*}[th]
    \centering
    \begin{subfigure}[h]{\linewidth}
        
    \includegraphics[width=0.33\linewidth]{figs/markdown_features/chat_bold_text_count-cropped.pdf}~
    \includegraphics[width=0.33\linewidth]{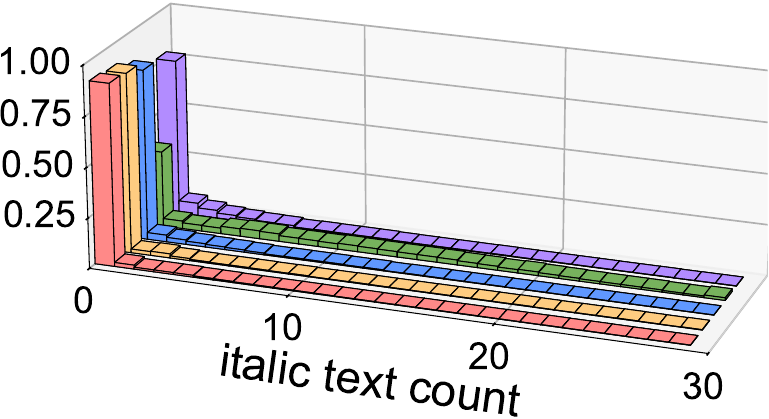}~
    \includegraphics[width=0.33\linewidth]{figs/markdown_features/chat_header_count-cropped.pdf}
    
    \includegraphics[width=0.33\linewidth]{figs/markdown_features/chat_enumeration_count-cropped.pdf}~
    \includegraphics[width=0.33\linewidth]{figs/markdown_features/chat_bullet_point_count-cropped.pdf}~
    \includegraphics[width=0.33\linewidth]{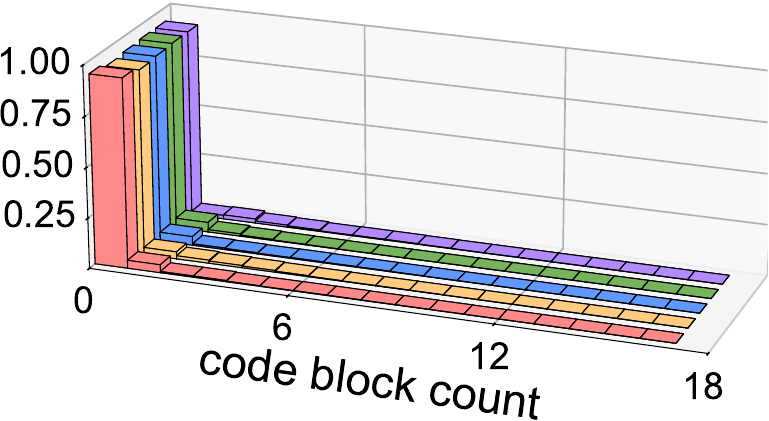}
    \subcaption{chat APIs}
    \vspace{2.5ex}
    \label{appendix:markdown_chatapis}
    \end{subfigure}
    
    \begin{subfigure}[h]{\linewidth}
        
    \includegraphics[width=0.33\linewidth]{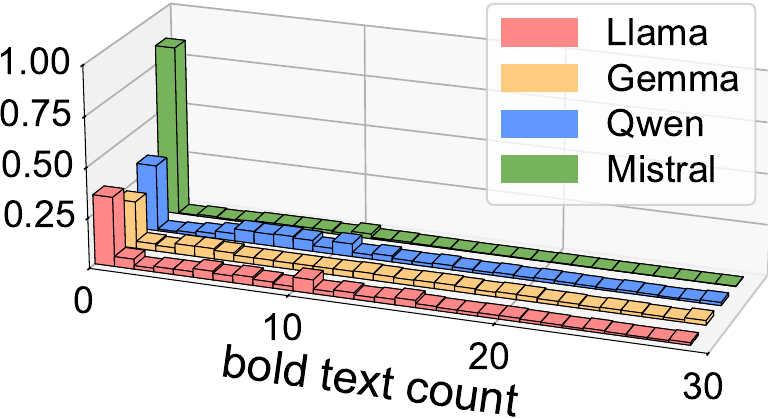}~
    \includegraphics[width=0.33\linewidth]{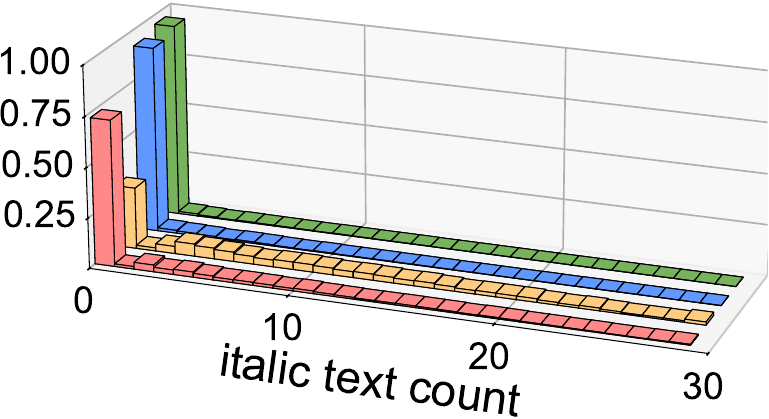}~
    \includegraphics[width=0.33\linewidth]{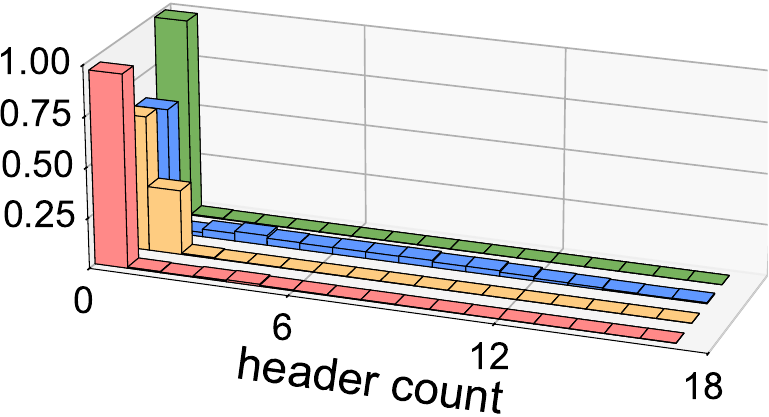}
    
    \includegraphics[width=0.33\linewidth]{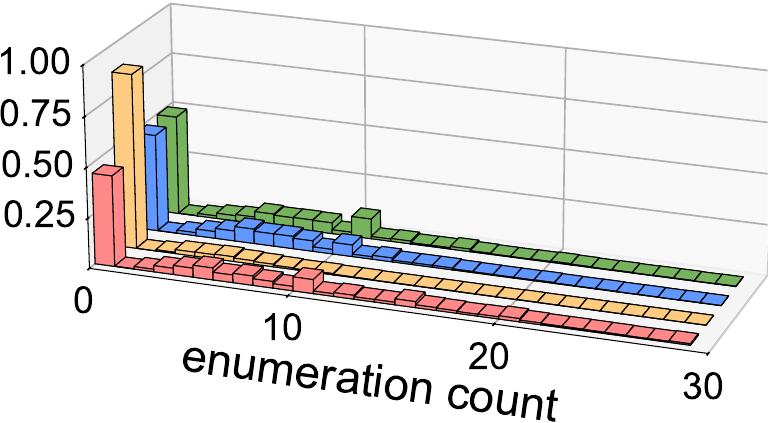}~
    \includegraphics[width=0.33\linewidth]{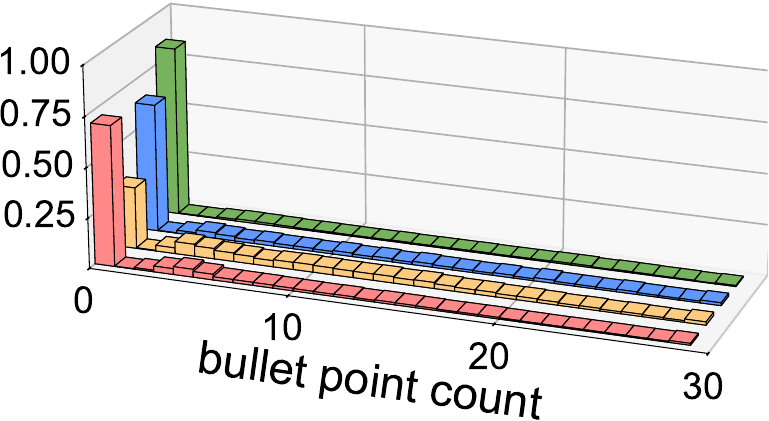}~
    \includegraphics[width=0.33\linewidth]{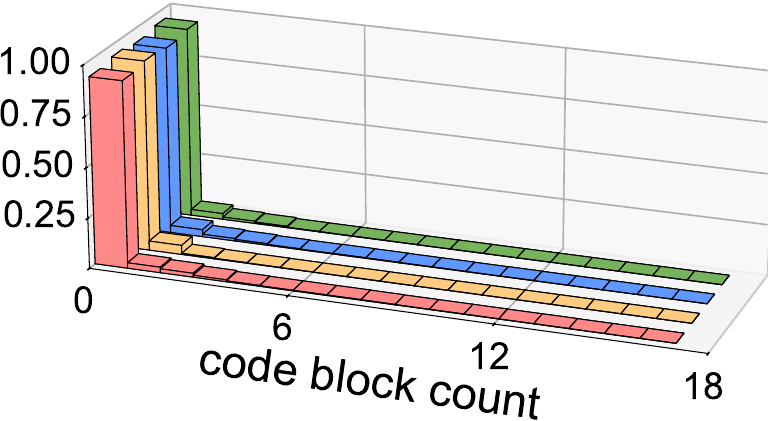}
    \subcaption{instruct LLMs}
    \label{appendix:markdown_instructllms}
    \end{subfigure}
    \caption{Markdown formatting elements for chat APIs (\emph{top}) and instruct LLMs (\emph{bottom}).}
    \label{fig:features_dist_instruct_base}
\end{figure*}

In this part, we provide additional results for the analysis of markdown formatting as presented in Section~\ref{sec:markdown}. Figure~\ref{fig:features_dist_instruct_base} illustrates the distribution counts of six markdown formatting elements across different models. For both chat API models (Figure~\ref{appendix:markdown_chatapis}) and instruct LLMs (Figure~\ref{appendix:markdown_instructllms}), we observe distinct differences in the usage of bold texts, headers, enumerations, and bullet points, while italic texts show less variation. Intriguingly, {\gemini} uses much more italic texts (a lower density at zero in the italic text) than other chat APIs, where similar observations can be found on Gemma2.

\clearpage
\newpage

\subsection{Rewriting LLM outputs}\label{appendix:semantics}
In Section~\ref{sec:semantics}, we used GPT-4o-mini to rewrite LLM outputs. Here, we present results using an alternative model: Qwen2.5-7B-Instruct. As shown in Table~\ref{appendix:tab:rewriting}, our observations remain consistent, indicating that our findings are robust to the choice of LLM used for rewriting.

\begin{table}[ht]
    \centering
    \tablestyle{6pt}{1.15}
    \begin{tabular}{lcccc}
   LLM for rewriting & original & paraphrase & translate & summarize\\
    \shline
    GPT-4o-mini  &  97.8 & 93.6 & 93.9 & 63.7 \\
    Qwen2.5-7B-Instruct & 97.8 & 92.6 & 94.3 & 71.5 
    \end{tabular}
    \vspace{-1ex}
    \caption{\textbf{Classification accuracies on rewritten responses.} The results are on Chat API responses. }
    \label{appendix:tab:rewriting}
    \vspace{-1ex}
\end{table}

\subsection{Open-ended Language Analysis}
\label{appendix:language}

\begin{figure*}[th]
    \centering
    \begin{subfigure}[h]{\linewidth}
    \includegraphics[width=.49\linewidth]{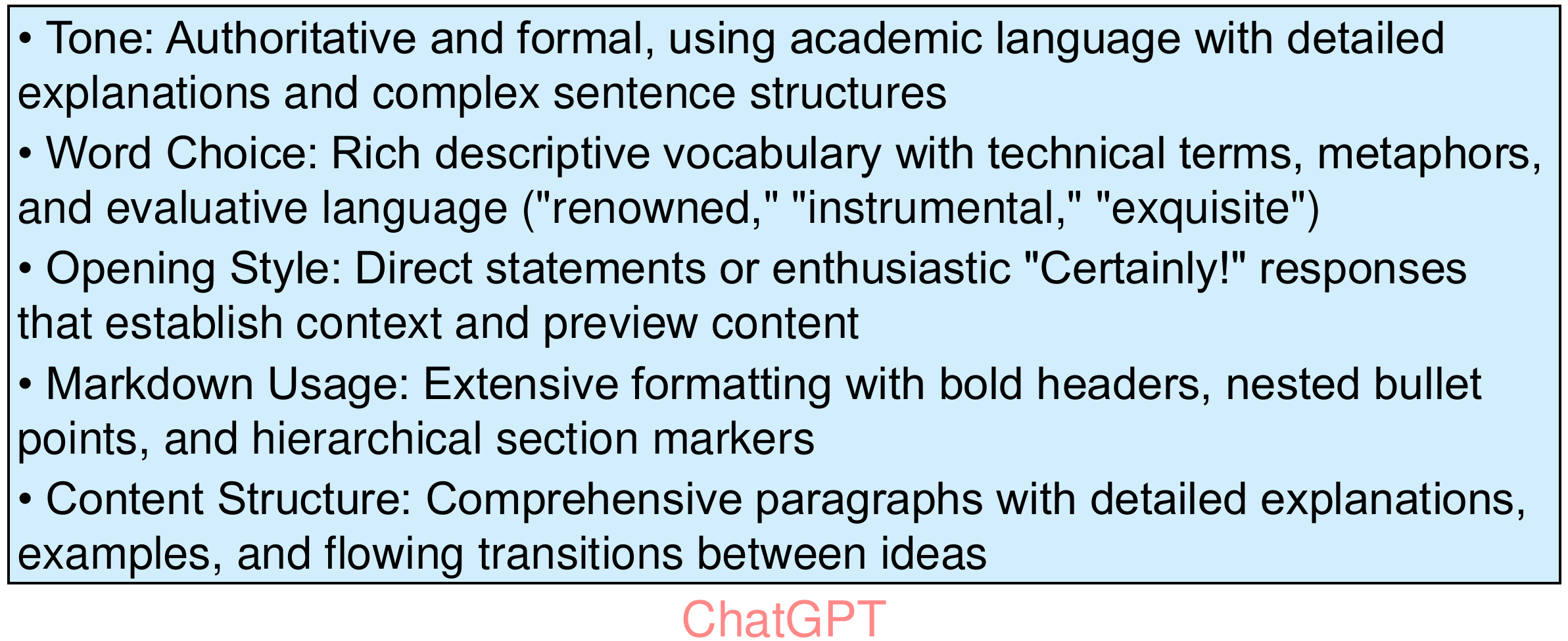}~
    \includegraphics[width=.49\linewidth]{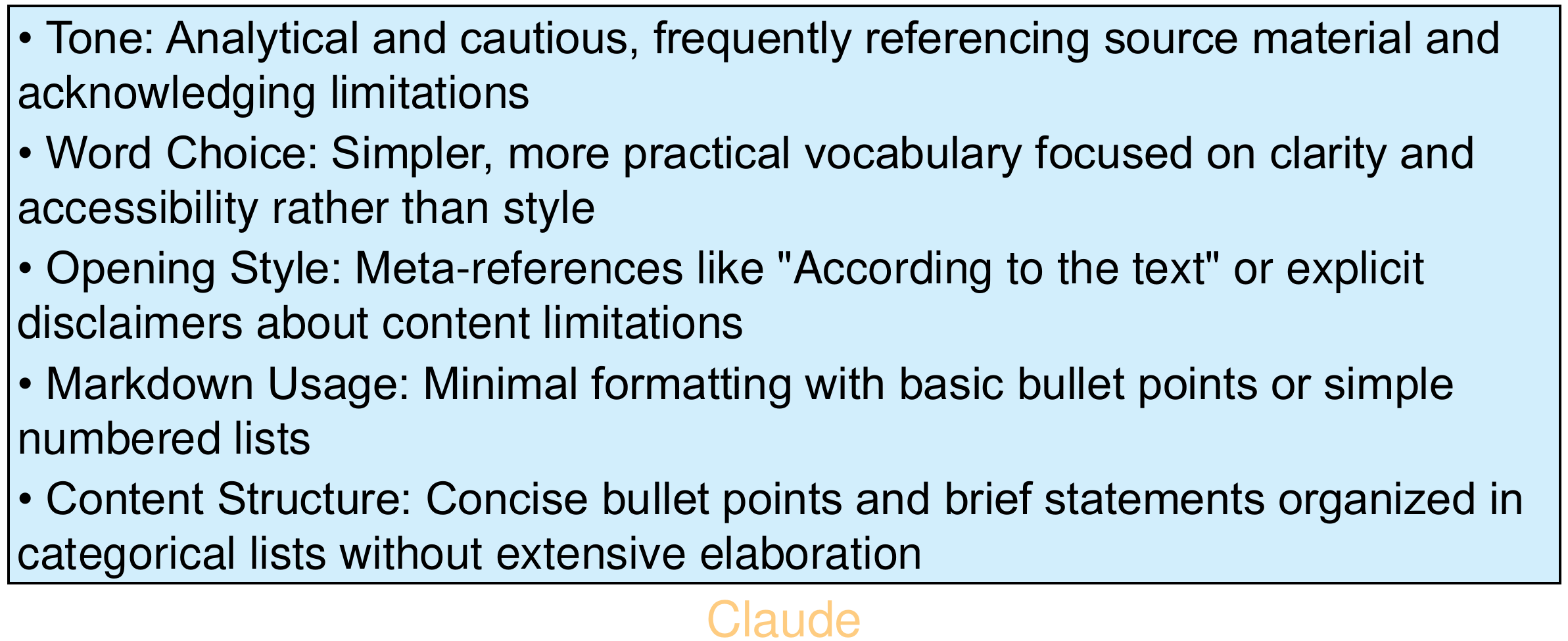}
    \vspace{-1ex}
    \subcaption{{\claude} as the LLM judge.}
    \vspace{2.5ex}
    \end{subfigure}
    
    \begin{subfigure}[h]{\linewidth}
    \includegraphics[width=.49\linewidth]{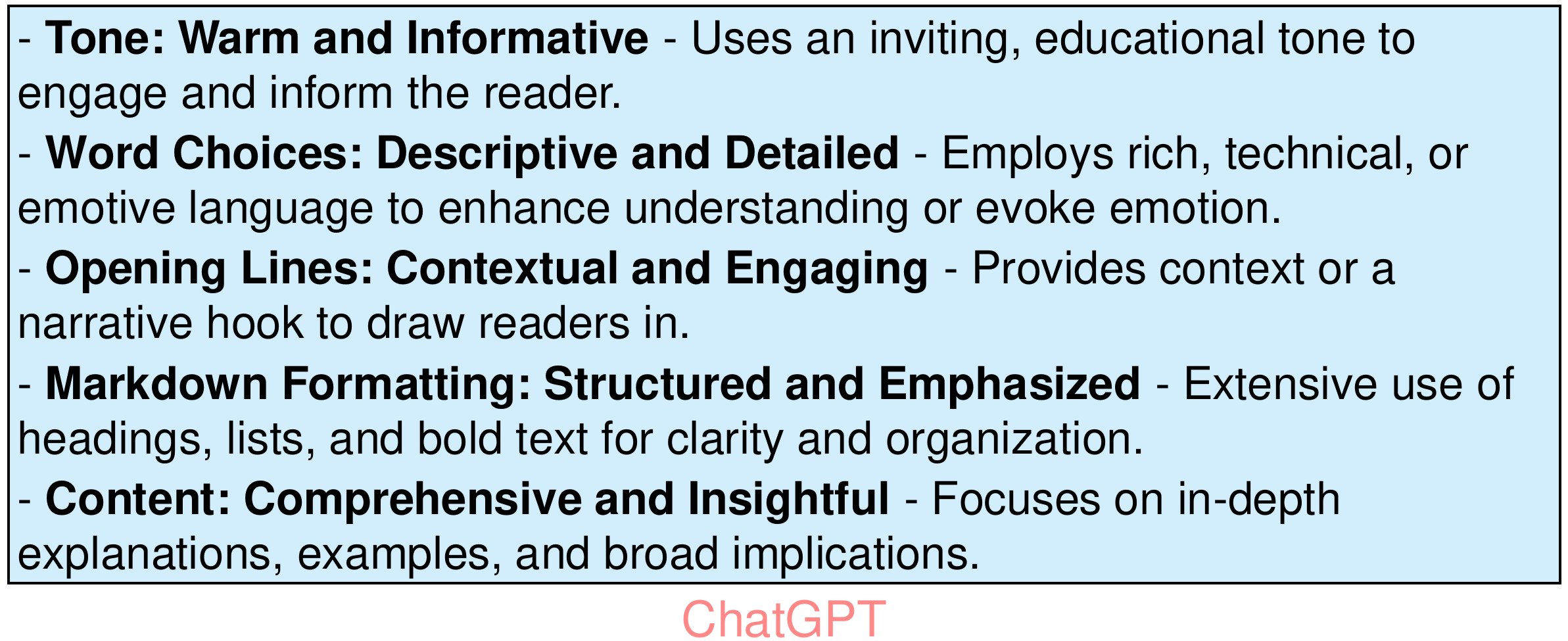}~
    \includegraphics[width=.49\linewidth]{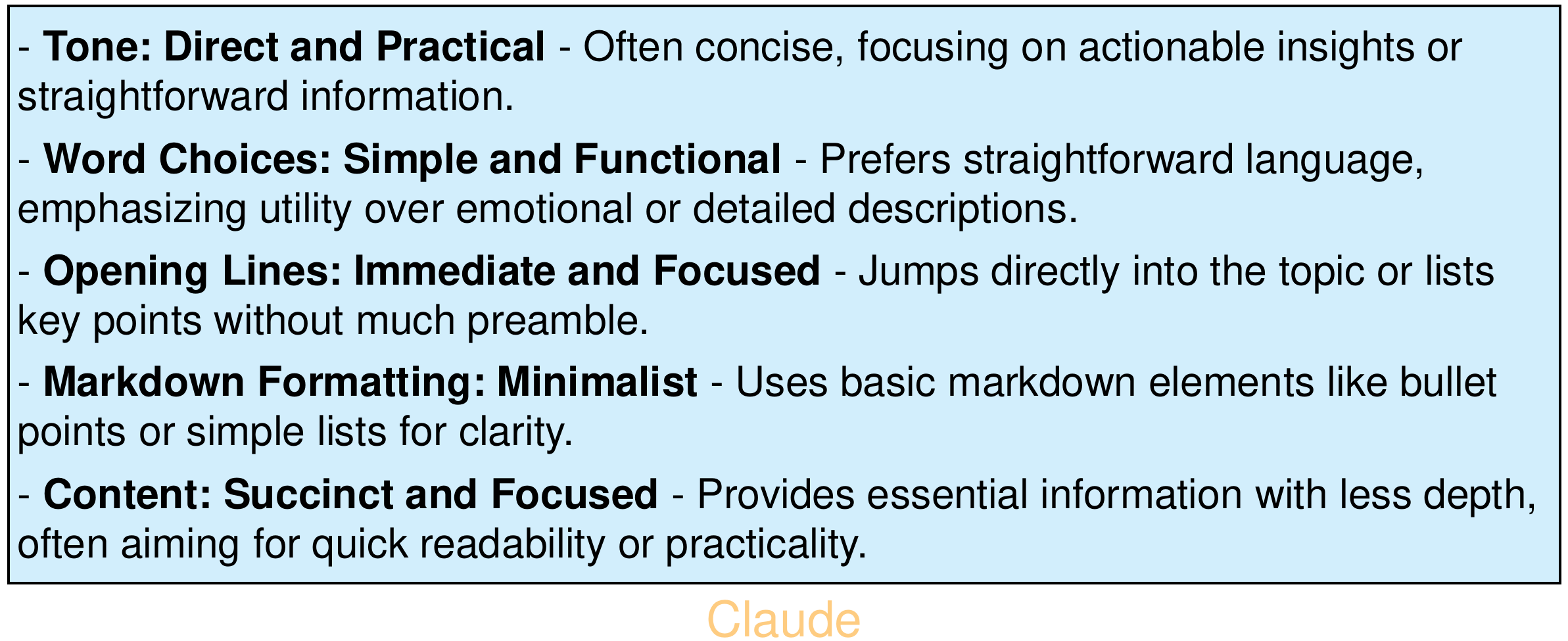}
    \vspace{-1ex}
    \subcaption{{\grok} as the LLM judge.}
    \vspace{2.5ex}
    \end{subfigure}

    \begin{subfigure}[h]{\linewidth}
    \includegraphics[width=.49\linewidth]{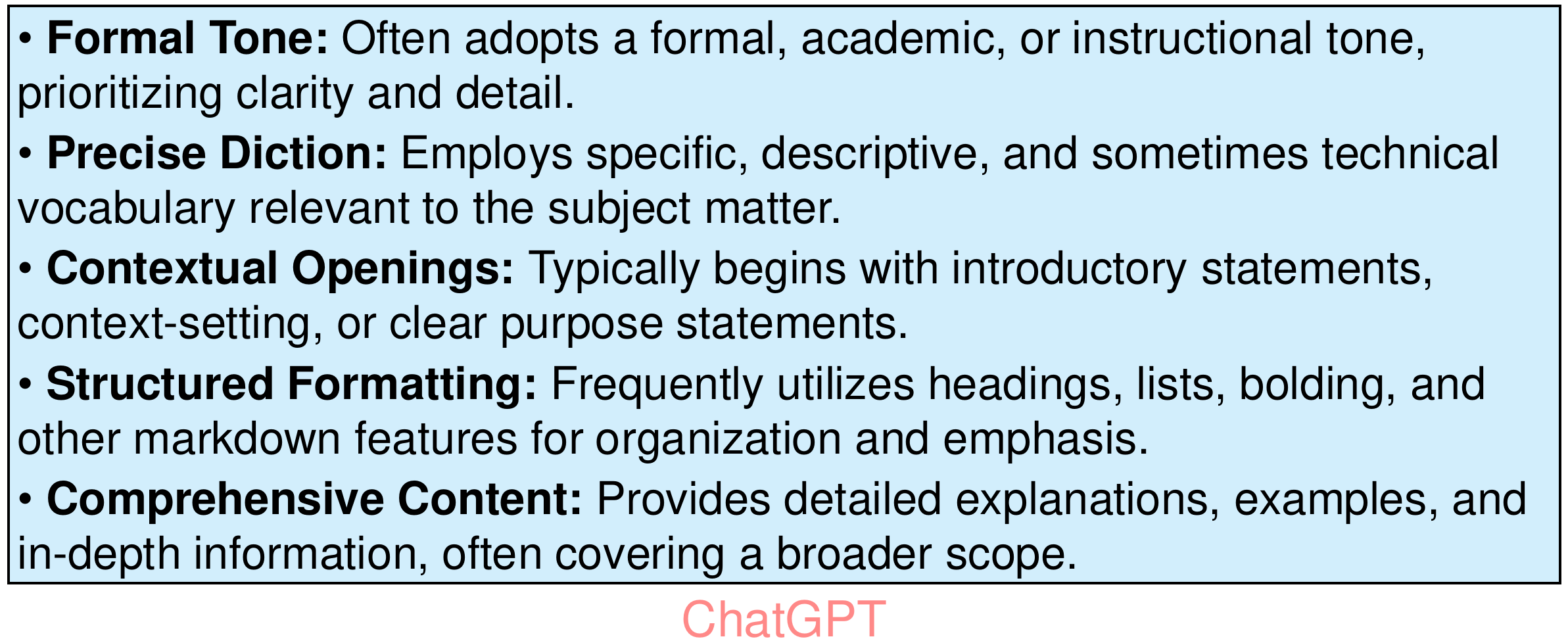}~
    \includegraphics[width=.49\linewidth]{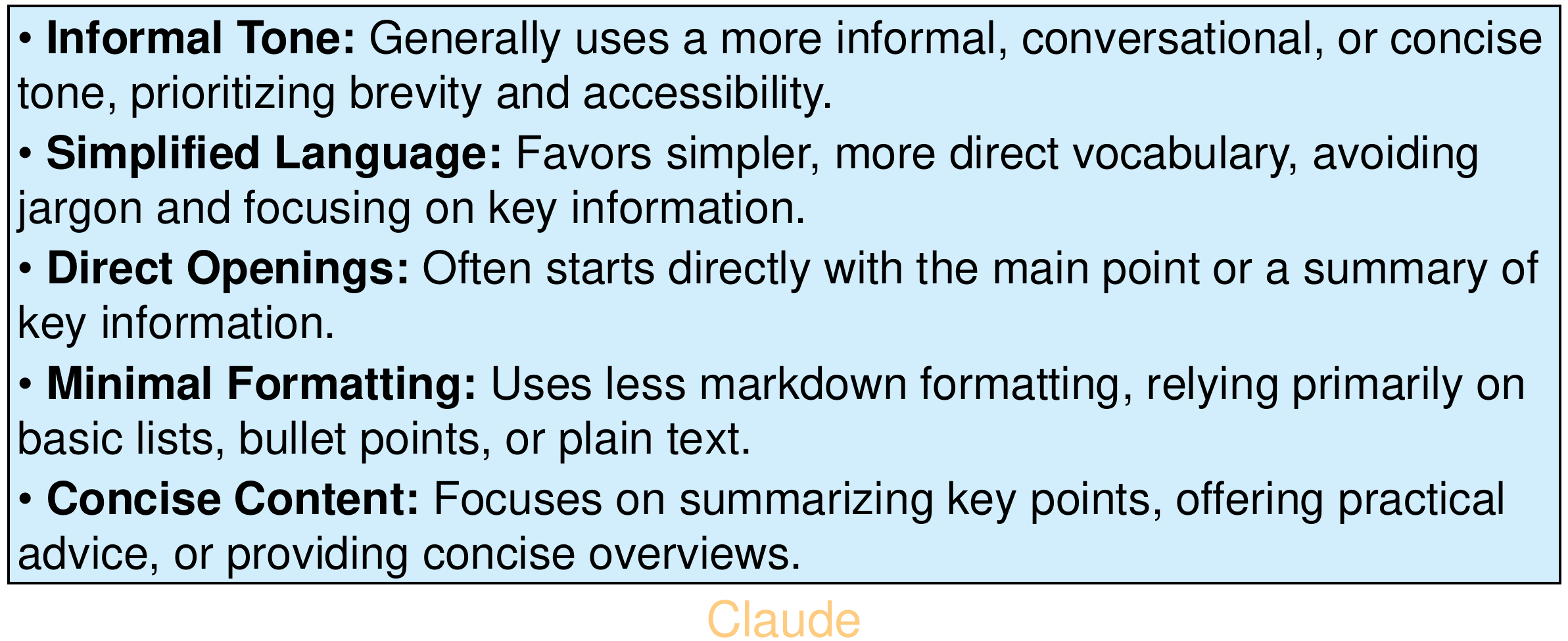}
    \vspace{-1ex}
    \subcaption{{\gemini} as the LLM judge.}
    \end{subfigure}
    \caption{Results of our open-ended language analysis on \textcolor{red}{{\gpt}} and \textcolor{yellow}{{\claude}} with different LLM judges.}
    \label{fig:open-ended_llm_judge}
\end{figure*}

\textbf{Ablation on LLM judges.} Here we demonstrate our findings in Figure~\ref{fig:open-ended} of Section~\ref{sec:semantics} remains consistent under several LLM judges. Specifically, we change the LLM judge from {\gpt} to {\claude}, {\grok}, and {\gemini}. We show the results in Figure~\ref{fig:open-ended_llm_judge}. Regardless of the choice of LLM judges, our language analysis reveals that {\gpt} often uses detailed explanations and complex formatting structures, whereas {\claude} emphasizes key contents without extensive elaboration. 

\begin{figure*}[th]
    \centering
    \begin{subfigure}[h]{\linewidth}
    \includegraphics[width=.49\linewidth]{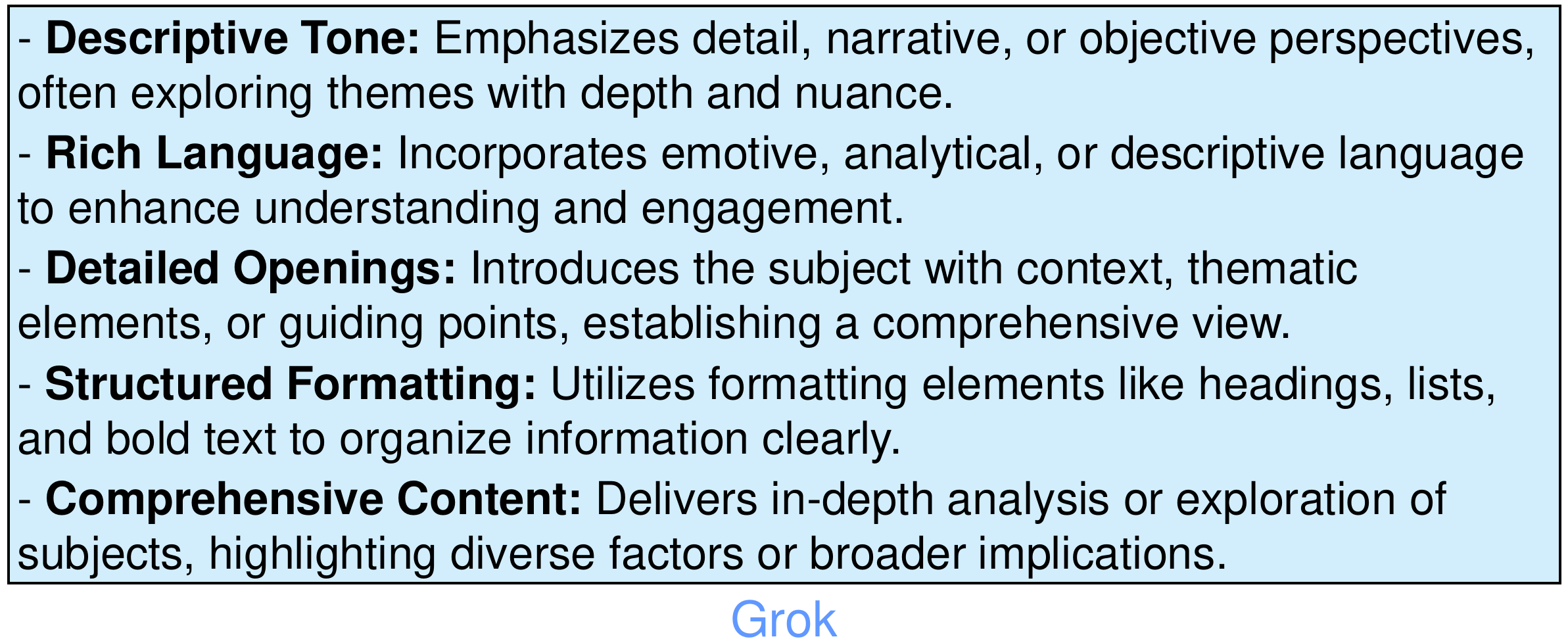}~
    \includegraphics[width=.49\linewidth]{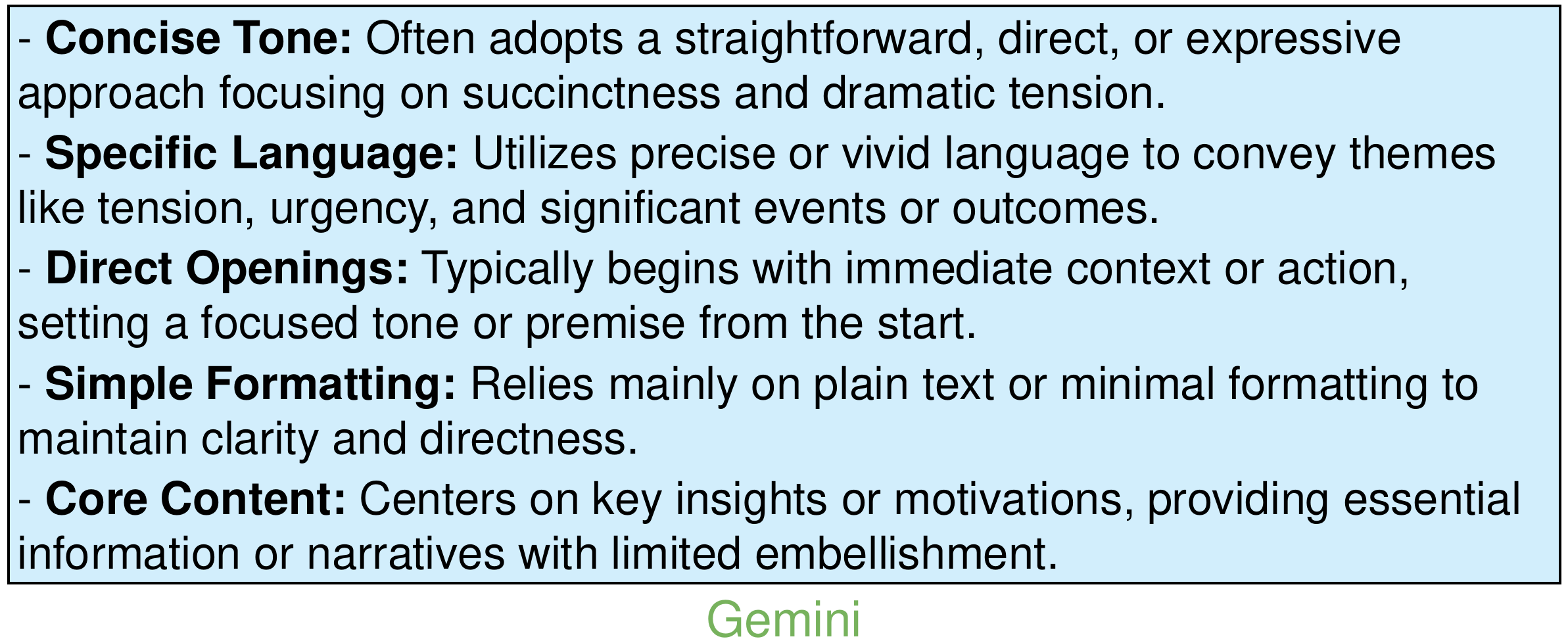}
    \vspace{-1ex}
    \subcaption{chat APIs}
    \vspace{2.5ex}
    \end{subfigure}
    \centering
    \begin{subfigure}[h]{\linewidth}
    \centering
    \includegraphics[width=.49\linewidth]{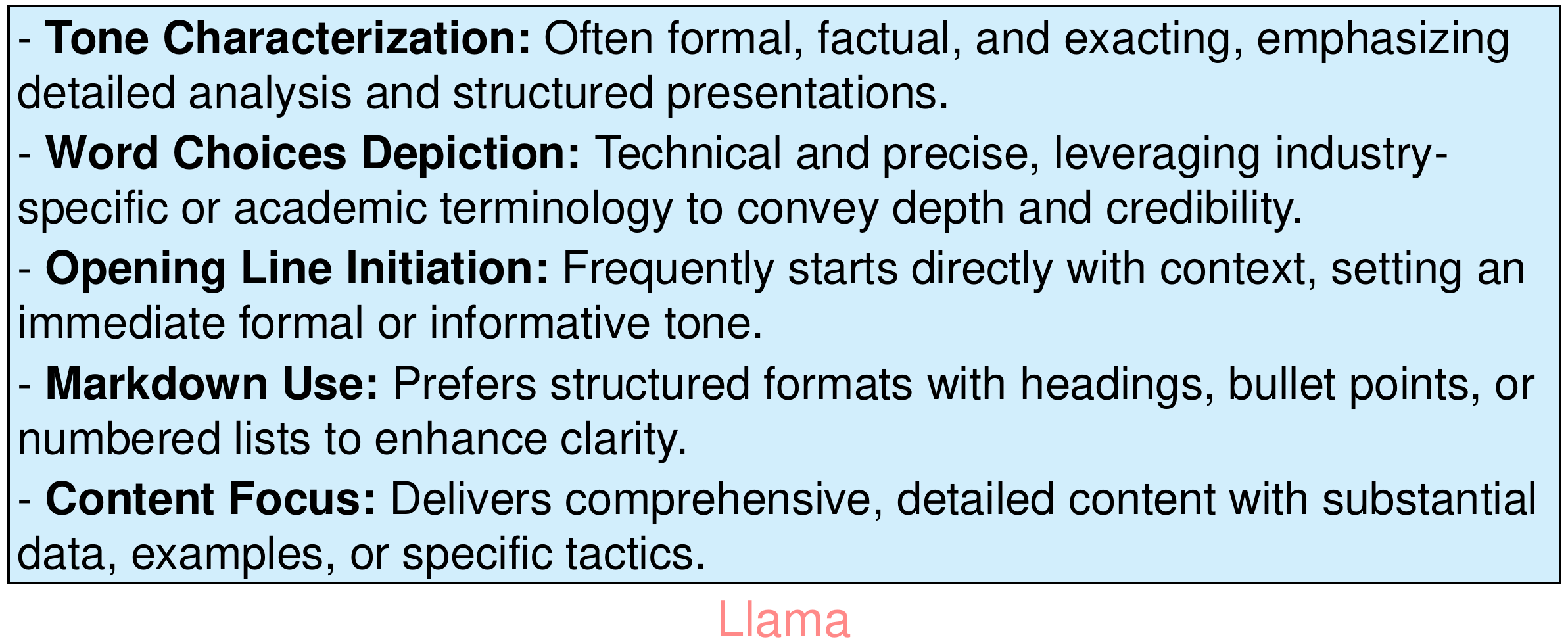}~
    \includegraphics[width=.49\linewidth]{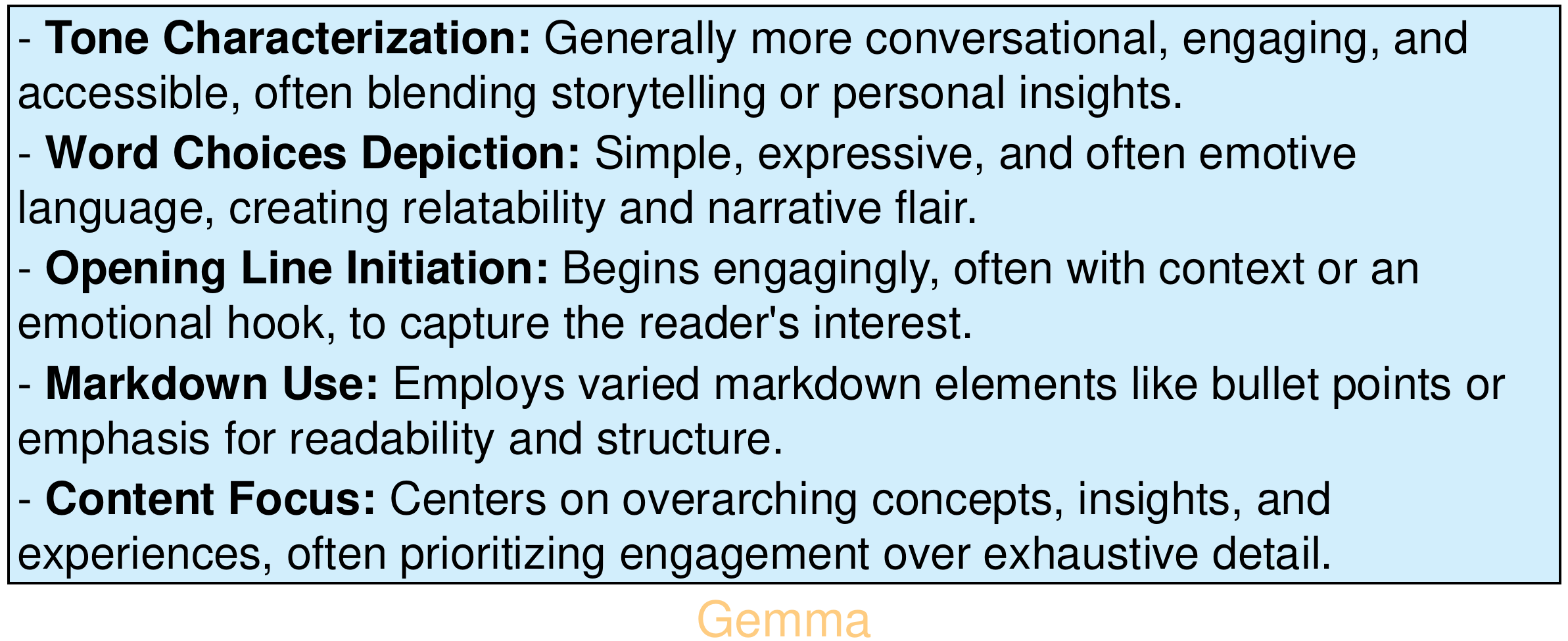}
    \vspace{2ex}
    \end{subfigure}
    \begin{subfigure}[h]{\linewidth}
    \includegraphics[width=.49\linewidth]{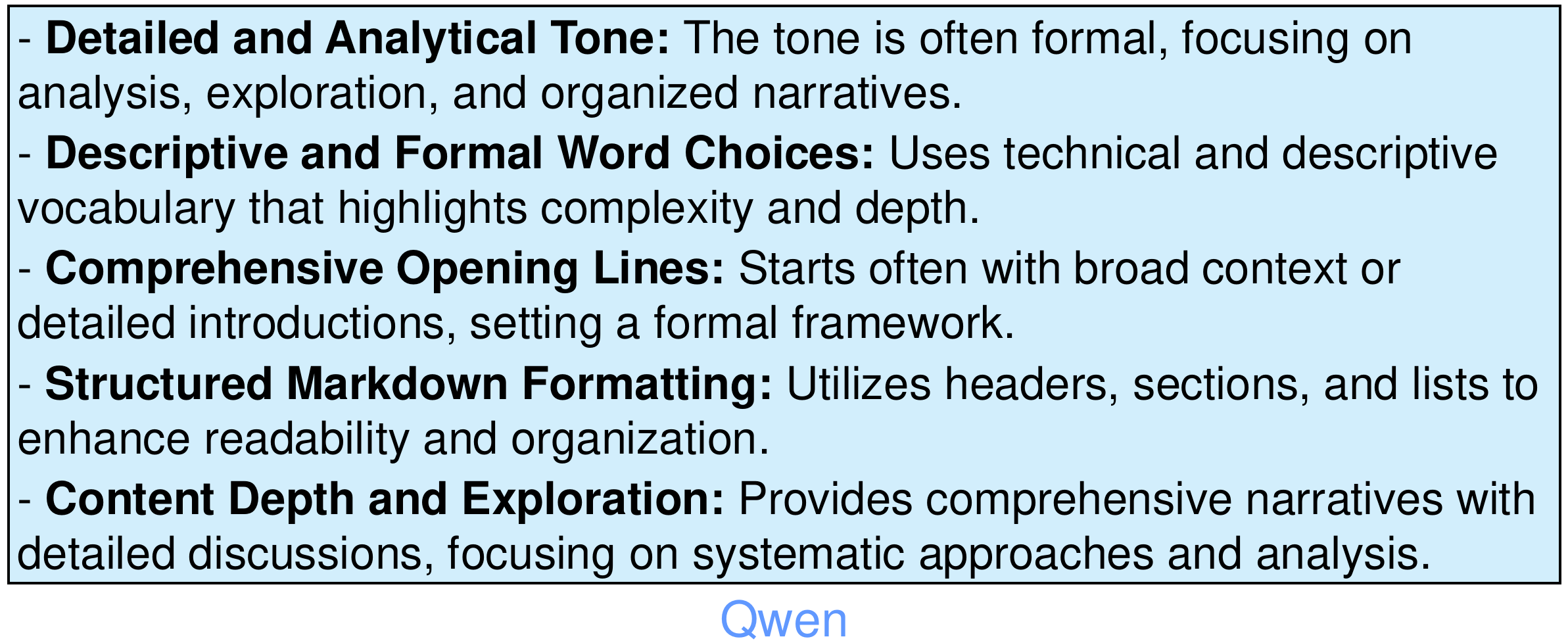}~
    \includegraphics[width=.49\linewidth]{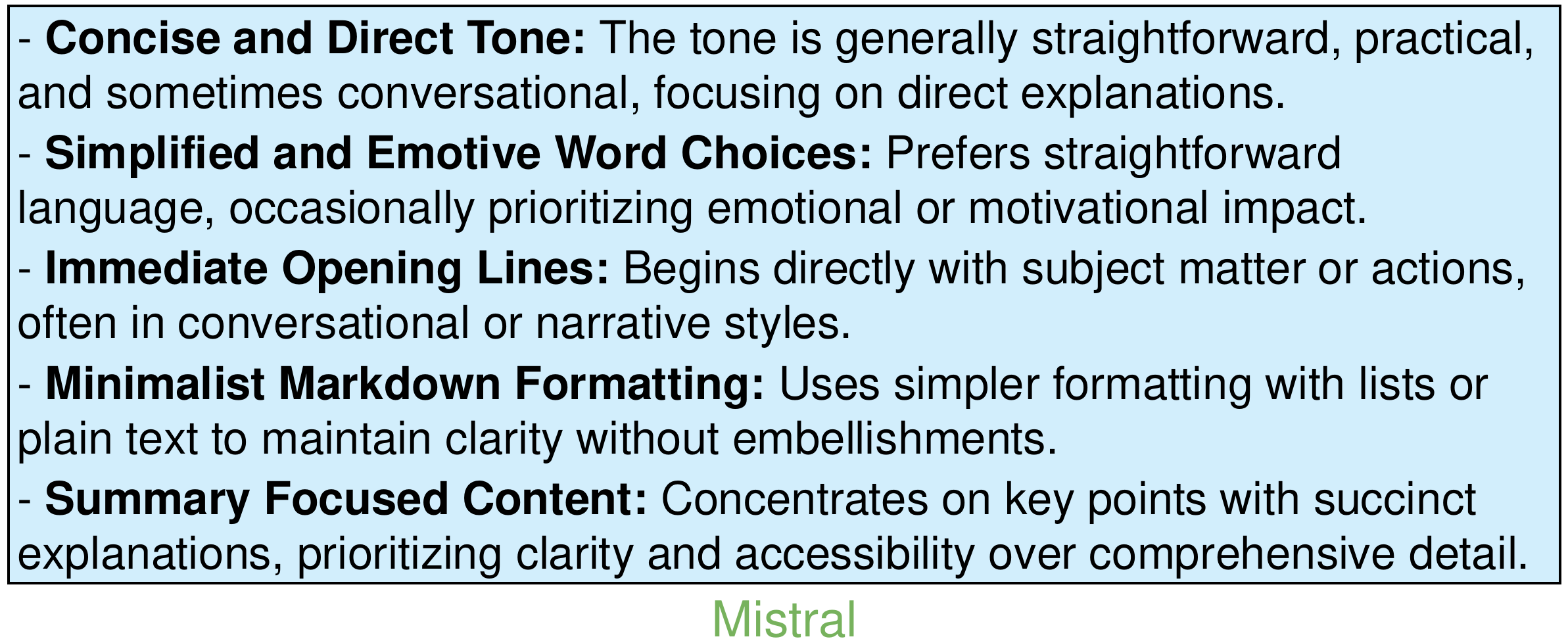}
    \vspace{-1ex}
    \subcaption{instruct LLMs}
    \end{subfigure}
    \caption{Additional results of our open-ended language analysis on chat APIs (\textit{top}) and instruct LLMs (\textit{bottom}).}
    \label{fig:open-ended_full}
\end{figure*}

\clearpage
\newpage
\textbf{Open-ended language analysis results on other LLMs.} In Section~\ref{sec:semantics}, we presented the results of open-ended language analysis for {\gpt} and {\claude}. Here, we extend our analysis to other chat API models and instruct LLMs. The full results are shown in Figure~\ref{fig:open-ended_full}, where we use {\gpt} as the LLM judge to compare responses generated by two models within the same category (chat APIs / instruct LLMs). Our analysis highlights several interesting characteristics of each model. For example, {\grok}'s responses tend to feature rich language and comprehensive content, whereas {\gemini}'s outputs are more concise with direct openings. 

\clearpage
\newpage
\section{Response Demonstrations}\label{appendix:response_demonstration}
In this part of the appendix, we present examples of LLM responses. Table~\ref{appendix:fig:prompt-gpt4o} and \ref{appendix:fig:prompt-llama3.1-8b-it} illustrate responses before and after our prompt-level interventions (Section~\ref{sec:main_control}). Table~\ref{appendix:fig:markdown_response} shows the transformed responses when only markdown elements are retained (Section~\ref{sec:markdown}). 

Table~\ref{appendix:fig:gpt-4o-rewrite} presents the LLM rewritten responses (Section~\ref{sec:semantics}). 
For paraphrasing, we find that formatting style remains largely unchanged, \eg, the number of enumerated lists are the same. Most of the differences lie in their word choices, paraphrased texts use different words with similar meanings but do not change the high-level semantic meaning of the original texts.

Additionally, Table~\ref{appendix:fig:open-ended-language-analysis} provides a pairwise comparison of responses, supporting our findings in Table~\ref{fig:open-ended}. Table~\ref{appendix:fig:response_chatgpt}, ~\ref{appendix:fig:response_claude}, ~\ref{appendix:fig:response_grok}, ~\ref{appendix:fig:response_gemini}, ~\ref{appendix:fig:response_deepseek}, ~\ref{appendix:fig:response_llama3.1-8b-it}, ~\ref{appendix:fig:response_gemma2-9b-it}, ~\ref{appendix:fig:response_qwen2.5-7b-it}, ~\ref{appendix:fig:response_mistral-v3-7b-it}, ~\ref{appendix:fig:response_llama3.1-8b-base}, ~\ref{appendix:fig:response_gemma2-9b-base}, ~\ref{appendix:fig:response_qwen2.5-7b-base}, ~\ref{appendix:fig:response_mistral-v3-7b-base} contain example responses from each LLM considered in Section~\ref{sec-evaluate-idiosyncrasies}, including {\gpt}, {\claude}, {\grok}, {\gemini}, {\deepseek}, Llama3.1-8b (instruct), Gemma2-9b (instruct), Qwen2.5-7b (instruct), Mistral-v3-7b (instruct), Llama3.1-8b (base), Gemma2-9b (base), Qwen2.5-7b (base), and Mistral-v3-7b (base).

\begin{table*}[h]
    \centering
    \includegraphics[width=.95\linewidth]{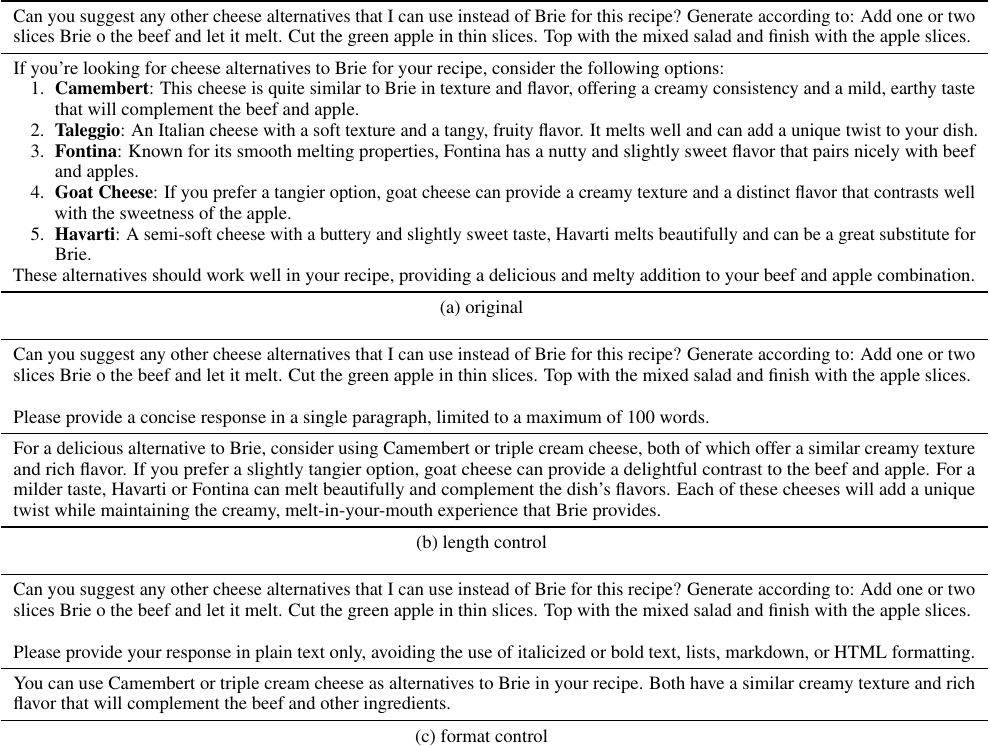}
    \caption{Examples of prompt-level interventions on \gpt.}
    \label{appendix:fig:prompt-gpt4o}
\end{table*}

\begin{table*}[h]
    \centering
    \includegraphics[width=.95\linewidth]{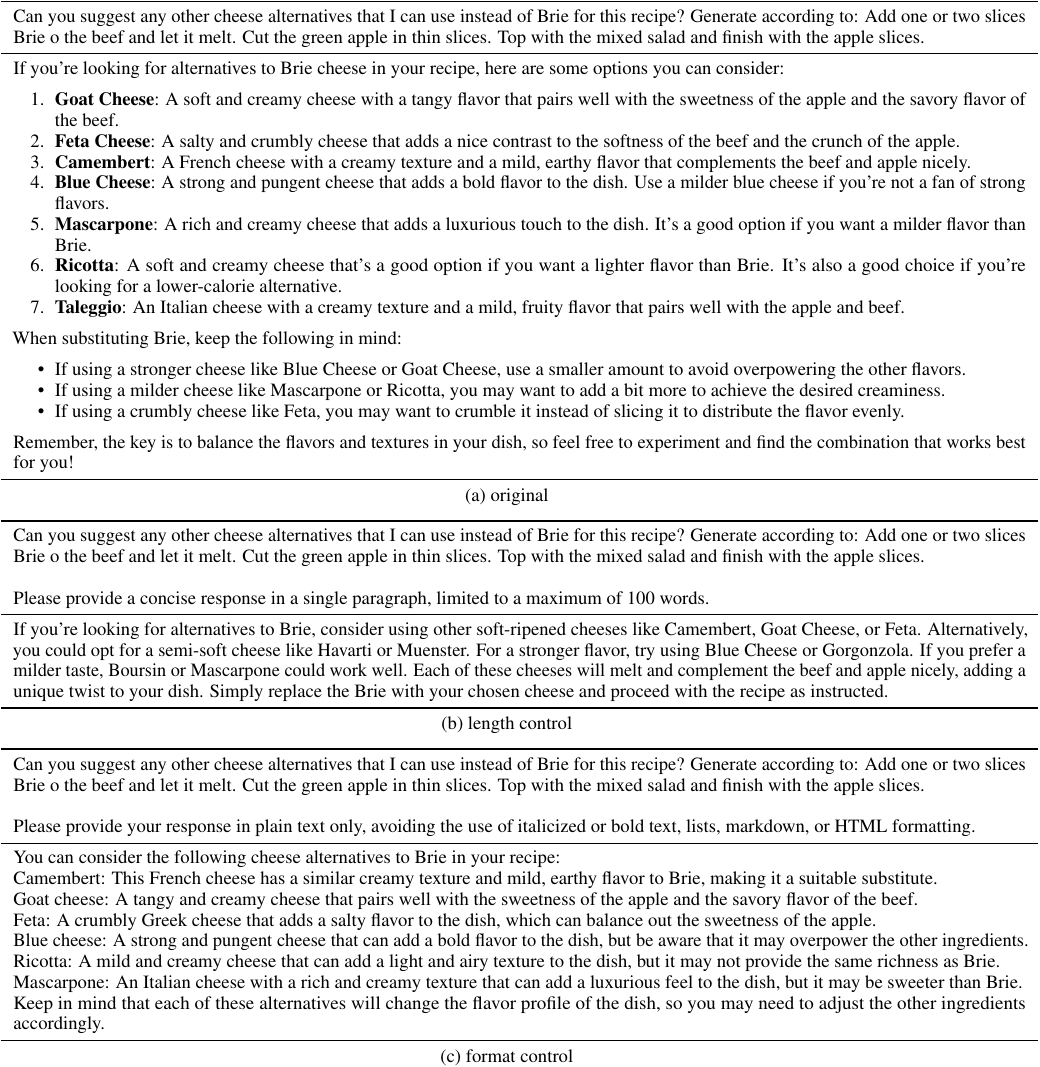}
    \caption{Examples of prompt-level interventions on Llama3.1-8b-Instruct.}
    \label{appendix:fig:prompt-llama3.1-8b-it}
\end{table*}

\begin{table*}[h]
    \centering
    \includegraphics[width=.95\linewidth]{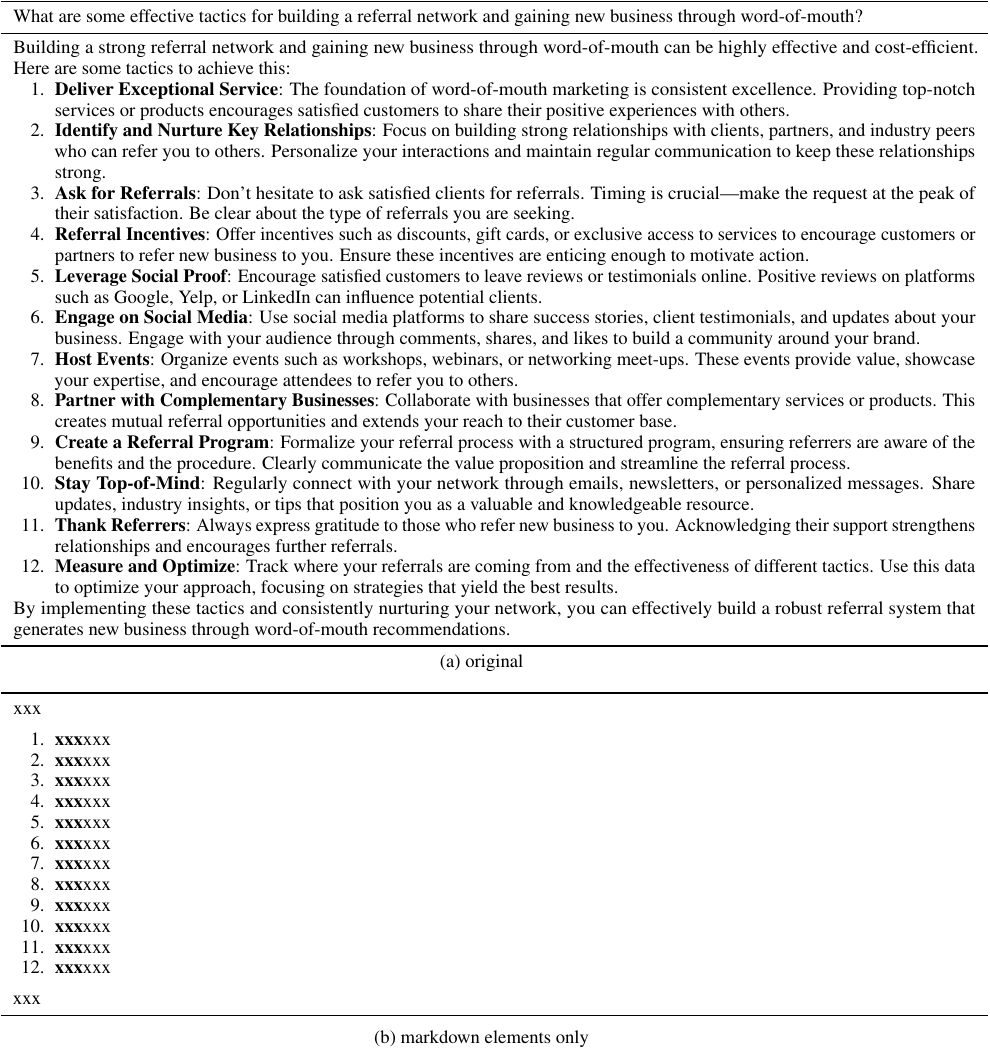}
    \caption{Examples of only using markdown formatting elements and replacing text content with ``xxx'' placeholders.}
    \label{appendix:fig:markdown_response}
\end{table*}

\begin{table*}[h]
    \centering
    \includegraphics[width=.95\linewidth]{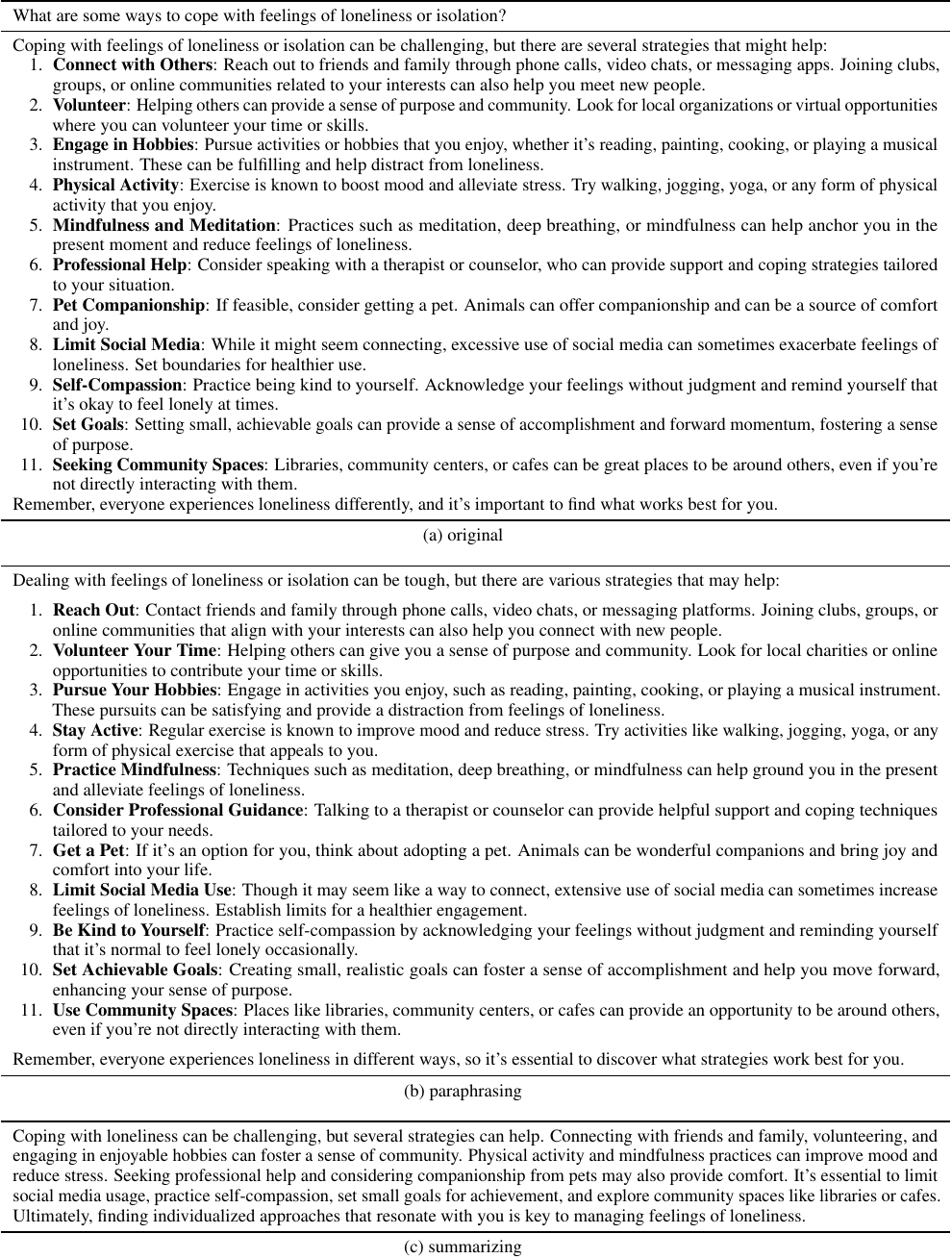}
    \caption{Examples of LLM rewriting using GPT-4o-mini.}
    \label{appendix:fig:gpt-4o-rewrite}
\end{table*}

\begin{table*}[h]
    \centering
    \includegraphics[width=.95\linewidth]{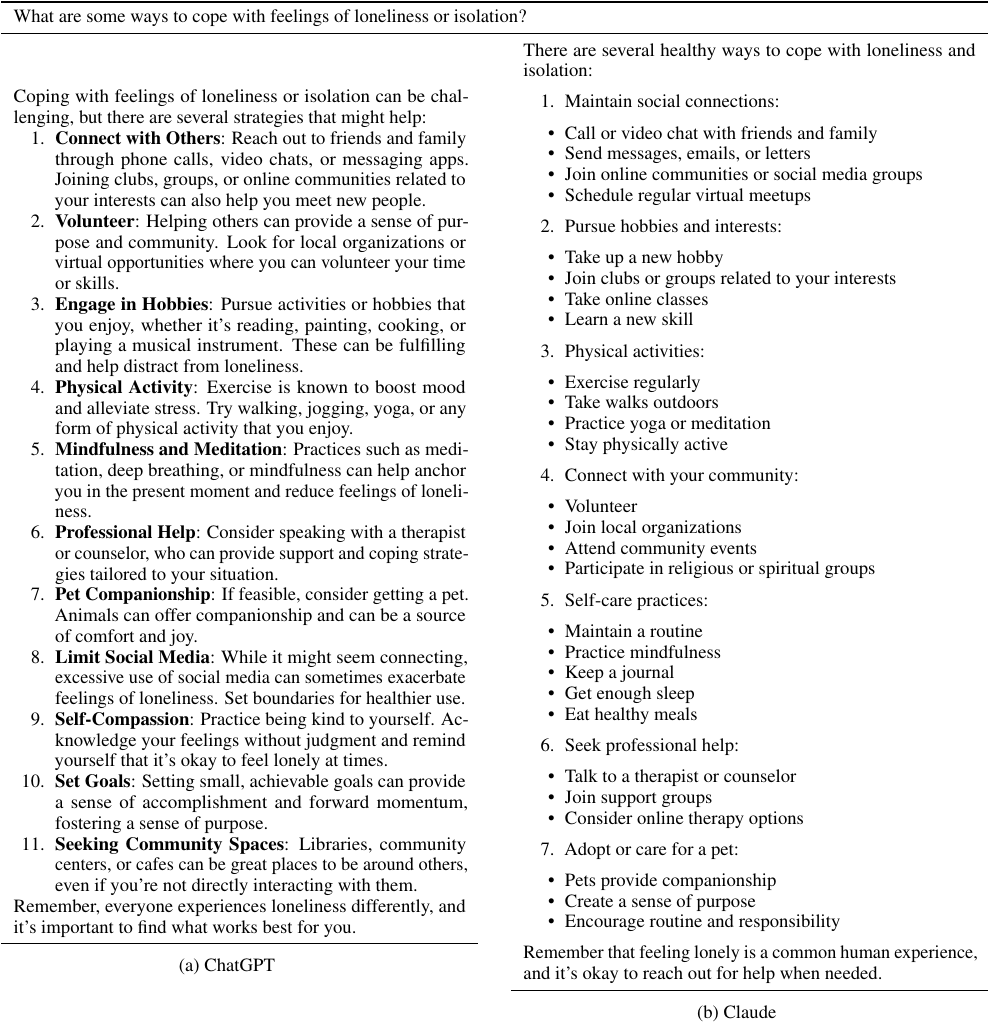}
    \caption{Examples from {\gpt} and {\claude} that illustrate results of our open-ended language analysis in Figure~\ref{fig:open-ended}.}
    \label{appendix:fig:open-ended-language-analysis}
\end{table*}

\begin{table*}[h]
    \centering
    \includegraphics[width=.95\linewidth]{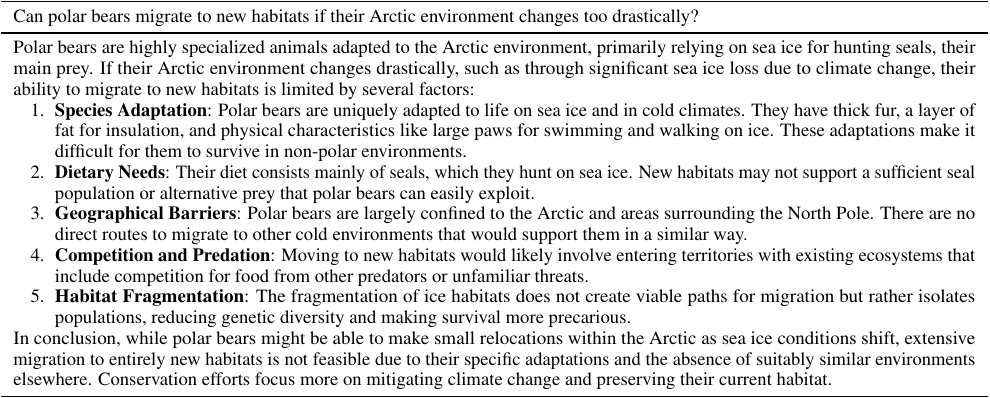}
    \caption{Example response of {\gpt}.}
    \label{appendix:fig:response_chatgpt}
\end{table*}

\begin{table*}[h]
    \centering
    \includegraphics[width=.95\linewidth]{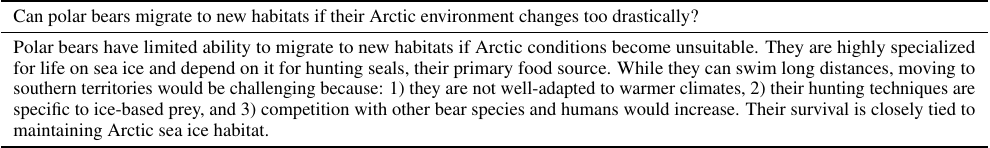}
    \caption{Example response of {\claude}.}
    \label{appendix:fig:response_claude}
\end{table*}

\begin{table*}[h]
    \centering
    \includegraphics[width=.95\linewidth]{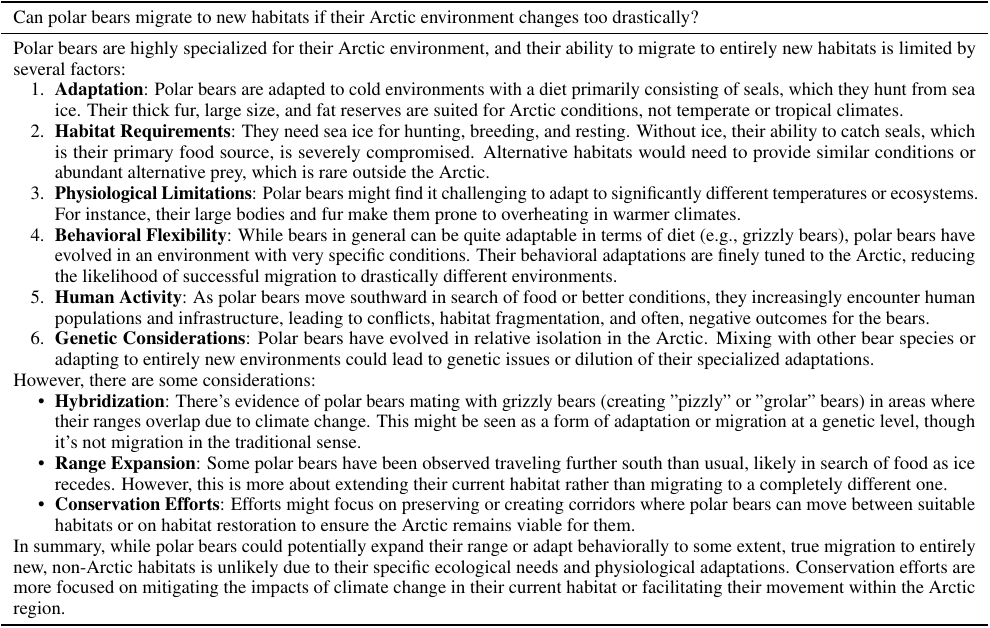}
    \caption{Example response of {\grok}.}
    \label{appendix:fig:response_grok}
\end{table*}

\begin{table*}[h]
    \centering
    \includegraphics[width=.95\linewidth]{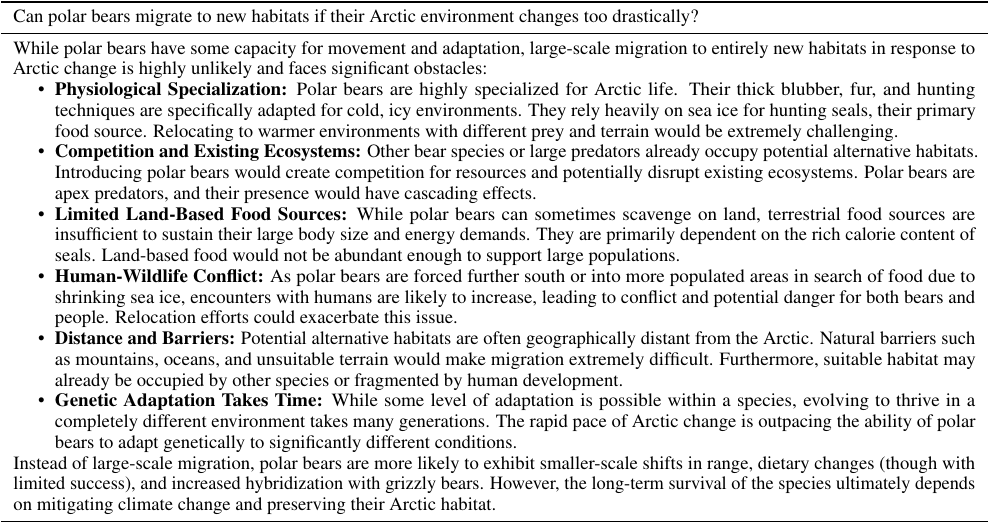}
    \caption{Example response of {\gemini}.}
    \label{appendix:fig:response_gemini}
\end{table*}

\begin{table*}[h]
    \centering
    \includegraphics[width=.95\linewidth]{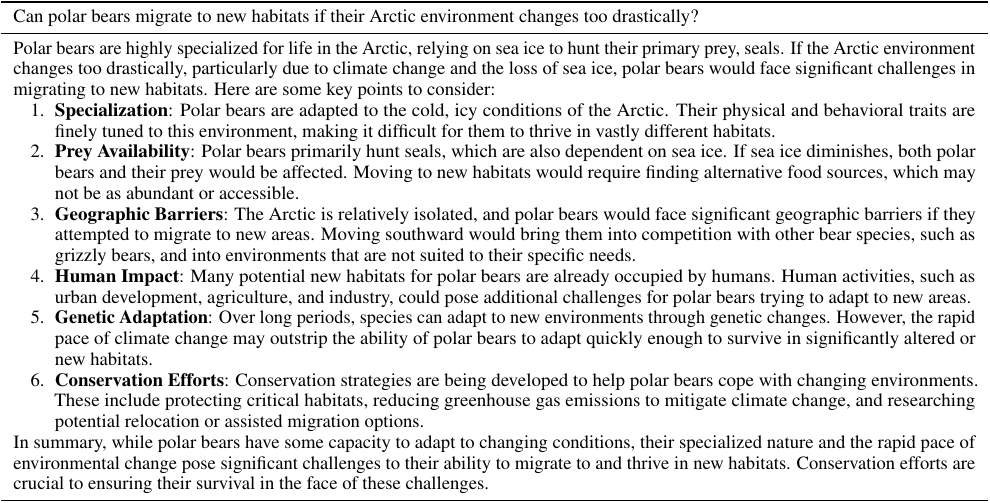}
    \caption{Example response of {\deepseek}}
    \label{appendix:fig:response_deepseek}
\end{table*}

\begin{table*}[h]
    \centering
    \includegraphics[width=.95\linewidth]{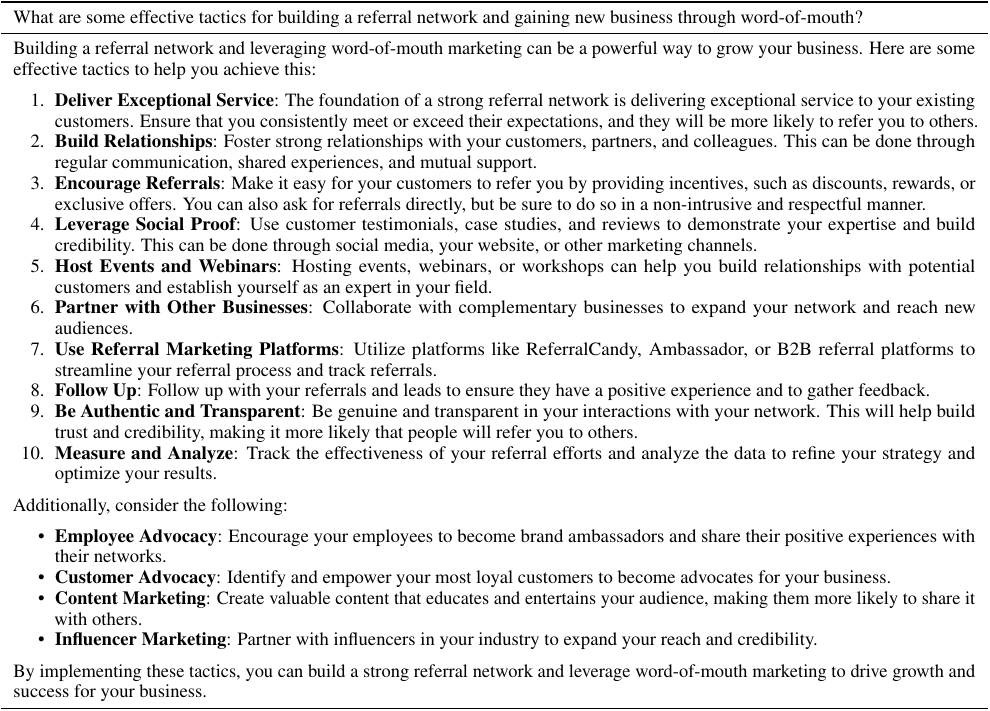}
    \caption{Example response of Llama3.1-8b (instruct).}
    \label{appendix:fig:response_llama3.1-8b-it}
\end{table*}

\begin{table*}[h]
    \centering
    \includegraphics[width=.95\linewidth]{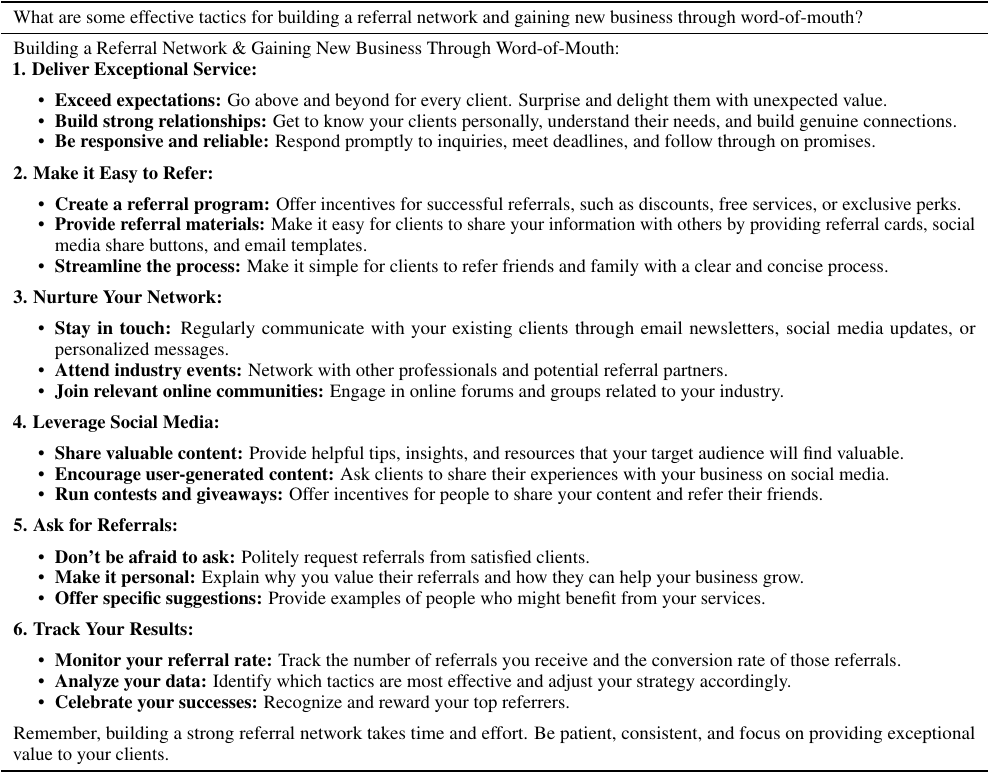}
    \caption{Example response of Gemma2-9b (instruct).}
    \label{appendix:fig:response_gemma2-9b-it}
\end{table*}

\begin{table*}[h]
    \centering
    \includegraphics[width=.95\linewidth]{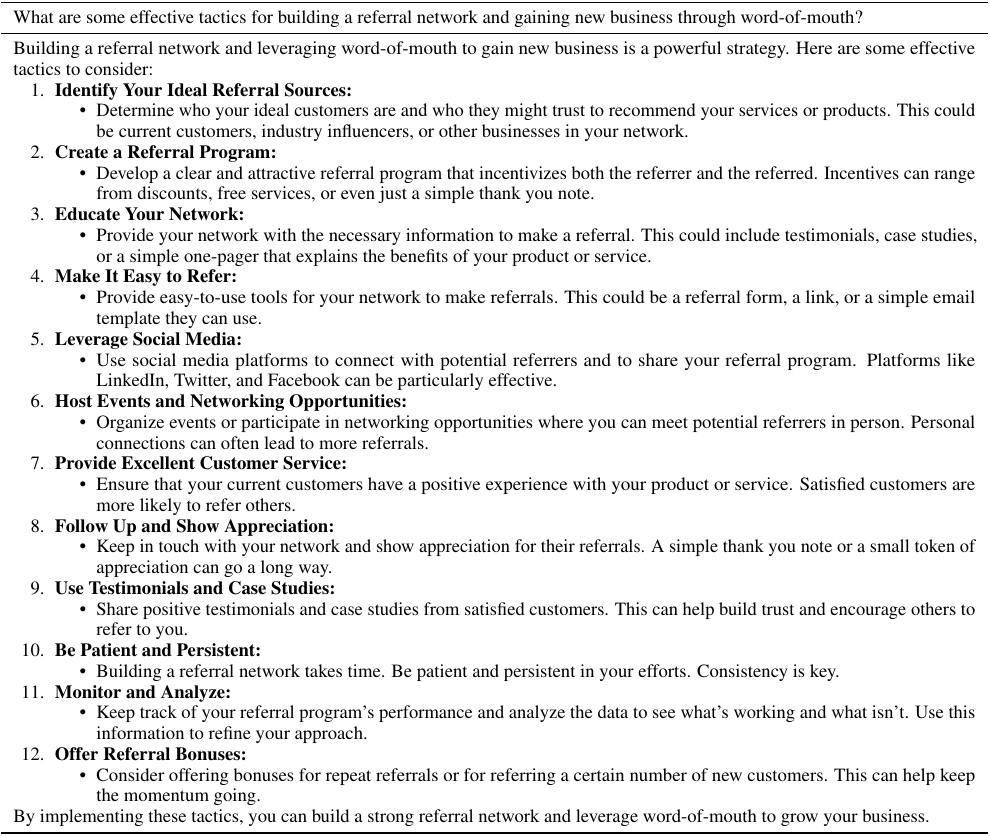}
    \caption{Example response of Qwen2.5-7b (instruct).}
    \label{appendix:fig:response_qwen2.5-7b-it}
\end{table*}

\begin{table*}[h]
    \centering
    \includegraphics[width=.95\linewidth]{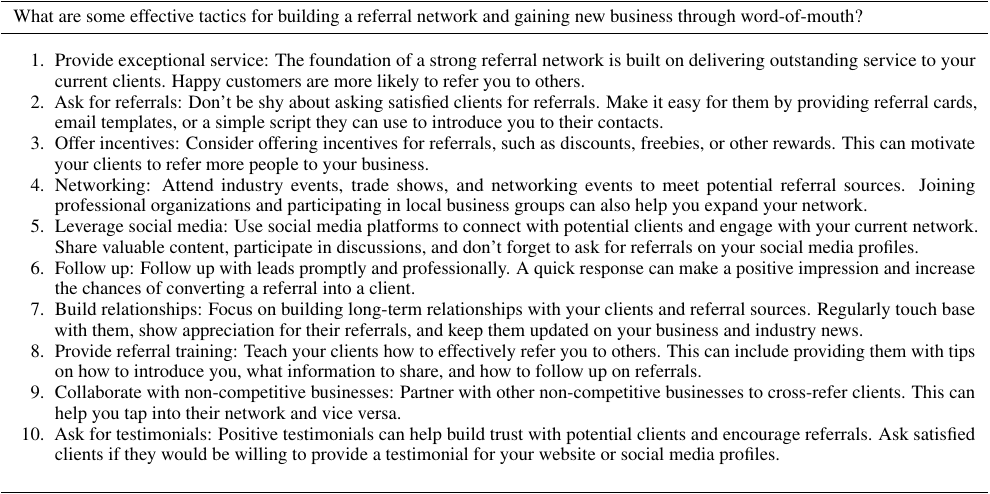}
    \caption{Example response of Mistral-v3-7b (instruct).}
    \label{appendix:fig:response_mistral-v3-7b-it}
\end{table*}

\clearpage
\newpage
\begin{table*}[h]
    \centering
    \includegraphics[width=.95\linewidth]{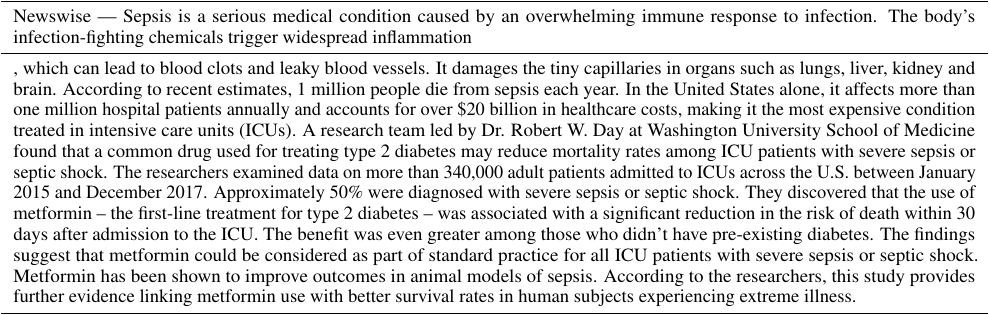}
    \caption{Example response of Llama3.1-8b (base).}
    \label{appendix:fig:response_llama3.1-8b-base}
\end{table*}

\begin{table*}[h]
    \centering
    \includegraphics[width=.95\linewidth]{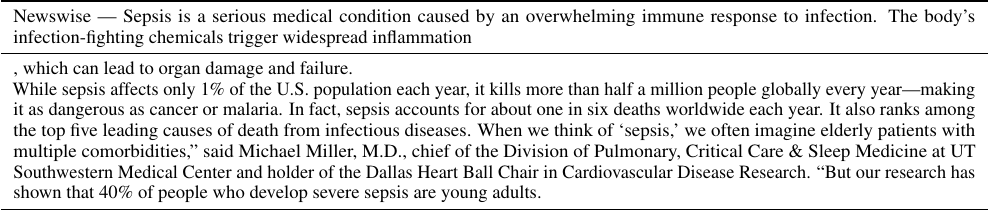}
    \caption{Example response of Gemma2-9b (base).}
    \label{appendix:fig:response_gemma2-9b-base}
\end{table*}

\begin{table*}[h]
    \centering
    \includegraphics[width=.95\linewidth]{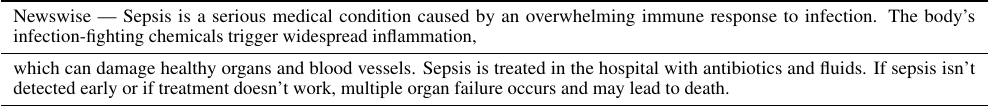}
    \caption{Example response of Qwen2.5-7b (base).}
    \label{appendix:fig:response_qwen2.5-7b-base}
\end{table*}

\begin{table*}[h]
    \centering
    \includegraphics[width=.95\linewidth]{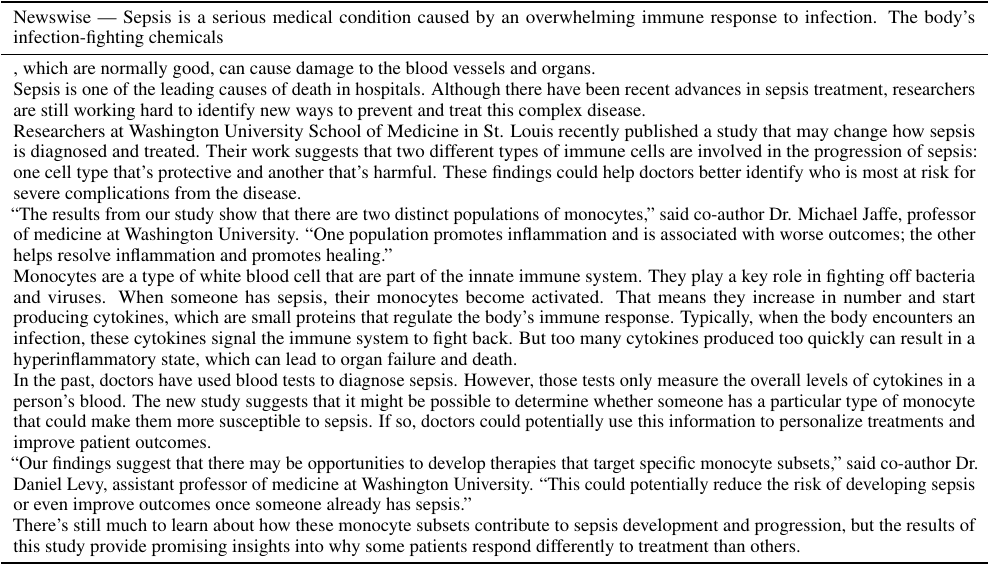}
    \caption{Example response of Mistral-v3-7b (base).}
    \label{appendix:fig:response_mistral-v3-7b-base}
\end{table*}



\end{document}